\documentclass[sort,numbers]{article}

\usepackage{microtype}
\usepackage{graphicx}
\usepackage{subcaption}
\usepackage{booktabs} 
\usepackage[utf8]{inputenc}
\usepackage[T1]{fontenc}    %
\usepackage{wrapfig}
\usepackage{tcolorbox}
\usepackage{lipsum}  
\usepackage{listings}%
\usepackage{textcomp}%
\usepackage{siunitx}%
\usepackage{float}%
\usepackage{tabularx}%
\usepackage{longtable}

\newcolumntype{Y}{>{\raggedright\arraybackslash}X}

\usepackage{arxiv}
\setcitestyle{authoryear,open={(},close={)}}
\renewcommand{\cite}{\citep}

\usepackage{amsmath}
\usepackage{amssymb}
\usepackage{mathtools}
\usepackage{amsthm}
\usepackage{wrapfig} 
\usepackage{amssymb} 
\usepackage{pifont}  
\usepackage{xcolor}  

\usepackage[capitalize,noabbrev]{cleveref}

\usepackage{listings}
\usepackage{comment}
\usepackage{adjustbox}
\usepackage{xspace}
\usepackage{fancyvrb}
\usepackage[subtle]{savetrees}
\usepackage{enumitem}
\usepackage{titling}
\usepackage{makecell}%
\usepackage{multirow}%

\definecolor{lpink}{cmyk}{0, 0.7808, 0.4429, 0.1412}
\definecolor{aqua}{cmyk}{0.91, 0, 0.09, 0.36}
\definecolor{ao}{rgb}{0.0, 0.5, 0.0}
\definecolor{amber}{rgb}{1.0, 0.49, 0.0}
\definecolor{dblue}{rgb}{0.0, 0.0, 0.61}
\definecolor{burgundy}{rgb}{0.5, 0.0, 0.13}

\date{}

\begin{document}
\title{Reflections and New Directions \\for Human-Centered Large Language Models}
\author{%
    \normalsize{
        \parbox{\textwidth}{\centering
            \textbf{Caleb Ziems*}, \textbf{Dora Zhao*}, \\
            Rose E. Wang, Matthew Jörke, Ahmad Rushdi, Advit Deepak, Sunny Yu, Anshika Agarwal, Harshvardhan Agarwal, Gabriela Aranguiz-Dias, Aditri Bhagirath, Justine Breuch, Huanxing Chen, Ruishi Chen, Sarah Chen, Haocheng Fan, William Fang, Cat Gonzales Fergesen, Daniel Frees, Tian Gao, Ziqing Huang, Vishal Jain, Yucheng Jiang, Kirill Kalinin, Su Doga Karaca, Arpandeep Khatua, Teland La, Isabelle Levent, Miranda Li, Xinling Li, Yongce Li, Angela Liu, Minsik Oh, Nathan J. Paek, Anthony Qin, Emily Redmond, Michael J. Ryan, Aadesh Salecha, Xiaoxian Shen, Pranava Singhal, Shashanka Subrahmanya, Mei Tan, Irawadee Thawornbut, Michelle Vinocour, Xiaoyue Wang, Zheng Wang, Henry Jin Weng, Pawan Wirawarn, Shirley Wu, Sophie Wu, Yichen Xie, Patrick Ye, Sean Zhang, Yutong Zhang, Cathy Zhou, Yiling Zhao, James Landay, \textbf{Diyi Yang*}
        }
    }\\  \\ 
    \large{
        Stanford University
    }
}

\maketitle
\noindent 
\vspace{-0.2in}
\begin{abstract}
Large Language Models (LLMs) are increasingly shaping the private and professional lives of users, with numerous applications in business, education, finance, healthcare, law, and science. With this rise in global influence comes greater urgency to build, evaluate, and deploy these systems in a manner that prioritizes not only technical capabilities but also \textit{human} priorities. This work presents a framework for developing Human-Centered Large Language Models (HCLLMs), which integrates perspectives from Natural Language Processing (NLP), Human-Computer Interaction (HCI), and responsible AI. Considering the ethics, economics, and technical objectives of language modeling, we argue that model developers need to address human concerns, preferences, values, and goals, not only during a cursory post-training stage, but rather with rigor and care at every stage of the pipeline. This paper offers human-centered insights and recommendations for developers at each stage, from system design to data sourcing, model training, evaluation, and responsible deployment. Then we conclude with a case study, applying these insights to understand the future of work with HCLLMs.
\end{abstract}

\clearpage
\tableofcontents
\clearpage

\hypertarget{introduction}{\section{Introduction}}
\label{sec:introduction}

\begin{figure}
    \centering
    \includegraphics[width=\linewidth]{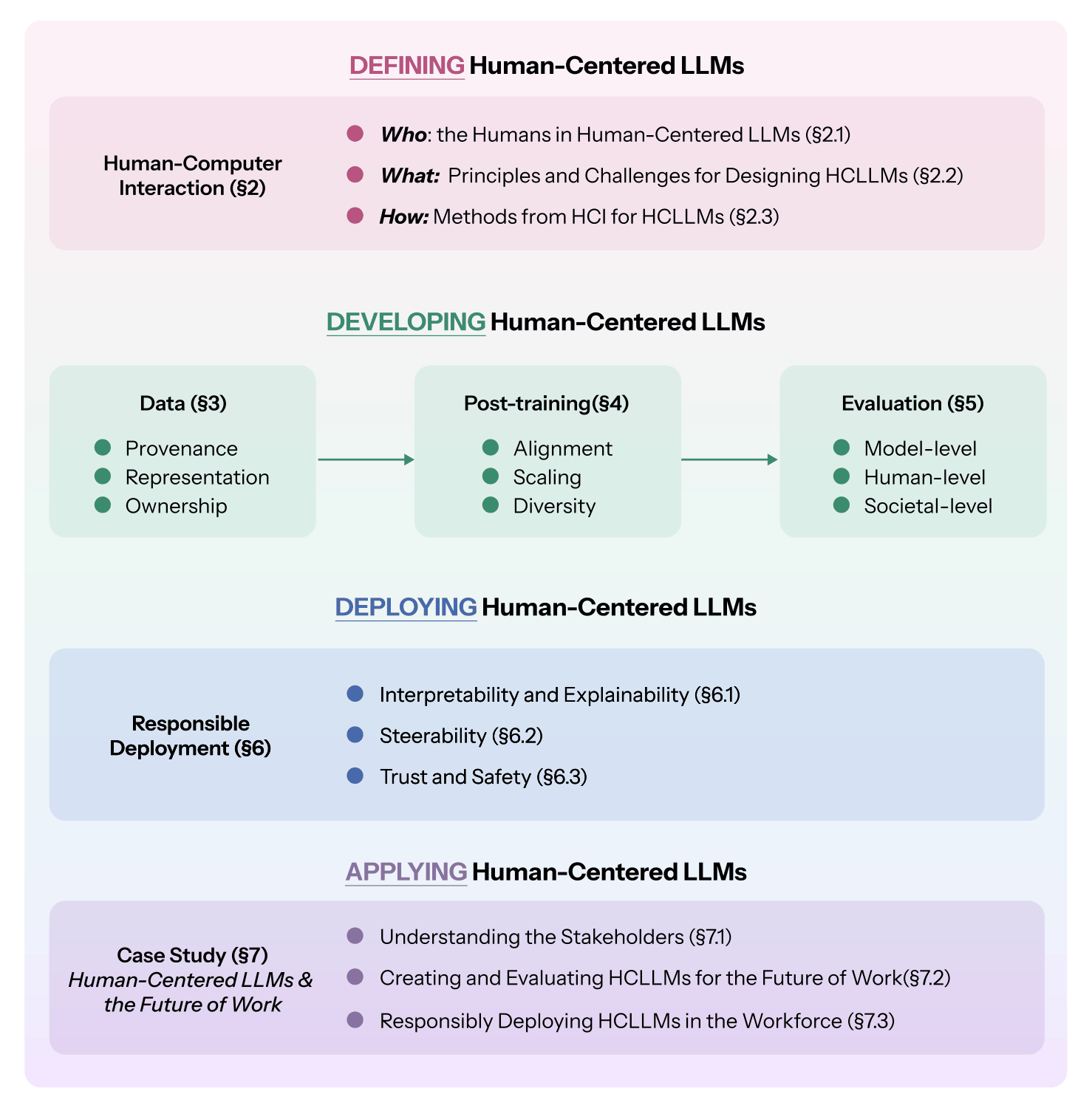}
    \caption{This survey has three core sections, focused on (1) defining, (2) developing, and (3) deploying human-centered LLMs (HCLLMs). In the first section, we conceptualize human-centeredness through HCI (\S\ref{sec:hci}) (who, what, and how). In the second, we illustrate how these principles appear across the LLM development pipeline (\S\ref{sec:data}, \ref{sec:nlp}, \ref{sec:evaluation}), and finally discuss considerations for deploying HCLLMs in a responsible manner (\ref{sec:responsible}). Finally, we synthesize takeaways from the three core sections into a case study on the deployment of HCLLMs for the future of work (\S\ref{sec:applications}).}
    \label{fig:overview}
\end{figure}
Large Language Models (LLMs) have transitioned from research artifacts to production infrastructure. They now power developer tools, enterprise copilots, search and recommendation systems, content moderation pipelines, and domain-specific assistants across healthcare \citep{thirunavukarasu2023large}, finance \citep{xie2024finben,nie2024survey}, education \citep{gan2023large,adiguzel2023revolutionizing,wang2024largelanguagemodelseducation}, science \citep{si2024llmsgeneratenovelresearch,zhang2024comprehensive} and law \citep{li2025legalagentbench,katz2024gpt,guha2024legalbench}. As LLMs are integrated into individual and collective processes, they can no longer be understood as isolated tools bounded by static performance metrics or leaderboard positions. LLMs are sociotechnical systems with global influence, and should be developed and evaluated in starkly more human terms. Are these models helpful, steerable, and safe under adversarial pressure, aligned across global markets, robust to distribution shift, and adaptable to evolving user goals and expectations? Do models comply with data governance regimes, privacy regulations, and ethical concerns around intellectual property? How can we build models that not only avoid harm but also actively contribute to human flourishing? Can LLMs do more than just passively assist humans; can they also actively collaborate with us as equal partners? 

This survey advances the framework of Human-Centered Large Language Modeling (HCLLMs) as a unifying lens for understanding and answering these questions. Rather than treating human-centered objectives as simple patches or alignment problems downstream of capability scaling, we argue that human-centered methods must be embedded across the entire LLM development pipeline, from data sourcing and filtering, to post-training and alignment, evaluation, deployment, and long-term maintenance (see Figure~\ref{fig:overview}). 

Importantly, we will demonstrate how human-centered objectives tend to resist universal solutions. The optimal path will depend both on who you ask and how you operationalize concepts like \textit{harm} and \textit{benefit}. Broad themes like \textit{transparency}, \textit{privacy}, \textit{safety}, and \textit{justice} frequently emerge \citep{jobin2019global}, but there will be significant variation in perspective on how these ideals should be implemented \citep{awad2018moral,jobin2019global}. Governments and non-profit organizations may codify the most dominant perspectives into laws and policies \citep{jobin2019global}, but high-level guidelines may fail to account for the nuances of real-world use \citep{hagendorff2020ethics}, and lag behind the rapid evolution of language models themselves \citep{auernhammer2020human}. In the face of these challenges, stakeholders often remain passive, which only endorses the status-quo \citep{kalluri2020don,crawford2021atlas,birhane2022values}.  

\begin{figure}[hb!]
    \centering
    \includegraphics[width=\linewidth]{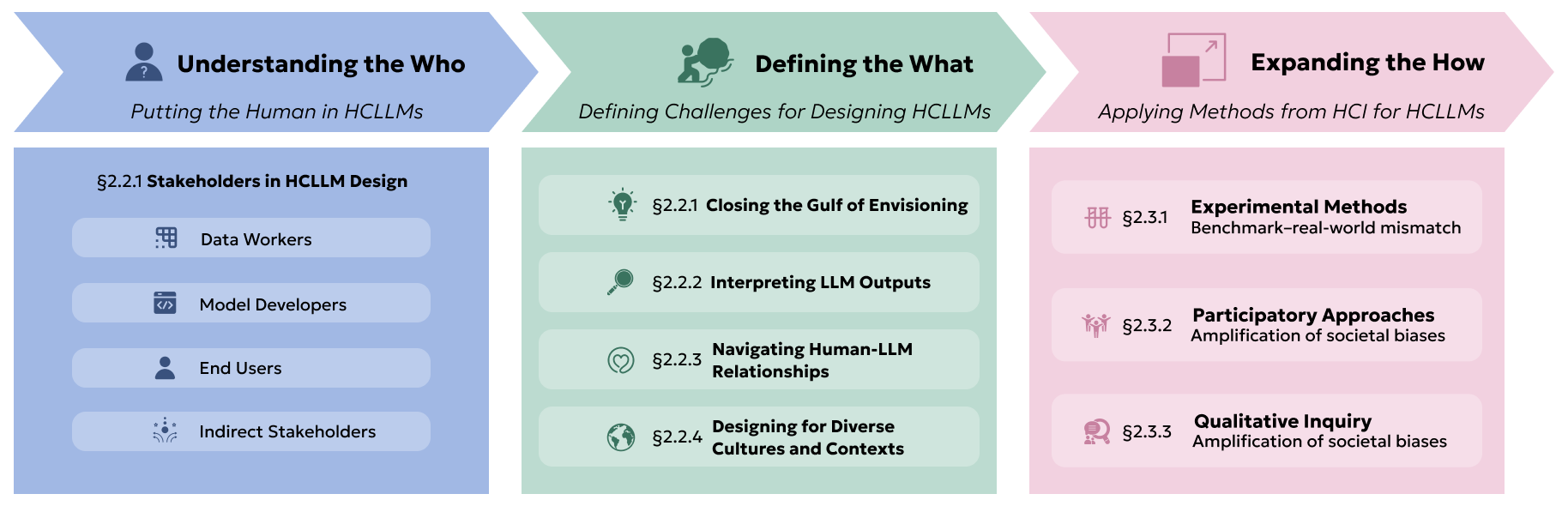}
    \caption{We can draw on the field of human-computer interaction (HCI) to help inform human-centered LLM design. The first in this process is understanding \emph{who} the relevant stakeholders are --- both direct and indirect --- in both the development and deployment of HCLLMs~\ref{subsec:hci_who}. Second, we identify a set of unique interaction challenges when it comes to designing HCLLMs~\ref{subsec:hci_challenges}. Finally, we discuss how HCI methods can be used for providing new design perspectives~\ref{subsec:hci_methods}. We synthesize these points in a case study on designing LLMs for motivating physical activity in Sec.~\ref{subsec:hci_case}.}
    \label{fig:hci}
\end{figure}

This survey paper elaborates on and endorses the alternative, Human-Centered Design \citep{capel2023human} (HCD), where users and other stakeholders are \textit{centrally} involved with ideating, building, evaluating, and deploying Large Language Models \citep{shneidermanHumanCenteredArtificialIntelligence2020, shneidermanHumanCenteredAI2022}. Their centrality at every stage of the design process is what distinguishes HCD from other instantiations of human factors design \citep{xu2019toward} that account for general user needs (e.g., \textit{transparency}) in only a small slice of the design or deployment process \citep{auernhammer2020human}. LLM development rarely \textit{centers} humans, but there is a growing body of research from Natural Language Processing (NLP) and Human-Computer Interaction (HCI) that points towards these ideals. We will cover this  foundation for \textit{HCLLMs} in an in-depth survey of relevant human factors approaches in \textbf{HCI} (\S\ref{sec:hci}) and  \textbf{NLP} (\S\ref{sec:nlp}), including more details on the \textbf{Data Pipeline} (\S\ref{sec:data}), and the \textbf{Evaluation} (\S\ref{sec:evaluation}) of LLMs. With this foundation established, we will return to the principle axes of \textbf{Responsible and Ethical Deployment} (\S\ref{sec:responsible}), like \textit{transparency}, \textit{privacy}, and safety. Synthesizing our discussion across these prior chapters, we will conclude with a concrete \textbf{Case Study} (\S\ref{sec:applications}) on the considerations of \textit{HCLLMs} for the future of work.
\clearpage
\clearpage
\hypertarget{hci}{\section{HCI for HCLLMs}}
\label{sec:hci}
How do we center humans in the design of LLMs? To start, we can turn to the field of human-computer interaction (HCI), which offers the foundational principles for realizing the vision of HCLLMs. In particular, HCI provides the established theories, methods, and frameworks for understanding and designing the critical interface between the human user and the complex system (see Figure~\ref{fig:hci}). This field has long grappled with how to make technology not just functional, but also usable, understandable, and aligned with human values and needs. 

In the first subsection of this chapter, we start by tracing how principles of \textbf{\textit{human-centered design}} apply to HCLLMs and understanding \emph{who} are the stakeholders for designing HCLLMs (\S\ref{subsec:hci_who}). We then discuss \textbf{\textit{design principles and challenges}} for creating HCLLMs in \S\ref{subsec:hci_challenges}. These challenges range in scope from the individual level (i.e., how we can improve end-user interactions with these models) to the societal level (i.e., how to account for the diverse cultures and contexts in which these models will be deployed). We then discuss how HCI \textit{\textbf{methodologies}} can be used in conjunction with techniques commonly used in NLP to build and evaluate HCLLMs. We conclude in \S\ref{subsec:hci_methods} with an overview of three methodological orientations --- experimental methods, participatory approaches, and qualitative inquiry --- from HCI and discuss how they can be adopted for HCLLMs. 

\subsection{Understanding the \emph{Who}: the Humans in HCLLMs}
\label{subsec:hci_who}
Human-centered design (HCD) is an approach that centers the needs and experiences of users in all steps of the design process. There are four guiding principles within HCD~\cite{hcd2025}:
\begin{enumerate}
    \item Centering real people, their needs, and experiences in the design process
    \item Solving the core problems 
    \item Thinking of everything as existing in an interconnected system 
    \item Engaging in iterative prototyping to find solutions
\end{enumerate}

When applying these principles of HCD to LLMs, it reframes the development away from solely optimizing technical performance toward understanding how these systems impact and fit into people's real lives. Rather than treating metrics such as accuracy or perplexity as the ultimate goal, HCD foregrounds the needs, values, and contexts of the humans who develop, use, and are affected by these models. Thus, to start, we must first understand \emph{who} the real people affected by LLMs are and \emph{what} are the problems they face. In our consideration, we account for people involved across the entire lifecycle of LLM development, starting from the data used to train these models or the objectives and principles that model developers prioritize to the way that end users interact with models. In this subsection, we provide an overview of important stakeholder groups and their corresponding roles in the development of HCLLMs: data workers, model developers, end users, and indirect stakeholders.

\paragraph{Humans as Data Workers.}
\textbf{Data workers} are group of stakeholders that are essential to the HCLLM ecosystem but also often overlooked. Data workers form the often invisible human labor that undergirds the datasets and methods that make LLM development possible. Within this category of data workers, we also have classes of individuals. For example, there are those who directly label datasets or rank model outputs as \textbf{data annotators}. Their judgments instantiate the ``human preferences'' that guide alignment, embedding positionality into what the model learns to value~\cite{ouyang2022training,kirk2024the}. While annotators are the ones directly labeling the data, selecting which values or ideologies are prioritized often comes from other parties who hold power or authority over data workers (e.g., project manager, client requesting annotated data)~\cite{miceli2020between,wang2022whose}. In addition to data annotators, there are also \textbf{safety and moderation workers}, who engage in various methods to expose and rectify potential harms in the model before deployment. Similar to content moderation in other domains, such as social media, this work often entails sustained exposure to toxic or disturbing material, raising serious questions about the ethics and mental health implications of this invisible labor~\cite{pendse2025testing}. Finally, there are \textbf{data subjects}, or the individuals whose internet data forms the bulk of the web corpora used to pre-train LLMs. Importantly, many of those who are data subjects are doing so unwittingly as their data is included without consent or even awareness~\cite{paullada2021data,birhane2021large}. Together, these data workers form the backbone of HCLLM development, yet their contributions are systematically undervalued and underprotected. 

\paragraph{Humans as Model Developers.}
Next, \textbf{model developers} of HCLLMs are the ones most immediately tasked with shaping the model architecture, training pipeline, and deliverable behaviors of HCLLMs. They make crucial decisions about what data to collect, how to preprocess it, and which training objectives and evaluation protocols to employ. Prior works have shown how the values of model developers can be implicitly imbued in resulting technical artifacts, such as in how quality filters are defined for data~\cite{gururangan-etal-2022-whose} and what research problems are prioritized~\cite{birhane2022values}. In designing HCLLMs, model developers must harmonize the trade-offs between performance, fairness, inclusivity, and safety. For example, they must ensure the fairness, privacy, and safety of the training data of LLMs (\S\ref{sec:data}), design comprehensive evaluation frameworks to capture real-world usage scenarios (\S\ref{sec:evaluation}), and devise governance mechanisms and transparency guidelines that help ensure responsible deployment (\S\ref{sec:responsible}). 

\paragraph{Humans as End Users.}
\textbf{End users} --- the individuals and groups who directly interact with LLM systems --- represent one of the largest and most diverse stakeholder groups. The general-purpose nature of LLMs means that end users span a vast range: students seeking homework help; professionals drafting reports; people with mental health concerns looking for support and companionship; creative writers brainstorming ideas; non-native speakers translating text; and countless others~\cite{handa2025economic,chatterji2025people,handa2025education}. Each user brings their own distinct goals, expertise levels, cultural backgrounds, and needs when interfacing with models. Thus, as LLMs shift from nascent technologies into being increasingly ingrained in users' lives, understanding their real-world impact becomes critical. Despite the outsized impact that LLMs may have on their lives, most end users have relatively little control or influence in the design and creation of these technologies. There are methods that allow end-users to exert some degree of agency over models, such as through fine-tuning, which offers a more heavyweight intervention~\cite{tan2024democratizing, wang2025end}. Research prototypes and product features also offer ways to tailor user interactions with LLMs through creating personalized memory stores or drawing on community knowledge banks~\cite{zhao2025knoll,memories,ryan2025synthesizeme,zhong2024memorybank}. While these personalization mechanisms offer some user agency, they also raise fundamental questions about the broader relationship between end users and LLMs.

Key questions related to end users include: How do these systems affect productivity, creativity, learning, and decision-making skills? What are the risks of over-reliance, deskilling, or perpetuating existing biases? And how can we design HCLLMs that empower users rather than constrain or harm them? Answering these questions requires moving beyond system capabilities --- bridging existing methods in NLP with those from HCI --- to examine how LLMs may reshape human capacities, agency, and well-being in practice.

\paragraph{Humans as Indirect Stakeholders}
Finally, even people who are not direct end users of LLMs, or \textbf{indirect stakeholders} can still be meaningfully impacted by them~\cite{friedman1996value}. If we consider an example of an LLM used to assist physicians with taking clinical notes, patients are also affected by this system even if they are not directly interfacing with it~\cite{haberle2024impact,korom2025ai}. Beyond patients, we can also think of many other potential indirect stakeholders in this example, such as families or caregivers, insurance providers, hospital administrators, and so on. Enumerating all indirect stakeholders can seem like an intractable problem; in part, this underscores that human-centered LLMs are not just technical artifacts; they are sociotechnical systems with many externalities to consider. There is no prescriptive formula for deciding which indirect stakeholders to prioritize; however, thinking about factors, such as who may not be well-represented in making design decisions, who is most likely to be harmed by such systems (both immediately and over longer time horizons), or conversely who may stand to benefit in ways that are not explicitly intended, can inform designers. More broadly, rather than treating indirect stakeholders as an overwhelming checklist, HCLLM designers should use this complexity as motivation to identify the most consequential stakeholders early and involve them throughout the design process, rather than only including them as an afterthought (Sec.~\ref{subsub:participatory}).

\subsection{Defining the \emph{What}: Principles and Challenges for Designing HCLLMs}
\label{subsec:hci_challenges}

While involving humans throughout the LLM lifecycle helps us understand who is affected by these systems, it also raises fundamental questions about \emph{how} to design for their needs. Designing human-centered LLMs requires more than simply optimizing model capabilities. It demands careful attention to how users interact with, make sense of, and form relationships with these systems across diverse contexts.

Designing human-AI interaction is not a new challenge. Pre-dating LLMs, prior work~\citet{yang2020humanai} has delineated the unique challenges of human-AI interaction, highlighting the (1) uncertainties about model capabilities and~(2) the complexity of AI outputs --- both of which are problems that remain even as model capabilities have improved. In tandem, many others have enumerated many best practices for designing human-AI interaction~\cite{amershi2019guidelines,yildirim2023investigating}. These principles span the lifecycle of model development, including the initial conceptualization (e.g., what values are imbued in the system, how is privacy and fairness handled), the model development process (e.g., what data is used, how is the model trained), the model deployment (e.g., how are errors handled, how are users' preferences accounted for), and the interface layer where humans interact with the system (e.g., how are users' expectations calibrated, how transparent are the model's outputs)~\cite{wright2020comparative}. While these principles provide a strong foundation, human-centered LLMs require adapting and extending them to address the unique characteristics of large language models, accounting for their open-ended generation capabilities, evolving social and relational roles, and wide-scale deployment.

In this section, we explore specific challenges that arise when designing HCLLMs. While of course there are challenges beyond those covered in this subsection, our goal is to illustrate the differing levels of consideration we must attend to. These range from  considerations at the unit of the individual, such as scaffolding users' interactions with models and model outputs, to societal level questions when it comes to deploying models across diverse cultures and contexts. 

\subsubsection{Bridging the Gulf of Envisioning.}
Within HCI, foundational frameworks for design are \citet{norman1988design}'s gulfs of execution and evaluation. The gulf of execution pertains to gaps related to users figuring out how to do certain actions that they want, whereas the gulf of evaluation arises when the user is not able to interpret the system's output. These gulfs are persistent across the design of many technologies, ranging from everyday, physical objects like doorhandles to cutting edge technologies. However, LLMs pose a new challenge for users: the ``\textbf{gulf of envisioning}''~\cite{subramonyam2024bridging}. This gulf refers to the distance that emerges between what users may intend to do with an LLM and the prompts that are ultimately produced. 

Why does this gulf of envisioning occur? Although LLMs have proven capable across a wide-range of tasks, they still require users to guide how they are being used. The de facto form of user interaction with LLMs is prompting, or providing textual instructions delineating the desired interactions. While prompting seems an ostensibly simple task, users consistently struggle with writing prompts, underspecifying the instructions and foregoing the appropriate level of detail, ultimately yielding unsatisfactory model outputs~\cite{zamfirescu2023johnny}. Furthermore, users must contend with the indeterminacy of model outputs as well as the ``black-box'' nature of how their inputs are transformed into outputs~\cite{subramonyam2024bridging,yang2020humanai,agrawala2023unpredictable}. 

Thus, one core challenge for designing human-centered LLMs is overcoming this gulf of envisioning. One approach to do so has been focused on designing interfaces that can better scaffold users' interactions with models. For example, recent works have introduced direct manipulation interfaces as an alternative to prompt writing~\cite{masson2024directgpt,wu2022ai,arawjo2024chainforge}, such as providing a visual programming environment that helps users create more complex prompt chains~\cite{arawjo2024chainforge}. Others have operationalized prompting pipelines as modular code components to provide a more systematic process for optimizing and refining inputs~\cite{khattab2024dspy}. There are also approaches beyond intervening at the prompt level that are intended to improve users' interactions with LLMs. One line of work has explored how to improve the model's understanding of the user, such as through building better user models or improving context that LLMs have about the user~\cite{shaikh2025gum,naous2025flipping,lei2026humanllm}. These approaches aim to reduce this gap between users' intentions and what they specify when interacting with an LLM. The improving capabilities of models are only beneficial to users in so far as they can harness them. Efforts coming from both directions --- making it easier for users to guide model outputs with less effort and improving models' understanding of users --- are required to reduce this gulf of envisioning.

\subsubsection{Interpreting LLM Outputs.}
So far, our discussion of human-LLM interaction has mainly focused on how users communicate intent to models. We now turn our attention to how users evaluate and make sense of model outputs. As-is, model outputs can be verbose and unstructured, making it difficult for users to understand and leaving them feeling overwhelmed~\cite{jiang2023graphologue}. To address this challenge, prior work in human-computer interaction has proposed new interfaces to support user \emph{sensemaking} --- or providing external representations to encode data for task-specific purposes. For example, works such as Sensecape~\cite{suh2023sensecape} and Graphologue~\cite{jiang2023graphologue} provide interactive visualizations that allow users to visually explore LLMs' outputs in a structured format rather than having to parse large amounts of text.

Beyond helping users understand the content that LLMs output, we also must interrogate how users subsequently interpret and act upon these outputs. Questions surrounding user trust and reliance on AI systems are long-standing issues that predate the rise of LLMs~\cite{papenmeier2022s,vasconcelos2023explanations}. Nonetheless, LLMs introduce new challenges to these established problems. The complex nature of these models means that understanding why and how models produced the outputs is inscrutable to experts and end users alike~\cite{ameisen2025circuit}. However, user trust and reliance are not solely model-specific issues; the design of LLM applications also introduces new challenges. For example, \citet{swoopes2025impact} demonstrate how the chat-based design of most user-facing LLMs can hide the inherent stochasticity of the models, making it difficult for users to calibrate trust. Furthermore, anthropomorphic features of models can further foster user trust, even when it may be unwarranted~\cite{cohn2024believing}. To combat these issues, researchers have proposed design features that can foster appropriate reliance, such as generating explanations, expressing uncertainty, and adding sources to claims~\cite{zhou2023navigating,kim2025fostering}. These solutions are not perfect; for instance, significant technical challenges remain in ensuring that cited sources are correct and relevant. The hypothesized benefit of features like sourcing is their potential to engage users in slower, more careful thinking, but how to design LLMs that effectively empower users' critical thinking over their outputs remains a key open question.

\subsubsection{Navigating Human-LLM Relationships.}
Finally, we consider the evolving role of LLMs relative to users. Traditionally, AI systems have functioned as assistants: tools that can augment human capabilities in restricted ways and that require users to delegate tasks. However, as model capabilities approach or exceed human performance in certain domains, visions of models as equal collaborators are becoming increasingly plausible. For instance, \citet{shao2024collaborative} articulates a framework in which humans engage in bidirectional collaboration with LLM-based agents across a diverse set of tasks. Shifting roles from assistant to collaborator carries significant implications for the design of LLMs. For example, while assistants wait for explicit instructions to execute tasks, a model that serves as a collaborator might proactively suggest alternative approaches, challenge assumptions, or redirect problem-solving strategies, demanding fundamentally different interface affordances. These questions echo and build upon classic debates in HCI between direct manipulation interfaces, which afford users high degrees of control, and interface agents, which can act autonomously on users' behalf~\cite{shneiderman1997direct}. Determining how much control is required is not prescriptive but rather will be modulated by contextual factors such as the type of task or user expertise.

Furthermore, as LLMs adopt more expansive roles beyond serving as functional tools in our lives, we must also consider the \emph{affective} dimension in human-LLM interaction. Models are increasingly anthropomorphized, meaning that they are perceived as having human-like characteristics --- a fact that is exacerbated by the linguistic expressions in generated outputs~\cite{cheng_anthroscore_2024,cheng-etal-2025-dehumanizing}. Already, users are interacting with these models not only as coworkers or collaborators but also as friends, companions, and romantic partners~\cite{zhang2025rise,pataranutaporn2025my}. LLMs are also increasingly used in sensitive domains, such as providing emotional support or being used for therapy purposes~\cite{zaosanders2025how}. These affective interactions are also not always intentional. For example, \citet{zhang2025rise} observed that companionship-oriented interactions emerge even when users are not primarily interacting with LLMs for emotionally laden tasks. While human-LLM relationships can have positive outcomes for users, including combating loneliness and reducing distress~\cite{de2025ai}, there are also many documented adverse impacts, such as fostering emotional dependence~\cite{laestadius2024too,pentina2023exploring}, encouraging harmful behavior~\cite{zhang2025dark,dupre2024aichatbots}, or, in the extreme, triggering cases of intense mental health crises~\cite{morrin2025delusions}. Balancing the trade-offs between these benefits that human-LLM relationships can have with these very real harms is an open challenge that necessitates interdisciplinary interventions from technologists, social scientists, ethicists, and policymakers. As an example of a work in this area, \citet{kirk2025human} called for the community to prioritize the socio-affective alignment of LLMs, accounting for how models fit into and actively shape individual users' social and psychological ecosystems. Nonetheless, how exactly we design human-LLM relationships that promote user well-being in the long-term or other pro-social outcomes remains an open area for exploration. 

\subsubsection{Designing for Diverse Cultures and Contexts.} 
Finally, we turn our attention to the critical challenge of designing LLMs that are sensitive and adaptive to diverse cultural contexts. LLMs are not culturally neutral artifacts. As discussed in more detail in \S\ref{sec:data}, they are trained predominantly on data from Western, Educated, Industrialized, Rich, and Democratic (WEIRD) societies~\cite{mihalcea2025ai}. As such, models inherently encode and propagate specific cultural values, communication styles, and social norms~\cite{naous2024havingbeerprayermeasuring,durmus2023towards,ryan2024unintendedimpactsllmalignment}. These misalignments can render models unhelpful, at best, and culturally insensitive or actively harmful to users.

HCI offers critical theoretical lenses to help us not only question the assumptions underlying LLM design but also offer more generative opportunities for design. For example, \citet{bardzell2010feminist}'s foundational work on Feminist HCI argues that we ought to be attending to marginalized user groups when designing rather than focusing only on a presumed ``default''. In this vein, other critical theories on postcolonial computing and literature on decolonial practices have urged designers to decenter dominant Western perspectives and account for the plurality of worldviews and epistemologies that exist~\cite{irani2010postcolonial,alvarado2021decolonial}. These works push designers to move beyond simply ``de-biasing'' models and instead question the fundamental assumptions embedded within them: whose knowledge is centered, whose values are prioritized, and whose ways of being are marginalized? For example, users from different cultural backgrounds may have different expectations from what they wanted out of an ideal AI system and use cases, requiring that we are able to localize models to these needs rather than assuming a single default~\cite{ge2024culture,qadri2025case,phutane2025disability}. Systematically prioritizing knowledge from certain groups or cultures over others is not simply a design challenge for better human-LLM interaction but can present quality-of-service differences with material impacts on users~\cite{wilson2025no,dev2022measures}. How to design, build, and evaluate LLMs that are genuinely context-aware and culturally adaptive remains a significant and vital open question for the field.
\subsection{Expanding the \emph{How}: Methods from HCI for HCLLMs}
\label{subsec:hci_methods}
Finally, in addition to providing new perspectives on how we ought to design HCLLMs, HCI also introduces a set of methods that researchers and practitioners can employ in pursuit of these goals. Given HCI's interdisciplinary roots, there are many ``ways of knowing'' or methodological orientations that researchers employ within the field. In this section, we will discuss the applicability of three classes of research methods --- experimental methods, participatory research, and grounded theory --- that are particularly applicable to HCLLMs and are less utilized within NLP. For a comprehensive overview of other methodological practices in HCI, we refer readers to \citet{olson2014ways}.   

\subsubsection{Experimental Research}
An important step in designing and evaluating HCLLMs requires measuring their impact on human outcomes, rather than only characterizing behavior. While traditional methods in NLP, such as benchmarking, provide a fast and scalable way to evaluate models, these numbers are often divorced from the context that LLMs are applied and do not capture their impact in practice~\cite{raji2021ai,mcintosh2024inadequacies}. Unlike observational studies, which are useful in informing us whether variables are related, experimental research helps reveal many types of relationships (e.g., association, causality) between variables of interest~\cite{gergle2014experimental}. Experiments can vary along the spectrum of research control, ranging from laboratory experiments to field experiments that are conducted in real-world settings but often have many factors that researchers cannot control for~\cite{oulasvirta2008field}. 

For HCLLMs, experimental methods enable researchers to move beyond measuring what models can do to understanding what they \emph{actually} do and \emph{why}. First, experimental methods also allow researchers to better understand mechanisms that explain observed patterns of behavior. For example, to better understand why users become dependent or overreliant on LLMs, recent work~\cite{cheng2025sycophantic} focused on the construct of \emph{social sycophancy}, finding through a series of laboratory experiments, that users are more likely to trust and also use sycophantic models. Isolating specific mechanisms can then inform design interventions that are first validated intrinsically (i.e., benchmarks) and then extrinsically with additional experiments. Second, experimental methods can quantify the utility of LLMs when deployed in practice, allowing researchers to evaluate outcomes rather than capability. A number of field experiments have started, looking at LLMs' impacts across domains including education~\cite{cuna2025hidden,wang2025tutorcopilothumanaiapproach}, software engineering~\cite{becker2025measuring}, healthcare~\cite{korom2025ai}, and so on. For example, \citet{becker2025measuring}'s work provides evidence that challenges the assumption that LLMs increase developer productivity, demonstrating through a field experiment that using these tools increases the time it takes for developers to complete their tasks. These findings complicate our understanding of models and point to the exogenous factors (e.g., institutional endorsement, workflow integration, user onboarding) that affect outcomes. Overall, moving from what a model can perform within a clearly scoped benchmark setting to how it impacts users or society writ large --- and why --- is a critical contribution from HCI's methodological toolkit that can complement traditional NLP evaluations.

\subsubsection{Participatory Approaches}
\label{subsub:participatory}
Building HCLLMs requires us to engage with the communities that are impacted by and using these technologies. A useful approach for understanding how we can work toward this engagement is to draw on participatory research methodologies. More than being a prescriptive methodology to follow, ``participatory research'' delineates an orientation towards involving relevant stakeholders in the knowledge creation process~\cite{bergold2012participatory}. Examples of methods that fall under this broader umbrella of ``participatory research'' include participatory action research~\cite{hayes2011relationship} and community-based participatory research~\cite{unertl2016integrating,wallerstein2017theoretical}. For a more extensive list of participatory research frameworks, we refer the reader to \citet{vaughn2020participatory}. Despite the many variants that fall under participatory research, a unifying emphasis is on fostering a democratic and inclusive process that treats stakeholders or community members as equal research collaborators rather than subjects~\cite{vaughn2020participatory,bergold2012participatory}. 

Participatory approaches have been adopted in creating machine learning solutions, such as developing context-specific models for detecting feminicide~\cite{suresh2022towards} or providing frameworks for communities to articulate algorithmic policies~\cite{lee2019webuildai}. \citet{tseng2025ownership} provides an example of how participatory research approaches can be applied to designing HCLLMs. Through interviews with different stakeholders in the journalism ecosystem (e.g., reporters, editors, executives), \citet{tseng2025ownership} articulate how LLMs must be designed to address journalists' needs, such as using an open-sourced model that can be fine-tuned for their tasks rather than relying on commercial offerings. 

While we have listed exemplars of work that have adopted participatory approaches, it is important to state that participation is neither a panacea to the many of the challenges with creating HCLLM nor is it easy to institute these methods in practice. These methods require building trust with communities, which can be a long-term process requiring significant time investment~\cite{le2015strangers}. To ensure that participatory approaches are equitable in practice, it further requires centering communities, especially those who do not hold positions of power, and ensuring that research tangibly benefits these individuals rather than acting as an extractive force~\cite{harrington2019deconstructing}. 

\subsubsection{Qualitative Inquiry}
Qualitative methods encompass a broad range of approaches, including ethnographic studies, interviews, participatory design workshops, thematic analysis, and other interpretive techniques~\cite{blandford2016qualitative}. For HCLLMs, existing work has also sought to extract insights by analyzing users' chat histories with models, such as common interaction patterns or high-level values reflected in model responses~\cite{tamkin2024clio,huang2025values}. While quantitative measures reveal how well models perform or what users do, qualitative approaches provide a deeper understanding of why and how individuals think about, interact with, and make sense of these systems. As \citet{geertz2008thick} illustrates through the notion of ``thick description,'' qualitative methods enable researchers to capture the meanings, contexts, and social dynamics that underlie behavior, rather than merely documenting surface-level patterns.

Qualitative methods offer several particularly valuable contributions to HCLLMs. First, they can uncover needs and harms in underrepresented populations, revealing how specific groups engage with LLMs and the nuanced benefits and harms they experience—insights that aggregate metrics often obscure (e.g., \citet{ma2024evaluating}'s interviews with LGBTQ+ individuals exploring LLMs in mental health support; \citet{qadri2025case}'s workshops with South Asian participants examining cultural misrepresentations in AI). Qualitative methods also help generate new design insights and hypotheses by surfacing unexpected use patterns, workarounds, and unmet needs that can inform future system design. Through providing this rich description about how users interact with these models, qualitative work enables contextual understanding of how LLM interactions are embedded in broader social and work practices, revealing dependencies and consequences that otherwise may be missed. For example, \citet{tamkin2024clio}'s analysis of user interactions with Claude revealed areas of unsafe behavior that their current guardrails were not designed to catch, allowing them to refine their system design through empirical insights. Finally, qualitative approaches complement and enrich quantitative findings by helping triangulate results and providing interpretive depth that explains why certain patterns emerge~\cite{zhao2025sphere}. 
\subsection{From Human-Centered Design Challenges to Technical Artifacts}
\label{subsec:technical_artifacts}

Addressing these design challenges for HCLLMs requires coordinated interventions and across the entire LLM development pipeline --- from the data curation and model training processes to the user interface design. Already when introducing these challenges, we have discussed interface-level interventions. For example, at the model level, improving model capabilities via post-training, such as instruction tuning and preference learning (as discussed in \S\ref{sec:nlp}), can help models better interpret underspecified prompts, helping bridge the gulf of envisioning. Decisions that developers make around what sources to include in the training data will affect models' capabilities across different cultures and context (\S\ref{sec:data}). Beyond the model training pipeline, choices around the technical design of the system, such as how to handle source attribution or what safety guardrails are put in place (\S\ref{sec:responsible}), will mold users' interactions, impacting relationships that may form. Nonetheless, the critical point of this chapter is that these technical interventions are most effective when informed by and evaluated against the human-centered principles outlined above. The challenges we have charted are fundamentally \emph{sociotechnical} problems; they cannot be solved by better models alone, nor by better interfaces alone, but through the careful co-design of both.

\subsubsection{Case Study: Motivating Physical Activity with HCLLMs}
\label{subsec:hci_case}

To make the human-centered LLM design process concrete, we present a case study based on two related systems for LLM physical activity coaching: (1)~GPTCoach~\cite{joerke2025gptcoach}, an LLM-based coach that implements motivational interviewing, and  (2)~Bloom~\cite{joerke2026bloom}, a mobile application that integrates GPTCoach with established, UI-based behavior change interactions. Physical inactivity is a major public health concern, with large portions of the population falling short of recommended guidelines for physical activity. LLMs present a promising opportunity to combine the scalability of existing mobile health interventions with the personalization of human coaching. Through this case study of designing an LLM health coach, we illustrate how  a \emph{human-centered} process can help realize this opportunity. 

Let us start by considering the status quo approach that treats training a good LLM health coach as an instruction-following problem. The approach is to collect user data (e.g., common barriers to activity, wearable data), feed it to a model, and have the model generate personalized nudges, exercise plans, or advice. This framing is an intuitive starting point, and it maps cleanly onto standard LLM training pipelines. However, it also implicitly encodes a set of assumptions: that users always want or need recommendations and advice, that more information yields better outcomes, and that the primary bottleneck is the model's ability to produce accurate health advice. 

GPTCoach and Bloom instead took a human-centered approach, exemplifying the following three concepts discussed in the chapter:
\begin{itemize}

\item \textbf{Working with stakeholders to shape the system.}
Rather than starting from the status quo, \citet{joerke2025gptcoach} conducted formative interviews with health experts and prospective end-users. These interviews revealed that health experts emphasized the importance of facilitative, non-prescriptive support that refrains from giving unsolicited advice---a mode of engagement that runs counter to how a standard LLM chatbot operates. Experts described their role as staying ``in the passenger seat'' and helping clients surface their own goals and barriers, rather than telling them what to do. This learning fundamentally reframed the problem to be solved, the system that was designed to address it, and the evaluation criteria. Moreover, engagement with experts did not end after the formative study. After the lab study, the authors hired trained experts to code all transcripts to measure adherence to motivational interviewing. Expert coding indicated that GPTCoach used conversational strategies that were consistent with motivational interviewing or neutral over 93\% of the time, but qualitative feedback revealed important gaps compared to skilled human practitioners, highlighting specific areas for improvement that would not have surfaced without expert involvement.
\item \textbf{Focusing on the interaction.}
A second theme from this case study is that model capability, while important, is not sufficient in and of itself for a successful human-centered system. While the authors could have devoted significant efforts to training, prompt chaining proved sufficient to enable LLM-based motivational interviewing. This created space for the authors to focus on how different aspects of the interaction design could shape users' experiences in substantive ways. For example, GPTCoach's non-prescriptive and non-judgmental communication style had a greater impact on participants' overall experience than its analysis of their health data. In Bloom, the authors represented the coaching agent as a \textit{bee} avatar named Beebo. Beebo's capabilities are the same as those of a generic chatbot, but it’s representation substantially changed the nature of the interaction. Many participants resonated with the avatar and described Beebo in relational terms, leading to increased engagement and adherence. Beebo's clear role as a ``coach,'' not a general purpose assistant, helped set expectations when Beebo redirected conversations back towards physical activity, or when guardrails triggered refusals for medical advice. This dynamic points to the design challenge of navigating human-LLM relationships from Sec.~\ref{subsec:hci_challenges}. Taken together, these design choices reflect a move beyond thinking narrowly about model performance toward a more holistic understanding of how users will interact with these systems.
\item \textbf{Evaluating with users.}
Finally, this case study showcases how the different methods from HCI offer new ways of knowing for evaluating HCLLMs. In a four-week randomized field study (N=54) comparing Bloom to a no-LLM control, the authors used a mixed methods evaluation, synthesizing insights across qualitative coding of participant interviews,  survey data, app usage logs, and wearable data. The quantitative data revealed a 5x increase in overall app usage time in the LLM condition, while mean physical activity levels stayed comparable in both conditions. Meanwhile, survey measures revealed substantial shifts in physical activity mindsets and satisfaction. Qualitative coding added rich nuance to these findings, with participants in the LLM condition reporting stronger beliefs that activity was beneficial to their health, greater enjoyment of exercise, an expanded appreciation of ``what counts'' as activity, and increased self-compassion when goals were missed. 
Most importantly, participants attributed these mindset shifts to interactions with Beebo that kept them in control of their own behavior change, such as finding flexible alternatives when plans fell through.
More broadly, this illustrates how qualitative ``thick'' understanding can surface the \textit{why} behind user experiences, which are insights that a purely quantitative evaluation might miss.
\end{itemize}

Overall, this case study demonstrates how critically engaging with humans \textit{early} in the design process can reframe the problem being solved and the evaluation target, leading to qualitatively different solutions that better serve human needs. Notably, the most consequential outcomes in both studies---positive changes in participants' beliefs about physical activity and their own abilities---emerged from the design process rather than from improvements in model capability.
\clearpage

\clearpage
\hypertarget{data}{\section{Data for HCLLMs}}
\label{sec:data}

Data is central to the language modeling paradigm, just as it has been throughout the history of machine learning \citep{halevy2009unreasonable}. Every stage of language model development, from pretraining to evaluation and deployment, depends on the availability of massive text corpora \citep{sun2017revisiting}. Critically, the scale, diversity, and quality of this linguistic data can determine a model’s downstream utility for users \citep{kaplan2020scaling,liu2024makes,zhou2023lima}. Data quality and quantity can quickly become bottlenecks \citep{10.5555/3692070.3694094}, limiting progress in AI \citep{longpre2024consent}. Thus one of the most urgent challenges in AI is in identifying diverse and representative sources of data.

\begin{figure}
    \centering
    \includegraphics[width=\linewidth]{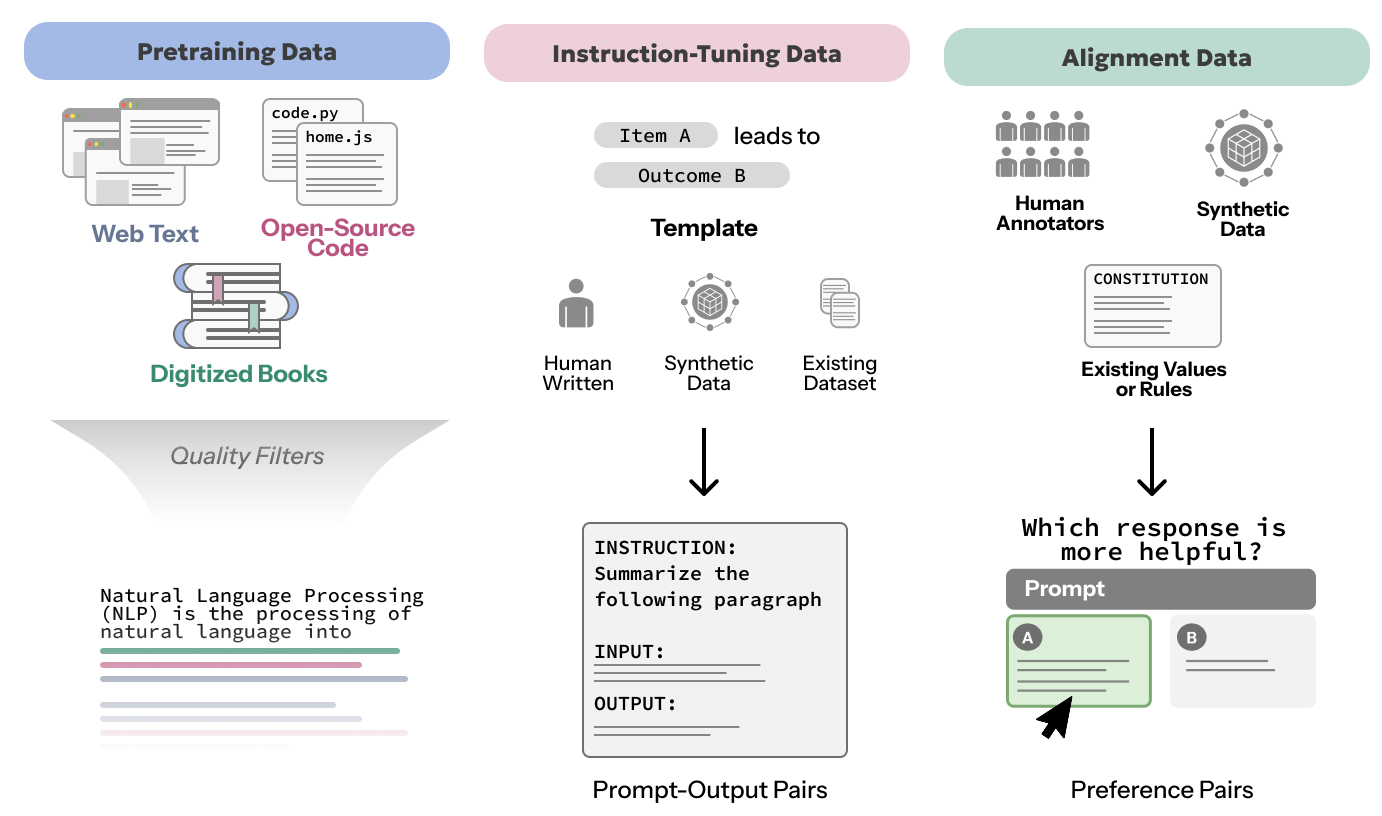}
    \caption{This chapter focuses on the human origins of data (\S\ref{subsec:data_provenance}), and how data encodes perspectives and values that impact HCLLM outcomes, from representation and bias (\S\ref{subsec:data_representation}) to consent and ownership (\S\ref{subsec:data_privacy}). In particular, we consider \textbf{\textit{pre-training data}}, \textbf{\textit{instruction-tuning data}}, and \textit{\textbf{alignment data}}.}
    \label{fig:data}
\end{figure}

From the human perspective, data is more than the fuel behind AI progress. Data is a dynamic reflection of lived human experience. It reflects the people, institutions, cultures, histories, and social contexts that produce it. In this sense, data is never neutral. Rather, data encodes viewpoints and values \citep{dotan2020value}, assumptions and biases \citep{paullada2021data}, and even political and social structures \citep{scheuerman2021datasets,miceli2022studying,capel2023human}. The origins of such data may be the subject of legal claims and privacy concerns, and its content may be highly personal or sensitive \citep{bender2018data}. To understand the human impact of LLMs, it becomes necessary to consider the human origins of data that shapes our models, particularly in pre-training, instruction tuning, and alignment (Figure~\ref{fig:data}).

In the first subsection of this chapter, we examine the \textbf{(\S\ref{subsec:data_provenance}) \textit{provenance}} of data used to develop LLMs. We ask where this data comes from, who produced it, under what conditions it was produced, and how it was transformed throughout this process. In this way, we recognize how data encodes implicit values, perspectives, and cultures that shape LLM behavior. From here, we are positioned to understand human-centered concerns around \textbf{(\S\ref{subsec:data_representation}) \textit{representation and bias}}, how the data's origins systematically skews, misrepresents, and erases the perspectives of underrepresented groups, leading to representational and allocational harms. While rich community and personal data may be used to mitigate some of these harms, we consider issues around \textbf{(\S\ref{subsec:data_privacy}) \textit{consent and ownership}}. Finally, we consider some of the biggest data challenges facing LLM developers today, and how proposed solutions like \textbf{(\S\ref{subsec:synthetic_data}) \textit{synthetic data}} account or fail to account for the human-centered objectives we have outlined.

\hypertarget{representation}{\subsection{Data Provenance}}
\label{subsec:data_provenance}

\textbf{Data provenance} \citep{longpredata}, also called \textbf{dataset genealogy} \citep{denton2021genealogy}, is the record of a dataset's origins, history, and transformations throughout its lifecycle. An understanding of data provenance is critical for achieving transparency in LLM development \citep{bommasani2023transparency}. Without transparent data practices, it becomes difficult to predict and understand why LLMs leak private information \citep{19_kandpal2022deduplicating,bubeck2023sparksartificialgeneralintelligence}, violate copyrights \citep{carlini2021extracting}, or perpetuate social biases \citep{denton2021genealogy}. But with data provenance, stakeholders become more equipped to audit models \citep{mokander2024auditing} and tackle these human-centered concerns. In this section, we will investigate the provenance of data used for pre-training, instruction-tuning, and aligning LLMs. In particular, we ask where data comes from, who produced it, and under what
conditions it was produced.

{\subsubsection{Pretraining Data}}
\label{subsec:data_provenance_pretraining}
\textbf{Data Sources.} The provenance of LLM pretraining data is often complex, layered, and opaque. Unlike the small, curated datasets of traditional machine learning, LLM pretraining corpora tend to be huge, multi-trillion token aggregations across heterogeneous and potentially noisy sources. These sources traditionally include web text, digitized books, and open-source code repositories \citep{wenzek2019ccnetextractinghighquality,raffel2020exploring, soldaini2024dolmaopencorpustrillion,devlin-etal-2019-bert}. Although different model developers use different data mixtures, most incorporate an open web crawl that at least partially intersects the \href{https://commoncrawl.org/}{\textit{Common Crawl}}. The Common Crawl contains monthly snapshots of ``open web'' --- partial samples of machine-crawlable sites reached from seed URLs that were initially crowdsourced in 2008 \citep{baack2024critical}. This is not a random sample of the internet. Large web crawls like this favor wikis, news sites, blogs, and other user-generated content platforms, which are generally multilingual, but heavily skew towards English \citep{baack2024critical}. Much of this data has been found to be socially undesirable, with a high prevalence of hate speech and sexually explicit content \citep{luccioni2021whatsboxpreliminaryanalysis}. The data can also reify social, cultural, and political biases \citep{naous2024havingbeerprayermeasuring,feng2023pretrainingdatalanguagemodels,navigli2023biases}. Finally, this web-scale data inextricably encodes the structural biases of the web itself, where the majority of content is produced by an active minority of users, and these users overly represent Western, Educated, Industrialized, Rich and Democratic (WEIRD) populations \citep{baeza2018bias}.

\textbf{Quality Filtering.} Because open web data is noisy, redundant, low-quality, and often socially undesirable, model developers use classifiers or heuristics \citep{chen2021evaluatinglargelanguagemodels,penedo2023refinedweb,rae2021scaling,soldaini2024dolmaopencorpustrillion} to filter their pre-training corpora for unique and high-quality documents that are information dense and free from toxicity or personally identifiable information \citep{longpre2024responsible}. By filtering pre-training data in this way, researchers can train safer models with better performance at lower computational costs \citep{du2022glamefficientscalinglanguage,rae2021scaling,albalak2024dataselection}. The distributions of these filtered corpora are shaped by sampling decisions, including the filters used to determine document quality.  These quality filters often systematically exclude both communities and discursive topics \citep{lucy-etal-2024-aboutme}. For instance, toxicity classifiers often exhibit racial and linguistic bias \citep{sap2019risk,dodge-etal-2021-documenting}. Quality filters trained on Wikipedia and OpenWebText tend to favor text from wealthy, educated, urban areas \cite{gururangan-etal-2022-whose}.

After language identification \citep{conneau2019cross,laurenccon2022bigscience} and deduplication \citep{lee2022deduplicatingtrainingdatamakes}, model-based filtering is one popular quality filtering approach. For example, perplexity-based methods filter noisy documents that appear as highly surprising to a much smaller language model. CCNet \citep{wenzek2019ccnetextractinghighquality} was constructed as a subset of the Common Crawl, filtered with 5-gram language models that the authors trained on Wikipedia data for each target language. They use a fastText classifier \citep{joulin2017bag} for language identification and run each deduplicated document through the appropriate 5-gram model to compute perplexity, filtering based on a heuristic and language-specific perplexity threshold. Disconcertingly, this pipeline effectively filters out minority dialects and low-resource languages \citep{albalak2024dataselection} for which language ID is unreliable \citep{caswell-etal-2020-language,kudugunta2023madlad}, or the perplexity model is overfit to only a small corpus \citep{feng2023pretrainingdatalanguagemodels,lucy-etal-2024-aboutme}. Perplexity-based filtering will retain text that matches the language distribution the filtering model was fit to. When the standard is Wikipedia, filtering will primarily preserve Standard American English in the third person, written in a neutral, semi-formal and broadly readable register, with clear declarative sentences. Another idea is to prompt existing LLMs to estimate the quality of pre-training data zero-shot using some manually-written definition of high quality data \citep{sachdeva2024train,wettig2024qurating,penedo2024fineweb}. This was the approach used for Llama-3 \citep{llama3modelcard}. However, manually-written definitions are brittle and may not encompass task-specific or user-specific notions of LLM utility \citep{held2025optimizing}. A third idea is to fine-tune a small model like fastText \citep{joulin2017bag} as a binary quality classifier. The binary classifier was the approach used by DataComp for Language Models \citep[DCLM;][]{li2024datacomp}, as this resulted in models with higher scores on general benchmarks like MMLU \citep{hendrycksMMLU2021}. However, methods like this are prone to overfitting on the training set, the construction of which itself reflects and reifies the values of model developers.  

Another popular filtering mechanism is to use content heuristics like domain name blacklists, and toxic keyword dictionaries, which were used to construct the Colossal Clean Crawled Corpus (C4)  \cite{xue2021mt5,raffel2020exploring}. The English C4 was found to skew heavily towards Wikipedia articles, patents, and United States news articles, such as the New York Times \citep{dodge-etal-2021-documenting,elazar2024s}. Most documents in the corpus had been published after the year 2011. The multilingual variant, mC4 \citep{xue2021mt5}, represents 101 identified languages, but many of these languages are under-represented \citep{snaebjarnarson2022warm}. For example, compared to 2.7T tokens of English, mC4 contains only 600,000 tokens of Javanese \citep{aji2022one}. Data for these lower-resource languages is also much noisier than that of the English subset \citep{kreutzer2022quality,van2024language}. 

Many pretraining corpora aggregate documents from a variety of sources. Popular examples include the Pile \citep{gao2020pile800gbdatasetdiverse}, RedPajama \citep{weber2024redpajama}, and Nemotron-CC \citep{su2025nemotron}, which contain 300B, 1T, and 7T tokens respectively, sampled from the Common Crawl, as well as academic texts, books, coding, medical, and legal documents \cite{biderman2022datasheet,weber2024redpajama}. Over half of the documents in these data mixes are duplicated at least once, and some of these contain personally identifiable information like email and IP addresses, as well as toxic language \citep{elazar2024s}. 

{\subsubsection{Instruction-tuning Data}}
\label{subsec:data_provenance_instruct}

Compared to the origin story of pre-training data, the provenance of instruction-tuning datasets is relatively well-known. Data is typically aggregated by a single organization for the purpose of fine-tuning a model to follow instructions. Therefore the provenance of instruction dataset development resembles the distribution of model developers, over half of whom originate in the US or China \citep{held2023material}. Notable examples include Google's FLAN \citep{chung2024scaling}, AI2's Natural Instructions \citep{mishra2021crossnaturallanguage}, Stanford's Alpaca \citep{alpaca}, and Cohere's Aya corpus \citep{singh2024aya}.

The instruction aggregation process often involves selecting a diverse range of tasks over which are constructed well-formatted prompt-output pairs. Some organizations may opt to annotate these pairs entirely from scratch, like in the \citet{databricks2023dolly15k} Dolly-15k. There are ethical and scientific benefits in such cases where data is sourced with explicit consent, attribution, and compensation. However, this is not the norm, especially since doing so demands significant human labor.

In many cases, instructions are sourced automatically from evaluation benchmarks via templates \citep{chung2024scaling,longpre2023flan}, which may be further translated \citep{muennighoff2022crosslingualBLOOMZ} or restructured using tertiary models. The templates themselves typically have a human origin. For example, the Natural Instructions dataset \citep{mishra2021crossnaturallanguage} was sourced from annotation guidelines that the benchmark developers constructed to onboard crowdworkers. Sometimes, humans also write templates from scratch, especially in the early days of UnifiedQA \citep{khashabi2020unifiedqa} and FLAN \citep{wei2021finetuned}, and in low-resource settings like the multi-lingual Aya corpus \citep{singh2024aya}. However, much of the data construction pipeline is automated. This trend is growing as instruction-tuning datasets are generated synthetically. For example, the instructions used to fine-tune Stanford's Alpaca model \citep{alpaca} were distilled from GPT-3.5, a larger model which was itself instruction-fine-tuned. This approach, called self-instruction tuning \citep{wang2023selfinstructaligninglanguagemodels}, has been adopted in a range of more recent work \citep{peng2023instructiontuninggpt4,li2023ottermultimodalmodelincontext}. As we will discuss in \S\ref{subsec:synthetic_data}, the use of synthetic data for self-instruction tuning complicates data provenance, and may exacerbate the human-centered concerns raised in this chapter. 

{\subsubsection{Alignment Data}}
\label{subsec:data_provenance_posttraining}
Additional datasets are used for \textit{model alignment}, or the process of training more helpful and less harmful models via supervised fine-tuning, preference tuning, and reinforcement learning from human feedback (RLHF) \citep{askell2021generallanguageassistantlaboratory}. Since alignment data is what produces models that are useful to humans, it constitutes a major force behind the sudden proliferation in the number of LLM users worldwide.

Since the notion of helpfulness or harmfulness is ambiguous and varies with different cultures and contexts, one might expect a commensurate heterogeneity in both the source and format of alignment data \citep{ethayarajh2024behavior}. This is generally not the case. With respect to the format, many alignment datasets assume a Bradley–Terry model of pairwise human preferences. Datasets like Anthropic's HH-RLHF \citep{bai2022traininghelpfulharmlessassistant}, OpenAI's InstructGPT \citep{ouyang2022training}, and Peking University's PKU-SafeRLHF \citep{NEURIPS2023_4dbb61cb} couple a user prompt with a pair of model responses: one preferred and one dispreferred. With respect to data sources, many preference judgments come from a very small pool of annotators, sometimes within the organization itself. For example, Peking University hired 28 internal annotators to construct PKU-SafeRLHF, and Anthropic's internal research team similarly hired and trained a small group of contractors to construct HH-RLHF. 

Crowdsourcing and citizen science can serve to democratize the process of collecting alignment data. One drawback of these approaches is  sampling bias, which may favor researchers, AI enthusiasts, and individuals from industrialized nations. Chatbot Arena \cite{DBLP:conf/icml/ChiangZ0ALLZ0JG24}, also known as LMArena, is one example of a public web platform with open-user participation in which volunteers engage with pairs of anonymous models and provide preference feedback in the standard binary format. The project was initiated at the University of California Berkeley in 2023, and covers 96 languages, although the vast majority are in English. OpenAssistant Conversations \citep{kopf2023openassistant} is a similar crowdsourcing effort, initiated by the German non-profit LAION in 2022. Over 13k volunteers contributed alignment data in 35 different languages, particularly in English (50\%), German (20\%), and Spanish (10\%). Of these annotators, 89.1\% identified as male, with a median age of 26. These clear demographic biases above will skew the values, perspectives, and interests represented by this data.

To address issues of demographic bias, some dataset developers intentionally target underrepresented demographics in their recruitment efforts. For example, the PRISM Alignment Dataset \cite{kirk2024the} is an academic project initiated at the University of Oxford, where the developers recruited Prolific workers from 33 underrepresented countries. The Meta Community Alignment Dataset \citep{zhang2025cultivating} is a similarly-motivated multilingual preference dataset in which its 15k participants were recruited from five countries on YouGov. Still, there remain limitations in recruiting diverse populations from crowdwork platforms, which have limited global coverage \citep{rinderknecht2025daily,douglasDataQuality,palan2018prolific}.

Synthetic data is an emerging trend among subsections in this chapter, and it is largely motivated by the need to scale AI beyond what human annotation labor can support \citep{casper2023openproblemsfundamentallimitations,santurkar2023opinionslanguagemodelsreflect}. Some LLM developers have considered synthetic data in the alignment step as well. Variants of this approach include Constitutional AI \citep{Bai2022ConstitutionalAH} and Reinforcement Learning from AI Feedback \citep[RLAIF;][]{lee2024rlaif}. Both approaches shift critical alignment decisions from data contributors to more centralized authorities: namely, the LLM-as-a-Judge, and those who prompt it. For example, in Constitutional AI, models judge their own output against the standards of a human-written constitution, and then re-write a better, constitution-aligned response. Anthropic's original 2022 constitution was sourced from Western liberal-democratic sources like the United Nations Declaration of Human Rights, the OECD, and Google's AI Principles. These frameworks employ individualist, rights-based moral reasoning \citep{haidt2012righteous}, which may not represent other global ethical traditions, or incorporate the voices of pluralistic user bases \citep{sorensen2024roadmap}.
\hypertarget{representation}{\subsection{Data Representation, Bias and Ethics}}
\label{subsec:data_representation}

The story of LLM training data is a story about whose voices become computationally legible and whose are overwritten or erased. In \S\ref{subsec:data_provenance}, we considered how the story of data is shaped by its sources, filtering decisions, annotation pipelines, and synthetic generation practices. Imbalances or biases in the provenance of LLM pre- and post-training data can exacerbate representational, allocational, and quality-of-service harms for those who use these models. Representational harms include stereotyping, denigration, and misrecognition, when LLMs perpetuate and amplify distorted and harmful portrayals of personal identities and social groups \citep{Blodgett2021SociolinguisticNLP}. Allocational harms arise when LLMs reinforce or amplify inequality in the distribution of opportunities and resources \citep{barocas2017problem,eubanks2018automating}. Quality-of-service harms involve performance disparities across different user groups, which may cascade into both representational and allocational harms. For further discussion on how to define and measure these harms, see \S\ref{subsec:bias_eval}.

Sociotechnical harms become harder to diagnose when data provenance is incomplete. Without visibility into the linguistic, cultural, and geographic origins of the data, as well as the filtering and curation pipelines, researchers cannot identify: (1) why the model stereotypes certain voices, (2) why specific groups are absent from generated outputs, or (3) how certain narrative tropes became dominant. We will briefly discuss the relationship between data provenance and each of these harm outcome categories,  as well as data-based mitigation strategies.

\subsubsection{Quality-of-Service Harms}
Quality-of-service harms are disparities in model utility for users from different sociodemographic groups \citep{shelby2023sociotechnical}. These disparities are often rooted in the composition and curation of data as discussed in \S\ref{subsec:data_provenance}. Pre-training data scraping, quality filtering, instruction-tuning templates, and alignment data collection tend to over-represent native English speakers from wealthy, Western nations and under-represent the language and perspectives of marginalized communities. As a result, LLMs introduce quality-of-service harms for individuals from these communities \citep{shah2020predictive}. 

LLM performance degrades for speakers of non-standard language varieties or dialects on a wide range of tasks \citep{kantharuban2023quantifying,joshi2025natural}, from text classification \citep{lwowski2021risk} and machine translation \citep{ahia2023all} to question answering \citep{ziems2023multi,fleisig2024linguistic} and conversational AI \citep{artemova2024exploring}. This inequitable distribution of utility to LLM users can result in allocational harms like inequitable wages and quality of life, especially as LLMs are integrated into the workplace \citep{shao2025future} and become \textit{general purpose technologies} \citep{eloundou2024gpts}. Quality-of-service bias also contributes to the representational harm of erasure, and may derive in part from representational biases. 

\subsubsection{Representational Harms}

LLMs demonstrate representational harms when they propagate negative or skewed representations of social groups, including cultural misrepresentation, stereotypes, essentialist language, and erasure \citep{chien2024beyond}. Representational harms can derive from pre-training data \citep{10.1145/3682112.3682117}, not only from its explicitly harmful, stereotypical, and toxic language \citep{luccioni2021whatsboxpreliminaryanalysis}, but also from implicitly biased language \citep{Caliskan2016SemanticsDA,navigli2023biases}, framing effects \citep{feng2023pretrainingdatalanguagemodels}, and the sparsity of socioculturally representative data \citep{naous2024havingbeerprayermeasuring}. Data quality filters exacerbate racial and linguistic biases that skew pre-training data away from in-group perspectives in favor of unrepresentative and misinformed out-group perspectives \citep{wang2025large}. Post-training data can further induce mode-collapse, effectively flattening their representational distributions to portray groups one-dimensionally \citep{bisbee2024synthetic,durmus2024towards,rottger2024political}. This kind of distributional flattening is a form of \textit{essentializing} that is particularly harmful for groups historically portrayed as one-dimensional \citep{wang2025large}.

Unsurprisingly, LLMs are known to generate harmful stereotypes in question-answering \citep{naous2024havingbeerprayermeasuring}, machine translation \citep{ghosh2023chatgpt}, and open-ended generation \citep{dhamala2021bold}. These issues are only exacerbated when the prompts are written in non-standard dialects, which may trigger demeaning or condescending responses from models \citep{fleisig2024linguistic}. LLM-simulated personas also collapse into stereotypical caricatures \citep{cheng-etal-2023-marked,cheng-etal-2023-compost,gupta2023calm}. These simulations systematically misrepresent, flatten, and essentialize the perspectives of underrepresented groups based on protected characteristics like age, gender, and disability \citep{wang2025large}. 

\subsubsection{Allocational harms}
Allocational harms are disparities in individuals' access to material resources like jobs, housing, credit, healthcare, childcare, education, and transportation \citep{cyberey2025prevalent}. When LLMs are embedded in decision-making systems, they can introduce, amplify, or otherwise reinforce allocational disparities, in part as a result of representational biases in the training data \citep{sen2025missing,chien2024beyond}. In pre-training, skewed representations can lead models to encode assumptions about who is qualified, creditworthy, employable, or deserving of services \citep{mehrabi2021survey}. During post-training, alignment data privileges the annotators' norms of professionalism, risk, and appropriate behavior \citep{conitzerposition2024}, which appear in downstream allocational biases as follows.

LLMs used in hiring decisions can be more likely to recommend less-prestigious jobs to speakers of marginalized dialects \citep{hofmann2024ai}. In content moderation, LLMs are prejudiced against speakers of African American English \citep{sap2019risk}. In automated exam scoring, LLMs show disparate performance for students with backgrounds not represented in training data \citep{schaller2024fairness}. And more broadly, LLM decisions are biased against underrepresented groups across domains such as business (i.e., funding a startup), finance (i.e., approving a credit card), relationships (i.e., resolving conflicts), law (i.e., issuing a passport), science (i.e., approving a research study), and the arts (i.e., awarding a filmmaking prize) \citep{tamkin2023evaluating,levy2024gender}.

\subsubsection{Mitigating Harms}
Mitigating sociotechnical harms requires interventions across the data pipeline. The first step is to establish transparent data provenance through documentation practices like \textit{Datasheets for Datasets} and \textit{Data Statements} \citep{gebru2021datasheets,bender2018data}. By explicitly recording the linguistic, demographic, and geographic composition of datasets, as well as filtering and annotation decisions, making it easier to diagnose representational gaps and biases. Model cards and system cards further extend this transparency to downstream users by documenting intended use, performance disparities, and known limitations \citep{model_cards}. Recent work argues that provenance-aware documentation should include not only source descriptions but also transformation histories, like filtering, deduplication, and synthetic augmentation \citep{scheuerman2021datasets}.

With transparent data provenance, a second mitigation step is to involve stakeholders in the process of data creation and diversification, using participatory methods \citep{vaughn2020participatory}, following \S\ref{subsub:participatory}. With community-level organization, it is possible to develop rich data resources for low-resource languages and underrepresented communities \citep{orife2020masakhane,heidt2025walking}. However, diversification alone may prove insufficient without governance structures that prevent extractive data practices and ensure ongoing community oversight \citep{benjamin2023race}. 

A third mitigation approach is to collect learnable data from user interactions with LLMs at the individual level. Personalized alignment methods may be considered in which individual preference data is collected from user interactions and used to shape subsequent model behavior through prompt-based \citep{hebert2024persomapersonalizedsoftprompt}, retrieval-based \citep{lamp}, or alignment-based methods \citep{ryan2025synthesizeme}. For more discussion on this direction, see \S\ref{subsec:personalization}.

\hypertarget{privacy}{\subsection{Consent and Ownership}}
\label{subsec:data_privacy}

The data used to pre- and post-train LLMs may include sensitive personal information (\S\ref{subsec:data_provenance}). Such personal data may actually help LLM systems become more capable, useful, and proactive. For example, LLMs can infer from users' hidden behaviors, personal habits, and broader computer use patterns, what the user might need before they even make a request \citep{shaikh2025gum}. However, the use of private or personal data comes with an array of legal and ethical challenges \citep{1_privacy_yan2024protecting, 2_subramani2023detecting}. Sensitive, private, or copyrighted information can be inadvertently leaked or reproduced without authorization, further complicating compliance with laws and regulations \citep{6_zhang2024privacy, 31_khan2022subjects, 32_wachter2019data}. Here, we will consider these concerns regarding the ownership of data. 

\subsubsection{Data Privacy Considerations}
Data privacy is broadly defined as the ability of individuals to control their personal information. Privacy leaks can occur in data, either when sensitive personal information is explicitly encoded, or when this information can be inferred \citep{1_privacy_yan2024protecting, kshetri2023cybercrime, staab2024memorizationviolatingprivacyinference}. In the former setting, LLMs can memorize personally identifiable information from training data and expose these details at inference time \citep{1_privacy_yan2024protecting, 3_memory_carlini2019secret, staab2024memorizationviolatingprivacyinference}. Attackers can exploit vulnerabilities in LLMs through methods such as backdoor attacks, membership inference attacks, and model inversion attacks, which can extract sensitive information embedded in the model during pre-training or fine-tuning \citep{1_privacy_yan2024protecting, 3_memory_carlini2019secret}. For instance, \citet{carlini2021extracting} demonstrated that it is possible to recover individual training examples, including names and phone numbers, by attacking the language model. Similarly, \citet{4_zhao2024measuring} revealed that LLMs could generate infringing content when prompted with partial information from copyrighted materials. These risks are more severe in larger models with more parameters and when longer contexts are used in prompting, making it increasingly challenging to address these vulnerabilities effectively for current LLMs \citep{karamolegkou2023copyright, carlini2021extracting, 11_carlini2022quantifying}.

Additionally, legal frameworks play a critical role in governing the use of personal and copyrighted data. Since 2018, the General Data Protection Regulation (GDPR) in the European Union has mandated data minimization, consent requirement, and the “Right to Erasure." This can be re-interpreted to apply to AI systems, though with limitations after the data collection process \citep{neel2023privacy, 32_wachter2019data}. More regulations and protocols are needed to comply with ethical obligations. Copyright law introduces another layer of complexity. Copyright law grants creators exclusive rights to use and distribute their work, with specific exceptions. Under §107 of the United States Copyright Law, the fair use doctrine permits limited usage of copyrighted materials without permission, typically for purposes such as commentary, research, or information extraction, but not for verbatim reproduction \citep{karamolegkou2023copyright}. With the increasing influence of LLMs, the use of online data has come under heightened scrutiny; justifications under principles like "Legitimate Interests" for personal data and "Fair Use" for copyrighted content are being questioned more rigorously \citep{33_franceschelli2022copyright}. Notably, companies such as OpenAI, Stability AI, and Microsoft have faced various legal challenges, including consumer privacy lawsuits and copyright infringement claims, underscoring the growing contention surrounding privacy and copyright issues in AI development \citep{news_1, news_2, news_3, news_4}. 

\subsubsection{Proactive vs. Reactive Privacy Strategies}
Adopting a proactive approach to privacy is essential. Rather than deferring mitigation until after model training, privacy considerations should inform every stage of data collection and curation. This includes implementing privacy-preserving data collection protocols, robust anonymization techniques, and consent-based frameworks from the outset. For instance, it is critical to obtain consent and minimize sensitive information collection, employ more tools to detect and remove personally identifiable information, and use more sophisticated data anonymization techniques to better protect aganst privacy leakage \cite{1_privacy_yan2024protecting, 2_subramani2023detecting}. Consent-Based Data Collection should be adopted in scenarios like web scraping to respect individual's rights \citep{2_subramani2023detecting}. Web architectures like
SOLID \citep{sambra2016solid} and Consent Tagging \citep{zhang2023tag} aim to streamline consent acquisition \citep{6_zhang2024privacy}. 

For more reactive privacy strategies after the data collection stage, various techniques have been proposed to mitigate these issues. Data cleaning methods aim to remove or generalize sensitive information from datasets before training \citep{brownLanguageModelsAre2020, ouyang2022training, bai2022traininghelpfulharmlessassistant, 19_kandpal2022deduplicating}. Federated Learning approaches decentralize the training process to enhance privacy by keeping data local and aggregating updates instead of sharing raw data \citep{20_chen2023federated, 21_yu2023federated, 22_xu2024fwdllm, 24_hoory2021learning}. Differential Privacy methods extract useful statistical information from datasets without revealing individual data by introducing controlled random noise or applying aggregation techniques \citep{24_hoory2021learning, 25_du2021dp, 26_li2021large, 27_shi2022just, 28_wu2022adaptive}. Additionally, Knowledge Unlearning techniques selectively forget or remove sensitive information from models to mitigate privacy risks \citep{5_seyitouglu2024extracting, 29_chen2023unlearn, 30_eldan2023s}.

\subsubsection{Open Challenges in Data Privacy}
Currently, privacy risks persist across the entire LLM lifecycle, encompassing not only model-centric issues but also human-centered factors. From the data side, stronger anonymization techniques and tools capable of identifying memorized personal information must keep up with LLMs’ evolving capabilities \citep{staab2024memorizationviolatingprivacyinference, 2_subramani2023detecting}. It should also be cautioned when scaling HCLLMs, as discussed in \S\ref{subsec:scaling}, that risks from memorization also increase with scale if repeated data are in the training stage \citep{hernandez2022scaling}. In addition, the complexities of obtaining consent, especially in scenarios involving third-party or inaccessible data sources, underscore the need for more robust frameworks to ensure transparent data sourcing and meaningful user control \citep{6_zhang2024privacy}. HCI researchers now also advocate for improved LLM interaction paradigms, a deeper understanding of user mental models, and systems that enable end-users to reclaim ownership over their personal data \citep{li2024human}. Despite significant progress in addressing data privacy concerns, much of the research focuses on well-known LLMs with relatively small scales. In contrast, recently released models with larger parameter sizes have received less attention due to the challenges posed by their scale, data transparency issues, and the lagging development of privacy-preserving technologies \citep{1_privacy_yan2024protecting}. Overall, greater efforts are needed to enhance legal frameworks, strengthen regulatory oversight, and advance research and technology to better safeguard privacy and copyright in the era of LLMs that developers, users, and policymakers can jointly share.

\subsection{Expanding Data Sources: Synthetic and Non-Traditional Data}
\label{subsec:synthetic_data}

\subsubsection{Synthetic Data}

We often lack high-quality, diverse, and privacy-compliant data \cite{almeida_2024_sdg_part1}. Filtering methods (\S\ref{subsec:data_provenance}) can filter out as much as 90\% of raw web text data from the Common Crawl. To replace this data, synthetic generation is one solution employed in Nemotron-CC \citep{su2025nemotron} and other popular pre-training corpora. Synthetic data generation preserves individuals’ confidentiality, replicating only the statistical properties of real datasets without retaining any personally identifiable information. LLM-generated synthetic text can also serve as fine-tuning and evaluation data \cite{vongthongsri2025synthetic}, where it is invaluable for addressing class imbalances \cite{moon2024synaug}, especially in domains like healthcare \cite{guo2024generative} where data is sensitive, and mathematical reasoning where gold examples are costly to produce \cite{chan2024balancing}.

\paragraph{Methods Used to Generate Synthetic Data.} 
Even medium-size language models can effectively expand pre-training corpora by paraphrasing existing data \citep{maini2024rephrasing}. Moreover, LLMs can effectively generate entirely new content from scratch, including textbooks for pre-training \citep{gunasekar2023textbooks} and instruction-tuning data for post-training \citep{wang2023selfinstructaligninglanguagemodels}. Procuring high-quality synthetic data with LLMs typically involves three stages: generation, curation and evaluation \cite{long2024llms}. Generation often involves prompt engineering to elicit LLM responses in the required format. This involves using strategies such as task definition, conditional prompting, in-context learning, and multi-step generation, which address context limitations and degradation over reasoning steps \cite{long2024llms,wang2024qstar}.

The generated data often contains noise or corrupted samples due to hallucination, and is generally curated using sample filtering and label enhancement techniques. Sample filtering could involve simple heuristic-based strategies or leverage the advanced language-understanding capabilities of LLMs to generate confidence scores for data points based on quality and reject samples with low scores \cite{chung-etal-2023-increasing}. Label enhancement strategies could include human inspection and annotation of low-confidence samples. These techniques are described in \ref{subsec:data_provenance_pretraining}.

After curation, the generated data must be evaluated for several components, including the statistical similarity between synthetic and real data, impact on model performance, and ensuring that synthetic data preserves essential patterns and relationships- \citet{xia2024advancing} capture these requirements in their proposed fidelity, utility, and privacy framework.

\paragraph{Making Synthetic Data More Human-Centric.} A human-centered approach to synthetic data creation should explicitly incorporate human values, perspectives, and audits at all stages of development, from generation to curation and evaluation. First, generation should serve to reflect authentic human interactions and preferences when real data collection proves slow or costly \cite{synthetic_hci}. Rather than simply increasing dataset sizes, synthetic data should contain realistic social interactions between individuals with diverse personalities and backgrounds. This requires persona alignment (\S\ref{subsec:steerability}) or role-play in which the LLM portrays a consistent identity \citep{tseng2024two}, possibly simulating a person from a particular sociodemographic background \citep{lutz2025prompt}, or an agent with a role, like a tutor or counselor \citep{li2024steerability, samuel2024personagym, shanahan2023role}. Persona alignment has been used to generate synthetic dialogues \citep{prodigy,stargate} and preference data \citep{castricato2024personareproducibletestbedpluralistic}. 

At the curation stage, stratified sampling should reflect real-world distributions along known axes of variation, such as opinions and preferences \citep{sorensen2025spectrum}. Finally, robust human-in-the-loop validation and auditing is essential. Human annotators and experts can review synthetic outputs, flag problematic patterns, and iteratively refine generation procedures. One major concern is that LLMs may reproduce biases and harms present in their training data, leaking private information or reinforcing existing social inequalities. \citet{compare} compare LLM-generated datasets with human-annotated benchmarks and highlight ethical concerns related to disparities in task performance and representational coverage. \citet{chen2024unveiling} identify several failure modes in LLM-generated query–answer pairs, including instruction-following errors. To mitigate these risks, \citet{chen2024unveiling} propose unlearning techniques to improve the reliability of synthetic queries. To preserve privacy, \citet{privacy_perserve} propose decentralized frameworks designed to reduce the likelihood of sensitive information exposure during data synthesis. These steps will ultimately enhance the quality, fairness, and usability of synthetic data to align with ethical standards and user expectations. 

\subsubsection{Non-traditional Data}
\label{subsec:non_traditional}

Recent progress in LLM research has shown the value of using non-traditional data to make models more human-centered. This discussion focuses on three primary areas. The first is multimodal data, which allows LLMs to work with inputs like speech, images, and touch. The second is human-AI interaction data, such as user feedback, edits, and eye-tracking, which helps improve how well LLMs understand and respond to user needs. Lastly, human-human interaction data uses examples of real human interactions to teach LLMs how people communicate, enabling models to better handle context, complex emotions, and relationships. 

\paragraph{Multimodal Data.}
Recent research has sought to expand Large Language Models to enable multimodality \cite{yin2024survey, li2024multimodal}, significantly enhancing human-LLM interaction by allowing systems to process and respond to a diverse range of input formats beyond text, such as speech \cite{rubenstein2023audiopalmlargelanguagemodel, huang2024audiogpt}, sound \cite{zhang-etal-2023-video, huang2024audiogpt}, vision \cite{achiam2023gpt, zhang-etal-2023-video, li2023ottermultimodalmodelincontext, pmlr-v202-li23q, fu2024a}, and tactile data \cite{fu2024a, yu2024octopi}, creating richer and increasingly human-like communication channels. By integrating multiple sensory modalities, AI can better mirror human communication, which could further improve Human-AI interaction. For instance, a multimodal AI assistant could analyze a user’s tone of voice, facial expressions, and spoken words to assess emotional states \cite{zhang2024llms, cheng2024emotion}, tailoring its responses accordingly. Recent works also explore integrating human physiological data (e.g. EEG, BVP) with LLMs to enhance empathic human-AI interaction \cite{dongre2024integratingphysiologicaldatalarge}. In applications such as education, healthcare, and accessibility, multimodality fosters inclusivity by accommodating users with diverse needs \cite{yildirim2024multimodal, belyaeva2023multimodalllmshealthgrounded, chang2024worldscribe}. Ultimately, multimodal AI systems bridge the gap between machine efficiency and human communication, making interactions more seamless, adaptive, and human-centered.

\paragraph{Human-AI Interaction Data.}
Expanding the scope of human-AI interaction data has opened new pathways for enhancing Large Language Models through both supervised fine-tuning and reinforcement learning with human feedback (RLHF). For example, Vicuna is trained with massive user-shared conversations with GPT to achieve high quality outputs \cite{vicuna2023}. Another valuable type of human-AI interaction data is human edits, where users adjust the outputs of LLMs to better match their desired results. This data can be leveraged to fine-tune LLMs for improved preference alignment \cite{shaikh2024show} or to extract user preferences more effectively \cite{ gao2024aligningllmagentslearning}. Beyond text-based interaction data, untraditional modalities such as eye-gaze signals offer additional interaction. Eye-gaze data, in particular, provides a real-time, implicit feedback mechanism that enhances context awareness and alignment with user intent \cite{konrad2024gazegpt, prokofieva2019eye, engel2023project, lopez2024seeing}. These gaze-based interactions have been shown to improve multi-modal conversational understanding and can be leveraged in RLHF workflows to refine LLM outputs dynamically \cite{lopez2024seeing}. The integration of gaze data into multi-modal frameworks would help create richer, contextually adaptive systems, fostering more intuitive, personalized, and effective interactions across diverse applications.

\paragraph{Human-Human Interaction Data.}
Real world human-human interaction data captures the nuances of human communication, including implicit cues, turn-taking dynamics, and diverse conversational contexts. Such data has the potential to improve LLMs by fostering deeper understanding of relational and situational context, thereby enabling models to generate responses that feel more natural, empathetic, and contextually appropriate. Recent advancements demonstrate how mining teacher-student interaction data, such as dialogue transcripts and collaborative problem-solving sessions, can align LLM outputs with human cognitive and emotional patterns, which allow LLMs to address complex, interdisciplinary challenges in fields such as education, psychology, and social science by emulating and learning from authentic human interaction styles \cite{wang-etal-2024-bridging, XU2024100325, wang2023sight, wang2024educonvokit}. By leveraging human-human interaction as an informative data source, we can expand the capacity of LLMs to foster meaningful, human-centered interactions in diverse real-world applications.

The integration of multimodal data, human-AI interaction data and human-human interaction data can all help LLMs more closely approximate the complexity of human communication, in turn making models more usable and reliable across high-impact domains like healthcare, education, and social services. As we exhaust traditional text data sources, recent efforts, such as MINT-1T, a multimodal text and image interleaved open-source dataset generated by \cite{awadalla2024mint1t} will be fundamental to advance the performance of frontier models.
\clearpage
\clearpage
\hypertarget{nlp}{\section{NLP for HCLLMs}}
\label{sec:nlp}

Human-centered LLMs are products of the multifaceted technical processes used to create them. NLP techniques determine not only what models can do but also the boundaries of what they \emph{cannot}. These limitations can have particular consequences as users across diverse linguistic and cultural contexts interact with LLMs.

\begin{figure}[hb!]
    \centering
    \includegraphics[width=\linewidth]{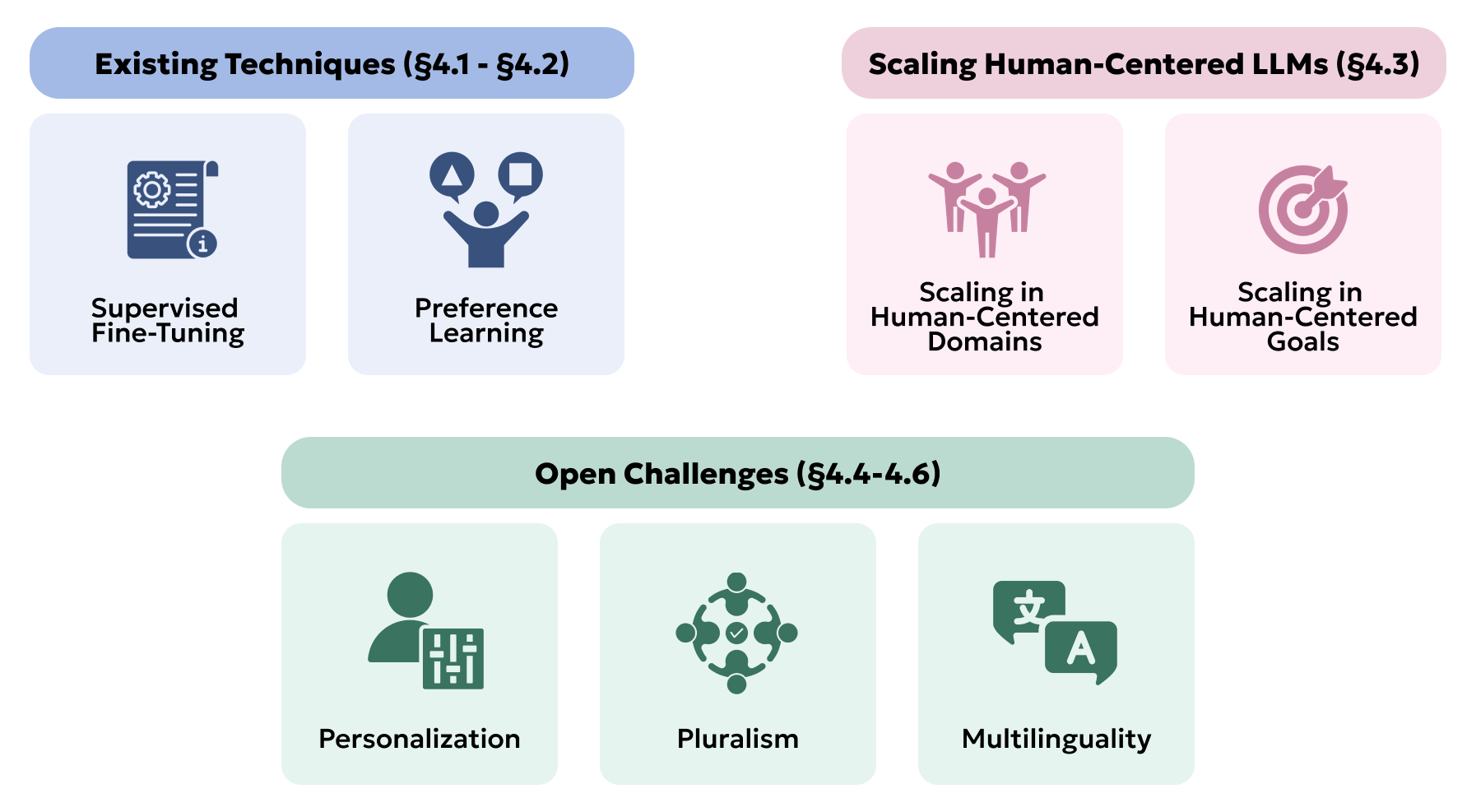}
    \caption{This chapter applies human-centered considerations to \textbf{\textit{existing post-training techniques}} like SFT and RLHF (\S\ref{subsec:instruction_tuning}-\ref{subsec:preference_tuning}), and explores the limitations of \textbf{\textit{scaling for human-centered outcomes}}  (\S\ref{subsec:scaling}). Finally, we cover open challenges in \textit{\textbf{personalization (\S\ref{subsec:personalization}), pluralistic alignment (\S\ref{subsec:pluralism}), and multilinguality (\S\ref{subsec:multilinguality})}}.}
    \label{fig:nlp}
\end{figure}

Prior survey papers cover the technical practicalities and details of NLP methods for LLMs~\cite{minaee2024large,zhao2023survey}. In this chapter, we instead focus on the human-centered considerations across the language model training pipeline. We have already discussed pre-training practices in \S\ref{sec:data} and will focus on \textbf{\textit{post-training techniques}} in this chapter. Although post-training recipes differ across models, two core components include a supervised fine-tuning (SFT) stage (\S\ref{subsec:instruction_tuning}) and a reinforcement learning stage that incorporates human preferences (\S \ref{subsec:preference_tuning}). We next discuss how the predominant paradigm of scaling applies to human-centered objectives (\S\ref{subsec:scaling}). Finally, we conclude by discussing three currently open challenges and future research directions for HCLLMs, covering \textit{\textbf{personalization (\S\ref{subsec:personalization}), pluralistic alignment (\S\ref{subsec:pluralism}), and multilinguality (\S\ref{subsec:multilinguality})}}. For a roadmap, see Figure~\ref{fig:nlp}.

\hypertarget{instruction_tuning}{\subsection{Supervised Fine-tuning for HCLLMs}}
\label{subsec:instruction_tuning}
Following pre-training, the subsequent stage in the pipeline involves some form of supervised fine-tuning (SFT), where models are trained on curated datasets to align their outputs with specific objectives or use cases. The goals of fine-tuning vary depending on the desired capabilities and target applications. For instance, existing models have employed SFT on step-by-step rationales and chain-of-thought reasoning examples to enhance their problem-solving and reasoning capabilities~\cite{muennighoff2025s1,olmo3_arxiv2025}. In other cases, SFT can also be used to adapt models for more bespoke, domain-specific applications~\cite{cheng2025finemedlm,yue2023disc}. Our discussion here focuses specifically on \textbf{instruction tuning} --- the supervised process of training LLMs on instruction-response pairs --- where models learn to follow diverse user instructions and generate appropriate responses. Instruction-tuning has become central to creating usable, general-purpose conversational AI systems. We briefly survey current practices in the literature before examining key tensions and exploring emerging frontiers for instruction tuning HCLLMs.

\subsubsection{Current Practices in Instruction Tuning} \label{subsubsec:succeses_instruction_tuning}
Instruction tuning is critical in achieving the instruction-adherance and generalized problem-solving capabilities that have helped popularized LLMs. By guiding the models to follow explicit instructions and domain-specific prompts, Instruction tuning improves LLMs' capabilities to communicate in a human-centered and user-friendly manner. While massive pre-training on self-supervised tasks improves the model’s understanding of language conventions and semantics, an LLM with an assistant-like ability to respond conversationally and complete tasks makes LLMs far more useful as human tools.

Already, we have seen the many successes of instruction tuning. It has improved performance across diverse language models, from zero-shot reasoning to domain-specific tasks. Beyond general alignment with human preferences, studies have explored ethical and domain-specific challenges, highlighting the versatility of instruction tuning. For instance, \citet{prabhumoye2023adding} demonstrated how simply attaching toxicity metadata as part of the instruction template significantly reduces toxicity present in model outputs. This approach suggests that explicitly encoding desirable or undesirable text qualities in instructions might enable the LLM to learn to promote or withhold similar texts with more ease. This demonstrates how instructions can guide models to better align with social norms and ethical considerations. 

Moreover, domain-specific instruction tuning for coding \cite{muennighoff2024octopackinstructiontuningcode}, dialogue systems \cite{ouyang2022training}, financial analysis \cite{xie2023pixiulargelanguagemodel}, and multilingual translation \cite{zhu2024finetuning} shows that modest sets of targeted instructions can unlock robust capabilities without sacrificing general knowledge. By specifying instructions to meet the requirements of each domain, LLMs can provide more reliable and appropriate outputs. These successes underscore instruction fine-tuning’s vital role in shaping LLMs into reliable, human-oriented assistants.

\subsubsection{Human-Centered Challenges with Instruction Tuning}
\label{subsubsec:challenges_instruction_tuning}
While instruction tuning can generally improve the capabilities and usability of LLMs, there are still tensions that emerge in human-centered contexts. First, as mentioned before, instruction tuning helps shape models to respond more conversationally, making them more usable for users. However, instruction tuning can lead to \emph{superficial} improvements, rather than improving the reasoning capabilities of models. For example, prior work found that models can merely learn to mimic the structure of the input data without heed to factual correctness or reliable reasoning and can fail to generalize to tasks outside of the training dataset~\cite{gudibande2023false,kung2023instruction}. When users interact with models, these patterns become especially concerning, as users may be more inclined to trust outputs as the instruction-following style appears more polished, confident, and authoritative, creating a veneer of competence that masks underlying reasoning failures~\cite{park2025critical,rathi2025humans}. This misalignment between surface-level fluency and actual reliability can lead users to over-rely on model outputs in high-stakes contexts where factual accuracy is critical.

A second tension with instruction tuning relates to \emph{safety} concerns, which we discuss in more detail in \S\ref{subsec:safety_eval}. Since instruction-tuned models are trained to comply with the provided prompt, instruction tuning can make models more susceptible to backdoor or poisoning attacks that embed malicious behaviors in datasets~\cite{prabhumoye2023adding, wan2023poisoninglanguagemodelsinstruction,shu2023exploitability}. As a result, models are more likely to produce unsafe responses, such as offensive or disallowed content. Furthermore, other work has demonstrated that instruction tuning can make models more susceptible to jailbreaking attacks, as they have been trained to follow human requests~\cite{zeng2024johnny}. Ostensibly, model designers can add more safety data to the training dataset or align models to avoid these harmful instructions. However, this practice can lead to overly cautious models that refuse to answer even benign queries~\cite{bianchi2024safety}. This resulting behavior is similarly unhelpful for users. Thus, drawing this line between what instructions are permissible to follow in service of helpfulness versus what instructions may lead to unsafe behavior remains a core design tension.

\subsubsection{Future of Instruction Tuning for HCLLMs}
What are the next frontiers for instruction tuning? A recurring theme when discussing tensions in instruction tuning is the role that data plays. As mentioned in \S\ref{subsec:data_provenance}, ensuring dataset diversity may be crucial for avoiding biases and enhancing generalization across varied tasks. To tackle this question, we can think about diversifying modalities of instruction tuning data and how data is sourced. In addition, there are new frontiers for evaluating how instruction tuned models along dimensions.

\paragraph{Multimodal data for instruction-tuning.} While our focus has centered on textual corpora for instruction-tuning, human-LLM interaction is not confined to text alone. For hands-free assistance, interacting via speech in addition to text is easier to manage for users~\cite{udandarao2025data}. For visual editing (e.g., figure design, poster-making), models ought to operate in both visual and textual modalities~\cite{pang2025paper2poster,si2025design2code}. Even in early demonstrations of intelligent assistants, our interactions are envisioned to seamlessly span multiple modalities. including speech, gesture, and images~\cite{bolt1980put}. Advancements in multimodal instruction tuning and dataset creation~\cite{li-etal-2024-multi} are thus critical for developing HCLLMs. Moving beyond text introduces modality-specific factors that are essential for human-centered applications. For instance, prosody in speech can help disambiguate user intent, and deictic gestures can provide spatial grounding~\cite{sasu2025enhancing,brooks2006working}. Furthermore, these multimodal capabilities require new approaches to both pretraining and fine-tuning on interleaved multimodal data, enabling models to process the rich integrated signals that characterize human communication~\cite{udandarao2025data,cui2025emu3}.

\paragraph{Synthetic data for instruction-tuning.}
How we obtain the requisite data for instruction-tuning remains an important open question. Synthetic data generation is an important research area that presents potential challenges for HCLLMs. From a technical perspective, synthetic data generation offers benefits for scaling data collection by reducing reliance on human labor while maintaining data diversity. There are some arguable human-centered benefits: synthetic data can democratize model development by making fine-tuning more accessible to researchers and practitioners without large annotation budgets. Empirical work has demonstrated that synthetic instruction data can be particularly valuable in low-resource settings, enabling more data-efficient fine-tuning~\cite{pengpun2024seed}. However, synthetic data generation also poses new challenges, exacerbating tensions discussed in \S~\ref{subsubsec:challenges_instruction_tuning}. Models trained on synthetic instructions may overfit to specific patterns present in the generated data, and despite claims of increased diversity, synthetic datasets can paradoxically reduce the authentic variation found in human-generated instructions~\cite{chen2024unveiling}. Figuring out how to address these flaws in synthetic data is important for leveraging this suite of methods to ensure models can handle the full range of real-world user needs and interaction styles.

\paragraph{Human-centered evaluation frameworks.} Emerging research shows that instruction tuning aligns LLMs with human brain activity, particularly in larger models with extensive world knowledge \cite{aw2024instructiontuningalignsllmshuman}. These findings open the door to more human-centered evaluation frameworks, where models are examined by how closely they mirror human-like reasoning, empathy, and context awareness. This direction has implications for designing systems that better respect ethical norms, cultural sensitivities, and user well-being. 
\hypertarget{preference_tuning}{\subsection{Learning from Human Preferences }}
\label{subsec:preference_tuning}

In recent years, the idea of fine-tuning LLMs on human preferences has seen remarkable success in improving their behavior. Previously, it was believed that training on more samples and increasing model size was sufficient to increase performance. However, researchers found that these scaling rules ignored \textit{alignment}, defined by \citet{askell2021generallanguageassistantlaboratory} as helpfulness, harmlessness, and honesty in model responses. To this end, it was discovered that incorporating human feedback directly into the training process achieved massive gains in human preference alignment \citep{askell2021generallanguageassistantlaboratory, leike2018scalableagentalignmentreward, bai2022traininghelpfulharmlessassistant}. 
As such, in this section we discuss these developments chronologically, starting with reinforcement learning from human feedback (RLHF), its spinoffs, including direct preference optimization (DPO), and recent frameworks like Constitutional AI which aim for a future of fully self-supervised AI alignment.

\subsubsection{RL-Based Methods} 
Much of RLHF is built upon landmark research by \citet{christiano2017deep}, \citet{stiennon2022learningsummarizehumanfeedback}, and \citet{ouyang2022training}. Together, these works demonstrated the feasibility of learning a reward function from human preferences and optimizing that function, first in the domain of simple robotics and Atari video games \citep{christiano2017deep}, then in the ability to improve LLM performance on summarization tasks \citep{stiennon2022learningsummarizehumanfeedback}, and finally in improving LLM behavior on a wide breadth of tasks, including open generation, chatting, and question and answering \citep{ouyang2022training}. 

Today, the canonical algorithm used to perform RLHF is proximal policy optimization (PPO) \cite{schulman2017proximalpolicyoptimizationalgorithms}, originally introduced as a simpler and more general improvement upon older RL methods like trust region policy optimization (TRPO) \citep{schulman2017trustregionpolicyoptimization}. However, there now also exists considerable research into exploring alternatives to PPO. To address PPO's high computational cost and sensitivity to hyperparamater tuning, \citet{ahmadian2024basicsrevisitingreinforcestyle} explore breaking PPO into its component pieces, and show that revisiting the formulation of human preferences in RL, discarding aspects that are unnecessarily complex for fine-tuning pre-trained LLMs, and returning to the most basic policy gradient algorithm, has yielded notable performance and efficiency gains. Other works propose alternatives to PPO entirely, such as bringing the process online for online iterative RLHF \citep{dong2024rlhfworkflowrewardmodeling} or scoring sampled responses from different sources and aligning these with human preferences (RRHF) \citep{yuan2023rrhfrankresponsesalign}.

Other works on extending RLHF focus specifically on the data that goes into aligning LLMs, whether that be improving accessibility by filling gaps in existing datasets or addressing issues of scale. For example, Okapi \citep{lai2023okapiinstructiontunedlargelanguage} is introduced as the first system and dataset to focus on RLHF for multiple languages, covering 26 different languages. For issues of scale, researchers from Google's DeepMind propose reinforced self-training (ReST) \citep{gulcehre2023reinforcedselftrainingrestlanguage}, which takes inspiration from growing batch RL to produce a dataset consisting of samples generated from the policy, which can then be used for offline training.

Beyond PPO and its proposed alternatives, there also exists considerable discussion on other portions of the RLHF pipeline, targeting better alignment through task formulation and dataset augmentation. Safe RLHF, proposed by \citet{dai2023saferlhfsafereinforcement}, explicitly decouples human preferences around helpfulness and harmlessness into two separate optimization objectives, and uses the Lagrangian method to balance trade-offs between the two.

\subsubsection{Non-RL Methods}
DPO has gained significant attention as an RLHF alternative because it enables preference tuning without an explicit reward model. It does so by directly including the probability ratio between preferred and dispreferred responses in its loss function \citep{rafailov2024directpreferenceoptimizationlanguage}. The original authors show that DPO-trained models generate responses that are preferred more frequently than those trained by PPO, and that DPO converges faster. 

Another  sample-efficient alternative, which not only avoids RL but also requires fewer than 10 samples, is Demonstration Iterated Task Optimization (DITTO) by \cite{shaikh2024show}. This method uses online imitation learning to create pairwise comparisons, treating user demonstrations as the gold standard and the model’s own outputs as dispreferred. DITTO's improvement in model alignment was shown through an average improvement of 19\% in win rates, compared to few-shot prompting and supervised fine-tuning on various human-centric tasks like news writing, emails, and blog posts.

\subsubsection{Beyond Human Feedback}
Despite the success of methods like RLHF and DPO, recent research has sought to address the potential drawbacks of only using human-sourced feedback for LLM alignment. One drawback is the incompleteness of human feedback, which may only represent a partial view of collective human values \citep{kirk2023pastpresentbetterfuture}. Furthermore, alignment is difficult to specify with explicit objectives \citep{tamkin2021understandingcapabilitieslimitationssocietal, bommasani2022opportunitiesrisksfoundationmodels}.Additionally, scaling quality and representative human feedback will become increasingly difficult with larger and more powerful LLMs \citep{casper2023openproblemsfundamentallimitations, santurkar2023opinionslanguagemodelsreflect}.

\paragraph{Constitutional AI.}
To resolve these intricate issues, recent work has shifted from using RLHF to enlisting AI help along with human collaboration to supervise other AIs to train helpful and harmless AI systems \citep{bowman2022measuringprogressscalableoversight, bai2022constitutional, saunders2022selfcritiquingmodelsassistinghuman}. As previously mentioned, one issue with alignment is that it is not clearly defined.

Anthropic's work on Constitutional AI establishes a framework designed to answer this exact question \citep{Bai2022ConstitutionalAH}. Rather than have humans provide explicit feedback, which may be inherently biased or incomplete, Constitutional AI only includes human feedback through the creation of a set of alignment principles (i.e. a "constitution"). Then, a model undergoes a training process similar to RLHF, but where the rewards are given by an LLM fine-tuned according to the values in the constitution \citep{ouyang2022training, bai2022traininghelpfulharmlessassistant}. This approach also uses chain of thought reasoning to maximize LLM self-reasoning capabilities throughout the entire process \citep{nye2021workscratchpadsintermediatecomputation, wei2023chainofthoughtpromptingelicitsreasoning}.

The end goal of Constitutional AI is not to get rid of human involvement or supervision entirely, but rather to have humans involved in only the most necessary aspects to move towards a self-supervised approach to alignment. Although Constitutional AI helped resolve many lingering issues with RLHF, this approach also brings up new questions in the ongoing research of alignment. First, how does the global AI research community come up with a widely accepted constitution that incorporates the pluralistic values of human beings \citep{hendrycks2023aligningaisharedhuman}? Second, how do we ensure a universal understanding and interpretation of the presumed constitution? How do we make sure there is a robust system for editing and improving the principles and rules as the society evolves? And when constitutional guidelines fail in ambiguous situations, how do we ensure that the models with minimal human supervision can still behave in a safe and useful way?

\hypertarget{scaling}{\subsection{Scaling Human Centered LLMs}}
\label{subsec:scaling}

In NLP, ``scaling'' refers to the relationship between a model's performance and factors such as the number of parameters $n$, dataset size $d$, and computational resources $c$ \cite{kaplan2020scaling}. Understanding these scaling laws is important for developing HCLLMs that balance efficiency and performance with accessibility and fairness.

\subsubsection{Scaling Laws in LLMs}

\citet{kaplan2020scaling} conducted foundational research on empirical scaling laws for language model performance, particularly focusing on cross-entropy loss. Their work established that model performance improves predictably with increases in $n$, $d$, and $c$, following a power-law relationship. Importantly, they discovered that returns diminish when either $n$ or $d$ is held constant, underscoring the need for a strategic, balanced approach to scaling in NLP. 

\citet{tay2022scaling} expand on \citet{kaplan2020scaling}'s work to investigate the effect of scaling properties of different inductive biases and model architectures. They find via extensive experiments that (1) architecture is an important consideration, (2) the best performing model can fluctuate at different scales, and (3) the choice of whether to scale depth (number of layers) or width (more neurons per layer) is important, especially in resource-constrained environments. Models often excel at pretraining but underperform on downstream tasks, underscoring the need to evaluate models based on human utility rather than just raw performance metrics. \citet{hoffmann2022training} introduced "Chinchilla scaling," which showed that smaller models trained on larger datasets achieve better performance per compute budget. This finding is applicable in resource-constrained human-centered applications where compute and data availability may be limited due to ethical or logistical constraints. 

\citet{ivgi2022scaling} investigate the applicability of scaling laws for different NLP tasks and find that benefits vary. Tasks aligned with pretraining objectives, such as question answering, show clearer scaling behavior compared to specialized tasks like sentiment analysis. Thus, for human-centered applications, practitioners should assess whether scaling laws are applicable to their specific use case or if full-scale testing is necessary. While scaling can improve performance for some tasks, it may not always be the most efficient path - a point further developed by \citet{liang2022holistic}, who show that similar or greater improvements to model accuracy can be achieved through more efficient human-centric means, such as training with human feedback. 

\subsubsection{Scaling in Human-Centered Domains}

Although scaling improves the overall performance of LLMs, it does not proportionally improve performance at the same rate for all subpopulations and human-centered knowledge domains. Representational biases in training data (\S\ref{subsec:data_provenance} and \S\ref{subsec:data_representation}) can lead to disparities in scaling \citep{rolf2021representation}. However, data is not the only cause of relative disparities in scaling, and may not even be the principal cause. \citet{held2025relative} found that, when holding the data scale constant, model-size scaling is responsible for \textit{widening} the performance gap between LLM performance on certain varieties of English relative to other varieties. In addition to dialect, the authors investigated AI risk behaviors \citep{perez2023discovering} and found that scaling mitigates some risks more than others.

Scaling model size alone cannot address human-domain-specific challenges such as cultural biases, as \citet{bommasani2022opportunitiesrisksfoundationmodels} highlight. The effectiveness of scaling laws varies across different domains, with some human-centered areas experiencing diminishing returns when either the number of parameters ($n$) or dataset size ($d$) is limited. For instance, \citet{brownLanguageModelsAre2020} observed that large models like GPT-3 show reduced performance gains on human-centered tasks unless they are fine-tuned with context-specific data, emphasizing the need for tailored approaches in these applications. 

\citet{gururangan2020don} conducted a study to investigate whether it is still helpful to tailor a pretrained model to the domain of a target task in a world where large-scale models, which form the foundation of today's NLP landscape, have found much success in a broad-coverage approach trained on a wide variety of sources. Overall they found that multi-phase adaptive pretraining offers large gains in task performance, implying that the quality and treatment of the data $d$ is more important than the quantity, which is relevant when dealing with sensitive or specialized human domains. 

\citet{bender2021dangers} contribute  by arguing that increasing $d$ without considering diverse and ethical data sources can lead to biased or non-representative outcomes, negatively impacting human-centered goals such as equitable access and cultural inclusivity. Furthering the discussion on fine-tuning for specific human-centric domains, \citet{zhang2024scaling} examine how different scaling factors influence the fine-tuning performance of LLMs. The study aligns with \citet{kaplan2020scaling} and \citet{hoffmann2022training}, exhibiting a multiplicative joint scaling law that links fine-tuning performance to model size, fine-tuning data size, and other scaling factors. It indicates that fine-tuning performance improves predictably when scaling both dataset size $d$ and model size $n$ and finds that LLM fine-tuning benefits more from LLM model scaling than pretraining data scaling. The work also highlights that the effectiveness of fine-tuning varies significantly based on the downstream task and the size and quality of the available fine-tuning data, underscoring the need for task-specific approaches in human-centered applications.

\subsubsection{Scaling in Human-Centered Goals}

Looking at specific human-centered evaluations, such as bias and fairness, there are potentially unexpected results when models are scaled. \citet{ethayarajh2020utility} show that leaderboard-driven scaling can create misaligned incentives in model development. By analyzing examples like the SNLI leaderboard, they demonstrate that focusing solely on state-of-the-art performance discourages practical models. For instance, with SNLI baselines at 78\% (n-gram) and 81\% (LSTM) versus a 92\% BERT-based SOTA, there is little motivation to develop lightweight models with ~85\% accuracy, which could balance performance and computational efficiency and be more practically useful. 

\citet{ganguli2022red} explored model safety through red teaming, where testers try to provoke harmful outputs. Testing models from 2.7B to 52B parameters, they found that RLHF-trained models became harder to `break' as they scaled, reducing the success of harmful attacks (the harmlessness score increased from approximately -0.5 with 2.7B parameters to 0.5 with 52B parameters). In contrast, other models showed no improvement in resisting such outputs with increased size. This study stresses the importance of incorporating human feedback during training to develop safer AI systems. Larger models  exhibit heightened privacy risks. \citet{hernandez2022scaling} demonstrate that an 800M parameter model could be degraded to that of a 400M model by repeating just 0.1\% of the training data 100 times, suggesting that larger models aren't automatically more robust to certain types of data-based attacks (see \S\ref{subsec:data_privacy}). 

In regards to emulating human values, \citet{biedma2024beyond} showed that as language models get larger, they show an increased preference for the task-oriented values like accuracy and factual consistency at the slight expense of social intelligence or moral fiber and adherence to ethical norms. While larger models may become more capable, their value systems have the potential to become increasingly misaligned with human values. 

Transfer learning is often a prerequisite for the application of LLMs in human-centered tasks. Scaling laws suggest that a model's ability to transfer knowledge improves as its performance increases. This relationship generally holds in human-centered evaluations, with studies showing that transfer learning benefits from scaling in broad tasks \cite{hernandez2021scaling, raffel2020exploring}. However, \citet{hernandez2021scaling} and \citet{raffel2020exploring} highlight the importance of domain-specific fine-tuning and alignment techniques to achieve human-centered objectives in specialized or sensitive areas such as legal and medical contexts.

\subsubsection{Inference time scaling}

Recent advances in inference-time scaling offer pathways to improve HCLLMs without retraining. Now more targeted approaches to inference-time adaptation are emerging that specifically address human-centered concerns. \cite{zhang2024controllablesafetyalignmentinferencetime} introduce Controllable Safety Alignment (CoSA), a framework that enables inference-time adaptation to diverse safety requirements without model retraining. Rather than following a one-size-fits-all approach to safety alignment where models refuse any potentially unsafe content, CoSA allows authorized users to modify safety configurations at inference time through natural language descriptions of desired safety behaviors. 

Despite these promising directions, there remain open questions about whether increased inference compute might undermine certain human-centered objectives. OpenAI’s emphasis on chain-of-thought (CoT) in their o-series of models, for example, underscores the prevailing focus on inference-time reasoning strategies \cite{learningtoreasonwithllms}. However, \citet{shaikh-etal-2023-second} find that explicitly prompting models to “think step by step” can inadvertently increase harmful biases and toxic outputs, observing an 8.8\% rise in biased responses and a 19.4\% increase in toxicity across relevant benchmarks. Their study suggests that while inference-time reasoning holds promise for improved performance or controllable safety alignment, it may also expose underlying biases.

\subsection{Personalization}
\label{subsec:personalization}
We established that the goal of the post-training stages is to align LLMs to human preferences. Nonetheless, the term ``human preferences'' is blanket statement that obscures the diverse and potentially conflicting desires that different users have. Rather than trying to create LLMs that can satisfy everyone, the goal of personalization is to align a model's outputs with the preferences of a single individual, both in terms of content and style~\citep{zhang2024personalization,tseng2024two}. 

\subsubsection{Current Approaches}
We cover three families of techniques for personalizing LLMs: prompting-based approaches, retrieval-based approaches, and personalized alignment (e.g., learning from human preferences). There is no ``best'' technique for personalization and instead depends on factors, such as what data is available, compute efficiency, degree of personalization required, and so on.

\paragraph{Prompting-based Approaches.} Given the capabilities of LLM, prompt-based approaches provide an efficient and popular approach to personalizing model outputs. One prompt-based technique is to provide a user persona directly in the text instructions provided to the model. Here, the critical research decision is figuring out what content to provide for personalization. For example, prior work has explored providing demographic information (e.g., race, gender) to the model for personalization, although these personas run the risk of stereotypes or caricaturization~\cite{cheng-etal-2023-compost, cheng-etal-2023-marked,gupta2023calm,huang2023humanity}. In addition, this approach has been adopted to imbue character traits upon the model, such as prompting the responses to reflect certain personality qualities (e.g., extroversion, warmth) or different tones~\cite{}. Since these methods can suffer from inefficiency and information loss~\cite{Liu2023LostIT}, other works have also explored embedding user information into tokenized prompt embeddings~\cite{li2024personalizedlanguagemodelingpersonalized,hebert2024persomapersonalizedsoftprompt,huang2024selectivepromptingtuningpersonalized}. 

\paragraph{Retrieval-based Approaches.} What information about the user is relevant depends on the context. Retrieval augmented personalization methods retrieve information from an external knowledge base that is then incorporated at inference-time to personalize the model's output. As an initial foray into this area, \citet{lamp} benchmark different retrievers, including BM25 and a dense retrieval model, on a series of personalization tasks. Other work has continued to refine retrieval methods, exploring how to improve capabilities while reducing the amount of retrieved data via summarization~\cite{rag_summary} or adopting techniques from other areas such as collaborative filtering~\cite{shi2025retrieval}. Although retrieval-based methods allow for on-the-fly personalization of models at inference-time, the performance is constrained by retriever quality. For example, lexical retrievers, such as BM25, may match on shallow keyword similarity rather than deep semantic understanding of user needs. Furthermore, this approach may be more constrained in cold-start situations or domains with sparse user data, where the knowledge base itself may be insufficient to support meaningful personalization.
    
\paragraph{Personalized Alignment.} Finally, rather than adapting model behavior at inference time, training-based approaches can embed individual preferences into alignment objectives. For example, several works present methods for creating personalized reward models, such as approaches that train multiple reward models that are later merged~\cite{jang2023personalized}. Critically, these approaches require that the dimensions for personalization are defined a priori, limiting the scope to which they can be applied. Alternative approaches have sought to loosen these constraints, proposing methods that learn preferences from historical user interactions, which can then be used to train personalized reward models or directly align LLMs~\cite{ryan2025synthesizeme,poddar2024personalizing,balepur-etal-2025-whose}. However, training-based personalization faces practical limitations: these methods require substantial amounts of user data in domains where data is often scarce, risk overfitting to individual user patterns, and incur significantly higher computational costs compared to retrieval-based or prompting approaches~\cite{ryan2025synthesizeme,shaikh2024show,zhang2024personalization}. These tradeoffs make training-based personalization most suitable for scenarios where there is sufficient data present and customization justifies the additional resource investment.

\subsubsection{Future of Personalization for HCLLMs}
A central challenge in personalization is deciding what the model should adapt to. For example, economists will distinguish between stated preferences --- what people say they want --- and revealed preferences—inferred from behavior~\cite{samuelson1948consumption}. In language model personalization, this tension also manifests. Behavioral signals such as query reformulations, response ratings, or conversation length may reflect immediate satisfaction but diverge from users' stated goals or long-term values. Consider a user learning a new subject who frequently requests direct answers. Based on inferred signals, the model might be personalized to comply whereas a model targeting learning outcomes might instead offer scaffolded hints. This raises fundamental questions about system objectives and user agency: which preferences should dominate when conflict arises, and how should systems handle patterns users exhibit but might not endorse upon reflection? 

Temporal dynamics present an additional challenge. User preferences and needs evolve over timescales, ranging from within-session learning to long-term skill development and shifting life contexts. Yet, current approaches treat personalization as a static task. While recent work has extended the length of conversation in personalization datasets, we lack real-world benchmarks that capture these longer-term dynamics or benchmarks for evaluating personalization over time~\cite{kirk2024the,zhang2024personalization,zhao2025llms}. There is some early work that explores updating user profiles over time, but critical questions remain open~\cite{wang2024lifelong}. For instance, how can systems distinguish transient preferences (a user exploring a new hobby) from enduring ones (a domain expert's consistent working style)? What mechanisms allow for efficient model updates under these circumstances? These temporal considerations compound the preference alignment challenges; even if we can perfectly identify and incorporate a user's current preferences into LLM outputs, those preferences themselves may be moving targets.
\subsection{Pluralism}
\label{subsec:pluralism}
While we have discussed methods for aligning models to human preferences and values, the important, unaddressed question is \emph{whose} values. Not all humans share the same set of values~\cite{durmus2024towards}, and what is permissible to some may be irrelevant or harmful to others. Rather than assuming values are monolithic and aligning models to a ``Silicon Valley default'' set of preferences, \emph{pluralistic alignment} is an approach of aligning language models to simultaneously serve diverse preferences. As \citet{DBLP:conf/icml/SorensenMFGMRYJ24} define, pluralistic alignment is the process of creating language models capable of representing a diverse set of human values and perspectives. 

\subsubsection{Current Approaches}
\paragraph{Defining Pluralism} Prior work has proposed three definitions as to how pluralism can be embedded into models~\cite{sorensen2024roadmap}:
\begin{enumerate}
    \item \textbf{Overton Pluralism}: An Overton pluralistc model generates all \emph{reasonable} perspectives for a given subject. The model draws on the concept of the ``Overton window,'' which encompasses the range of ideas and perspectives that are considered acceptable within the mainstream. Scholars such as \citet{lake2024from} have posited that Overton pluralism is compatible with the status quo alignment process, as longer conversational outputs typically share a few varied perspectives. 
    \item \textbf{Steerable Pluralism}: Steerable pluralism advocates for using situational context to steer the model towards a particular perspective. One popular use case of steerable pluralism is improving the cultural alignment of language models to particular groups and cultures \cite{masoud2024culturalalignmentlargelanguage, tao2024cultural}. Personalization also falls under steerable pluralism as the model is steered to the values of a particular user. 
    \item \textbf{Distributional Pluralism}: Finally, in distributional pluralism, developers steer models to produce responses that roughly correspond with a population distribution. That is, if 70\% of the population holds a particular opinion, the model will generate that opinion 70\% of the time. As \citet{lake2024from} note, base models are somewhat distributionally aligned to the opinions present in pre-training data.    

\end{enumerate}

\paragraph{Methods for Pluralistic Alignment.}
Several works have proposed different methods for alignment, depending on what type of pluralism they seek to achieve. For Overton pluralism, researchers have explored prompting techniques~\cite{meincke2024promptingdiverseideasincreasing} and approaches that generate multiple perspectives before synthesizing them~\cite{feng-etal-2024-modular, hayati-etal-2024-far, li2024largelanguagemodelsecretly}. For steerable pluralism, personalized reward models~\cite{jang2023personalized, poddar2024personalizing, chen2024pal} and optimization techniques (e.g., Group Preference Optimization~\cite{zhao2023group}) enable alignment to specific user or group preferences with minimal context. Other common implementation methods include targeted prompting \cite{alkhamissi-etal-2024-investigating} and generate-then-filter approaches \cite{feng-etal-2024-modular}. Finally, for distributional pluralism, researchers generate diverse perspectives and filter based on population distributions \cite{feng-etal-2024-modular}, with surveys serving as key measurement tools since they enable matching LLM probability distributions to actual population responses. OpinionsQA~\cite{santurkar2023opinionslanguagemodelsreflect} and GlobalOpinionsQA~\cite{durmus2024towards} are two widely-used survey benchmarks for evaluating distributional alignment.

In tandem, a data-centric approach to pluralistic alignment has focused on collecting datasets that represent a diversity of values and perspectives. Chatbot Arena \cite{DBLP:conf/icml/ChiangZ0ALLZ0JG24} and PRISM \cite{kirk2024the} are two such collections of preference data that retain user labels.  The PERSONA dataset attempts to simulate 1,500 users with synthetic personas for the purpose of studying pluralistic alignment, providing synthetic prompts and feedback pairs \cite{castricato2024personareproducibletestbedpluralistic}. The ValuePrism dataset \cite{sorensen2024value} is a collection of values, rights, and duties applied to various scenarios.  They also release the KALEIDO model for measuring how certain statements agree or disagree with particular values.  ValueConsistency \cite{moore-etal-2024-large} is a dataset of 300 controversial topics in four languages with corresponding controversial questions.  Moral Stories \cite{emelin-etal-2021-moral} is a dataset of narratives with moral dilemmas with multiple endings.  The stories were written from annotators of various demographic backgrounds and express different social and moral norms.

\subsubsection{Future of Pluralistic Alignment for HCLLMs}
Achieving effective pluralistic alignment faces several challenges. First, figuring out how to appropriately and effectively model diverse perspectives is a key precursor to effective pluralistic alignment. Existing literature has shown that models tend to emphasize stereotypical representations when asked to simulate perspectives from different demographic groups~\cite{cheng-etal-2023-marked, cheng-etal-2023-compost, deshpande2023toxicity}, meaning that approaches that generate and synthesize varied perspectives require careful design to avoid misrepresenting communities. While prompting LLMs with demographic features can improve alignment with group opinions~\cite{alkhamissi-etal-2024-investigating}, the choice of base model sometimes has a more significant effect than the sociodemographic features themselves~\cite{beck-etal-2024-sensitivity}. Effective perspective modeling remains an open problem.

A second challenge involves understanding and balancing the societal impacts of pluralistic models on public opinion. It is well-established that aligning LLMs to human preferences can exacerbate model sycophancy \cite{sharma2024towards}, where models agree with users regardless of validity. Personalized models developed through steerable pluralism risk creating personalized echo chambers that reinforce rather than challenge user beliefs. Already there are  concerns about ``echo chambers'' in media consumption, which may only be further augmented with LLMs~\cite{cinelli2021echo,barbera2020social}. Even though pluralistic alignment aims to broaden the viewpoints that models may espouse, these methods still define what perspectives are considered ``in bounds'' of acceptability or follow ostensible public opinion distributions. In doing so, these methods may inadvertently silence marginalized viewpoints that fall outside mainstream acceptability. Future work should empirically investigate how different pluralistic alignment strategies affect opinion formation and marginalization in practice, moving beyond theoretical concerns to measurable impacts on diverse user populations.

Finally, pluralistic alignment raises questions of data sovereignty and consent. Many communities, particularly indigenous groups, do not want their data and opinions collected for training AI systems \cite{rainie2019indigenous}. Even when motivated by the goal of broad representation, developers must carefully consider whether all communities want their perspectives embedded in AI systems, respecting the right of groups to opt out of technological representation entirely.
\subsection{Multilinguality}
\label{subsec:multilinguality}
Beyond being able to capture the needs, values, etc. of users, users ought to also be able to interact with models in their preferred language. Although there are over 7,000 languages spoken worldwide, most of LLM development focuses on English \citep{held2023material}. Expanding the multilingual capabilities plays a critical role in helping democratize access to language models, particularly bridging the gap between high-resource and low-resource language technologies.

\subsubsection{Current Approaches}
Multilingual large language models (MLLMs) are systems capable of both understanding and generating text in multiple languages. While most LLMs perform best in English, there is a concerted effort to improve their multilingual capabilities. From a data-centric perspective, these efforts focus on curating diverse linguistic resources, including pre-training corpora and post-training datasets for supervised fine-tuning (SFT) and preference learning. Many pre-training corpora, such as RedPajama~\cite{weber2024redpajama} and CC-100~\cite{conneau_etal_2020}, are collected via web scraping. Furthermore, many existing pre-training corpora that are primarily in English already include a small percentage of non-English data; incorporating even this small amount of data during pre-training can improve cross-lingual capabilities~\cite{blevins2022language}. Nonetheless, a core challenge is that the Internet remains heavily English-centric, leading to performance gaps for less-resourced languages that appear infrequently online or, in some cases, lack standardized written forms. For an in-depth survey on multilingual LLMs, please refer to \citet{qin2025survey}. 

Rather than relying solely on naturally occurring multilingual text, researchers have turned to translating high-resource English text into other languages. For example, \citet{wang2025multilingual} translate the pre-training corpus FineWeb into multiple languages for pre-training. Similar approaches exist for creating post-training datasets, such as the Aya~\cite{ustun2024aya} and CrossAlpaca~\cite{ranaldi2024empowering} corpora, which rely on either machine translation or existing MLLMs to assist with translation~\cite{lai2023okapiinstructiontunedlargelanguage,yue2024pangea}. However, translated datasets can introduce artifacts (referred to as \emph{translationese}) that shift linguistic patterns and degrade data quality, particularly for downstream reasoning tasks and open-ended generation~\cite{dang2024rlhf,vanmassenhove2021machine}. While translation is an efficient way to generate large amounts of multilingual data, another question we must ask is what gets lost in translation? For example, when translating from text in English, the resulting output might lack important cultural context and nuance that would be found in text originating in places that speak the selected language~\cite{qin2025survey}. 

Finally, complementary to training interventions are non-training approaches enabled through prompting. A wide range of prompting strategies for multilingual use has been proposed, including those surveyed in recent work by \citet{vatsal2025multilingual}. One common strategy is ``translate-test'', or to translate user input into English before performing the task, leveraging the stronger English proficiency of most current LLMs~\cite{liu2025translation,artetxe2023revisiting,huang2022zero,etxaniz2024multilingual,huang2023not}. While this approach often boosts performance, it can fall short for tasks requiring cultural nuance, idiomatic understanding, or language-specific world knowledge, where remaining in the original language is critical for faithful interpretation and generation~\cite{liu2025translation}.

\subsubsection{Future of Multilinguality for HCLLMs}
Looking forward, efforts toward \emph{human}-centered multilingual capabilities must consider the following areas. First, there is a growing interest in multilingual safety. As \citet{yong2025state} identify, multilingual safety remains massively underrepresented as a research domain, resulting in safety standards built for English that do not translate effectively to other linguistic contexts. Treating English as the universal reference point obscures sociolinguistic variation and produces a gap between how models behave and how safety norms should operate for real users across languages. Understanding and addressing multilingual safety thus remains an open frontier.

A second challenge involves determining what it means to represent language in ways that do not exploit the communities that speak it. The desire for multilingual NLP is not new. Projects such as Meta’s No Language Left Behind demonstrate longstanding investment in broad language coverage. However, as \citet{bird2024must} argues, these efforts often treat language as a detached artifact rather than something rooted in communities, cultures, and social practices. When language is treated as a pure optimization target, scraped data, or a resource to be ``unlocked,'' the resulting systems risk extraction without contributing tangible value to the communities whose linguistic labor enables them.

Beyond technical considerations, achieving authentic multilinguality in human-centered systems demands rethinking how data is gathered, whose language practices are modeled, and for what ends. Simply scaling web pre-training and technical fixes may improve multilingual benchmarks, but it does not confront the deeper question of whether these systems advance the needs, agency, and self-determination of speakers. Efforts to ``democratize'' access to LLMs echo earlier narratives that cast computing as a universal solution. As the Information and Communication Technology for Development (ICT4D) literature reminds us~\cite{toyama2015geek, harris2016ict4d}, such narratives risk reproducing existing inequities when social, cultural, and political contexts are ignored.
\clearpage
\hypertarget{evaluation}{\section{Evaluation}}
\label{sec:evaluation}

Evaluation methods allow model developers, users, and stakeholders to compare the capabilities and limitations of different LLMs, and to understand the scope of their utilities and risks in particular domains \citep{10.1145/3641289}. Evaluations can inform all stages of model development, from mixing pre-training data \citep{held2025optimizing,mizrahi2025language} to selecting and optimizing reward models for alignment \citep{lambert2024rewardbenchevaluatingrewardmodels,frick2024evaluate}. After models are trained, evaluations arguably become even more important. They act quality filters to decide whether and how companies deploy models \citep{liang2022holistic}. They inform researchers on the most important and promising directions for model improvement \citep{srivastava2022beyond}, and help anticipate models' future capabilities \citep{kaplan2020scaling,hoffmann2022training}. And finally, they shape public perceptions \citep{liao2022designing}, and inform policy and other regulatory decisions \citep{eriksson2025can}. 

\begin{figure}[hb!]
    \centering
    \includegraphics[width=\linewidth]{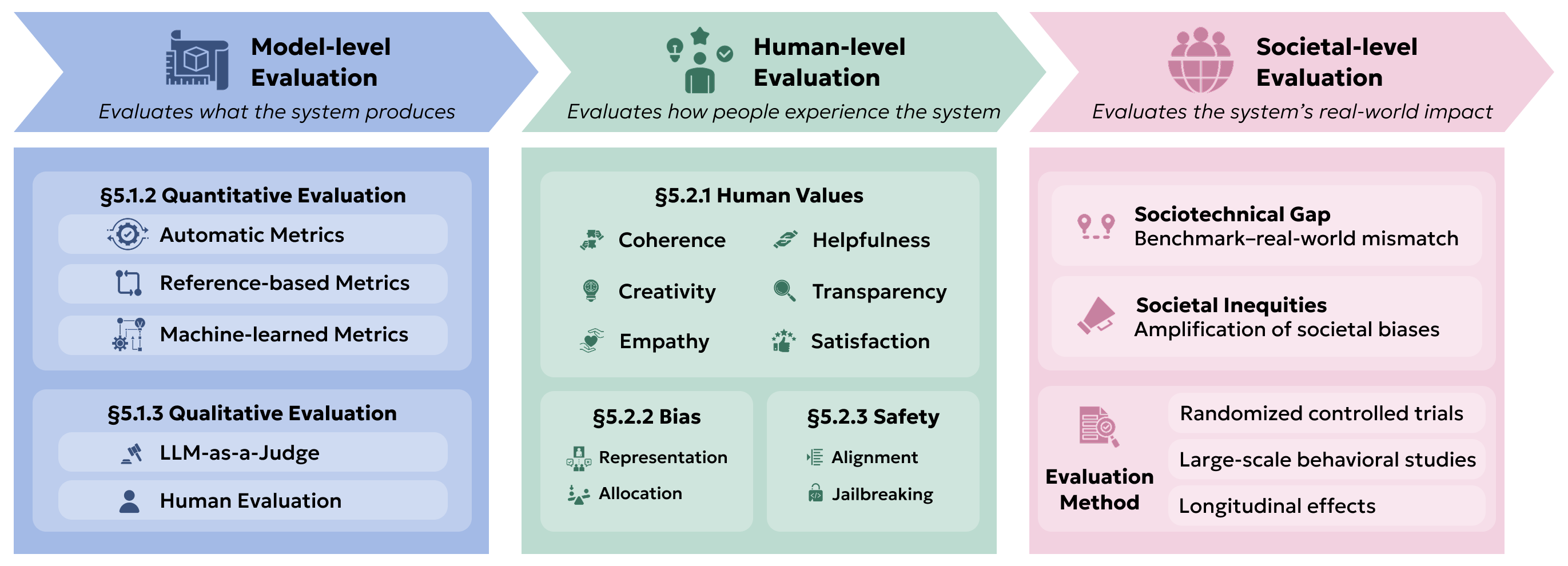}
    \caption{In this chapter, we discuss common pitfalls and best practices for evaluating HCLLMs, considering three distinct levels of evaluation: the model level (\S\ref{subsec:model_level_eval}), the human level (\S\ref{subsec:human_centered_eval}), and the societal level (\S\ref{subsec:impact}).}
    \label{fig:evaluation}
\end{figure}

Without a human-centered evaluation of LLMs, model development, deployment, and governance may be oriented not towards the long-term and collective good, but rather towards profit incentives and short-term gains on surface-level heuristics \citep{eriksson2025can}. In this chapter, we observe common pitfalls and highlight best practices in human-centered evaluation, spanning three levels as shown in Figure~\ref{fig:evaluation}.  First, we consider evaluations at the level of model outputs (\S\ref{subsec:model_level_eval}), using both quantitative metrics (\S\ref{subsub:quantitative_methods}) and qualitative evaluations (\S\ref{subsub:qualitative_evaluation}). Beyond raw outputs, we also consider how people experience LLMs (\S\ref{subsec:human_centered_eval}), considering human values (\S\ref{subsub:human_values}), as well as concerns over bias (\S\ref{subsec:bias_eval}) and safety (\S\ref{subsec:safety_eval}). Lastly, we discuss extrinsic evaluations at the societal level (\S\ref{subsec:impact}), measuring the system's real world impact. 

\hypertarget{benchmarks}{\subsection{Model-Level Evaluations}}
\label{subsec:model_level_eval}

\hypertarget{benchmarks}{\subsubsection{Benchmarks}}
\label{subsec:benchmarks}

Benchmarking has long been a key driver in the development of AI systems. Benchmarks act as a compass, encoding the values, priorities, and goals of the AI research community \cite{ethayarajh2020utility, birhane2022values}. They help determine not only how capable a model is, but also what we consider meaningful progress, allowing us to compare the strengths of different models. 
Recent research has argued for a shift in what and how we evaluate, especially for human-centered applications. As LLMs increasingly influence real-world decision-making, especially in domains like education, law, healthcare, and customer support, the limitations of traditional benchmarks become even more critical. Benchmarks must evolve to better represent human values such as fairness, robustness, usability, and positive societal impact. 
Below, we discuss a few general principles for thinking about evaluating human-centered LLMs. 

\paragraph{Moving Away from "Exams" and Rethinking What We Evaluate.}
Traditional benchmarks often mimic academic exams, assessing LLMs by how well they can replicate human outputs or solve static problems in standardized formats like multiple-choice questions \citep{hendrycksMMLU2021}, math problems \citep{sun2025challenging}, and code generation \citep{jimenezswe}. While useful, this framing compresses complex, multidimensional model behavior into a single metric. Even interaction-based, human-voting evaluations like ChatbotArena \citep{DBLP:conf/icml/ChiangZ0ALLZ0JG24} are limited by their brittleness or their misalignment with how humans actually use AI \citep{singh2025leaderboard}.
In real-world use, LLMs are collaborators, copilots, or tools embedded in workflows, so it becomes necessary to evaluate them in natural, complex, and multi-step human-AI interaction settings, not just in isolation. 
One promising alternative is centaur evaluations \cite{haupt2025ai} where humans and models collaborate. Here, we care about the outcome of the combined system. These setups get closer to how AI is actually used in practice, whether it's writing, analysis, customer support, diagnosis, or decision-making.

\paragraph{Ecological Validity.}
A central challenge in evaluating LLMs for real-world use is ecological validity, the extent to which a benchmark setting reflects the complexity of how systems are actually used. Controlled evaluations may offer cleaner signals, but they often fail to generalize to interactive, user-facing deployments. Recent work \cite{li2025mind} has shown that no single benchmark strongly correlates with interactive performance for audio models across 20 existing datasets. A model that excels at standard static tasks might still struggle in dynamic or collaborative environments. This mismatch suggests a need for richer, context-aware evaluations.
One promising direction is to build evaluations bottom-up from in-the-wild data. For example, \citet{rottger2025issuebench} evaluate perspective and framing biases in LLM responses to natural user queries. Benchmarks should also be robust and reliable, correlating good performance with success in real tasks. This requires vetted examples with accurate annotations and sufficient statistical power \cite{bowman2021will}. Finally, effective benchmarks should 
reveal potential biases, artifacts, and any dual uses, as well as ways to mitigate such unintended consequences \cite{weidinger2022taxonomy}. 

\paragraph{Data Contamination and Dynamic Alternatives.} 
With LLMs trained on massive web-scale corpora, the risk of benchmark contamination has become a serious issue. Many popular benchmarks are at least partially contained in training data, making their validity as evaluation tools less convincing. The line between training and testing becomes blurry, especially for static tasks. This is one benefit of dynamic, evolving benchmarks. Examples like DynaBench \citep{kiela2021dynabench}, Chatbot Arena \citep{DBLP:conf/icml/ChiangZ0ALLZ0JG24}, WebArena \citep{zhou2023webarena}, and WildVision-Arena \citep{lu2024wildvision} introduce a degree of human involvement that better mirrors real-world interaction. Such dynamic setups are promising for evaluating generalization and interaction aspects and mitigating issues around saturation and contamination. 

\paragraph{General-Purpose vs. Domain-Specific Evaluations.}
For example, DR Bench \cite{gao2023dr} assesses LLMs’ diagnostic reasoning abilities, PubMedQA \cite{jin2019pubmedqa} targets biomedical research question-answering, and LegalBench \cite{guha2024legalbench} is designed for legal reasoning, including statutory interpretation and contract analysis. In education, benchmarks are emerging to evaluate LLMs' effectiveness in providing innovative and meaningful feedback to teachers \cite{wang2023chatgpt} and emulate expert decision-making in providing tailored math remediation help bridge the gap between technological capability and educational needs \cite{wang-etal-2024-bridging}. 
Recently, GDPval measures model performance on economically valuable, real-world tasks across 44 occupations \cite{openai_gdpval_2025}. 
These specialized benchmarks provide useful signals that evaluation is grounded in each specific context, offering contextualized and real-world assessment of model performance compared to simply on math and coding tasks. Collaborative efforts across domains are crucial to developing benchmarks that reflect the full complexity of human-LLM interactions and the contexts in which LLM systems are deployed.

Overall, a human-centered framework often transcends traditional metrics and benchmarks that continue to prioritize efficiency and profitability above all else. While these measures are useful in providing objective algorithmic reviews on quantitative criteria, they fail to, or sometimes not even attempt to, capture the human factors and societal patterns that are inherently present in these systems.
\subsubsection{Quantitative Evaluation} 
\label{subsub:quantitative_methods}

\textbf{Automatic Metrics.}
Notably, among automatic metrics, foundational methods have played a critical role in shaping intrinsic evaluations. These metrics provide a systematic way to evaluate model performance through standardized benchmarks, making the evaluation process more efficient and scalable \citep{10.1145/3485766, askell2021generallanguageassistantlaboratory, hu2024unveilingllmevaluationfocused}.

Metrics like BLEU \cite{papineni2002bleu} and ROUGE \citep{lin2004rouge} are valued for their simplicity and reproducibility. However, their shortcomings include reliance on strict token matching, which often penalizes valid paraphrases and fails to capture deeper semantic equivalence \citep{wieting-etal-2019-beyond}. Even embedding-based metrics like BERTScore~\cite{hanna-bojar-2021-fine} can be fooled by lexical similarity, ranking a more similar incorrect translation higher than a dissimilar but correct one.

Generally, quantitative metrics also are limited in addressing other human needs, such as interpretability, latency, cognitive load, and user satisfaction. Optimizing solely for a metric like perplexity can lead to monotonous responses from a language model \cite{celikyilmaz2021evaluationtextgenerationsurvey}, which would be less appealing to a user. In the high-stakes domains such as healthcare, where applications are highly critical, existing metrics have been found to fail to capture trust, personalization, and empathy \cite{abbasian2024foundation}. Finally, while these metrics may be automatic, they are often not scalable to tasks such as open-ended question answering and complex planning \cite{gehrmann2023repairing}. These limitations have led to the development of complementary and alternative evaluation methods. 

\textbf{Reference-based Metrics.} Reference-based approaches measure the similarity between the system output and the predefined reference samples, such as cosine similarity \cite{agarwal2024aisuggestionshomogenizewriting}, the E2E benchmark \cite{banerjee2023benchmarkingllmpoweredchatbots}, HUSE \cite{hashimoto-etal-2019-unifying}, and Reward Bench \cite{lambert2024rewardbenchevaluatingrewardmodels}. These methods maintain the benefits of standardized and objective evaluation for automatic metrics, but they are also limited to the quality of the used standard. For instance, they may be inconsistent or disprove themselves against new references or optimize for closeness to a single gold standard, even if the overall response quality is worse \cite{nguyen2024referencebasedmetricsdisprovequestion}. For creative tasks, such a gold standard may not even exist. 

\textbf{Machine-learned Metrics.}
Machine-learned metrics such as reward models \cite{ryan2024unintendedimpactsllmalignment} and classifier-based scoring \cite{shaikh2024rehearsal} show some promise in capturing nuances of human judgment. However, it can be challenging to build pipelines to ground LLMs, such as with specific sources for factual correctness \cite{tang-etal-2024-minicheck}, or to social science theories that reflect human behavior and preferences \cite{shaikh2024rehearsal}. Additionally, these methods face limitations in generalizing to out-of-distribution settings, particularly in addressing discrepancies in preferences across different groups of people worldwide \cite{ryan2024unintendedimpactsllmalignment}.

\subsubsection{Qualitative Evaluation}
\label{subsub:qualitative_evaluation}

In contrast to quantitative evaluations (\S\ref{subsub:quantitative_methods}), qualitative evaluations require a nuanced approach to evaluation as they work directly with humans (or LLMs). They are perhaps more human-centered than automatic or machine-learned metrics due to their subjects, while requiring more careful considerations to design fair and effective evaluations. We first discuss two paradigms of qualitative evaluations, LLM as a Judge and Human Evaluation. We then end the section with a coverage of Extrinsic Evaluation.

\textbf{LLM-as-a-Judge.} The rise in popularity of LLMs has led to the ``LLM-as-a-Judge'' paradigm, which caters towards more human-centered systems. Given the cost and subjectivity of human evaluation, LLM evaluation proves to be a feasible alternative, and the results are generally consistent with results from human experts on some tasks \cite{chiang2023largelanguagemodelsalternative}. Within the LLM judge paradigm, there are various use cases, such as LLM-derived metrics (embedding-based, probabilities, etc.) \cite{jia-etal-2023-zero, xie-etal-2023-deltascore}, prompting, fine-tuning LLMs with human evaluations \cite{xu2023instructscoreexplainabletextgeneration, ke-etal-2024-critiquellm}, and human-LLM collaborative evaluations \cite{gao2024llmbasednlgevaluationcurrent}. More recent methods employ multiple LLMs to engage in multi-agent debates for evaluations and have shown better alignment with human assessment \cite{chan2023chatevalbetterllmbasedevaluators}. However, LLM-based evaluators exhibit systematic limitations, including self-preference bias \cite{panickssery2024llmevaluatorsrecognizefavor}, where models favor their own outputs, and inconsistent application of evaluation criteria \cite{hu2024llmbasedevaluatorsconfusingnlg}, both of which reduce the reliability of their judgments. One set of limitations stems from hallucinations and lack of consistency and reproducibility that impacts accuracy of responses. Furthermore, LLMs can exhibit biases similar to human cognitive biases, e.g., gender and authority bias \cite{chen-etal-2024-humans}. They also show self-preference to LLM-generated outputs \cite{panickssery2024llmevaluatorsrecognizefavor}. Researchers study agreement between human and LLM evaluations using metrics such as Intraclass Correlation Coefficient (ICC) \cite{bartko1966intraclass} and Cohen's Kappa \cite{li2024llmsasjudgescomprehensivesurveyllmbased, warrens2015five}. However, these issues are exacerbated by humans over-trusting LLM outputs for supposed objectivity in application settings \citep{bansal2021does}. One approach to address this issue is the ``LLM as a jury’’ paradigm proposed by \citet{verga2024replacingjudgesjuriesevaluating}, to check back on bias perpetuated by a single judge and thus better align with human evaluation.

\textbf{Human Evaluation.} Crowd-sourcing platforms such as Amazon Mechanical Turk  (MTurk)\footnote{https://www.mturk.com/} and Prolific\footnote{https://www.prolific.com/} have enabled large-scale experiments within budget. Researchers have access to a wider range of evaluators than they would have in in-person studies. Nevertheless, human evaluators online may exhibit biases and quality issues~\cite{ipeirotis2010quality}. In addition, evaluators' demographics could be skewed depending on the platform~\cite{difallahMechanicalTurk}. Correspondingly, the data quality between the platforms might differ. \citet{douglasDataQuality} shows that Prolific and CloudResearch are more likely to produce high-quality data, in comparison to MTurk, Qualtrics, and SONA. However, these trends may be shifting as AI agents more readily mimic human respondents and bypass AI detection methods \citep{westwood2025potential}.

Such human evaluations must be designed according to best practices. Relevant questions are, how are human ratings collected? What questions are asked? We must design human evaluations carefully to avoid low-quality annotations. There exists a difficulty in standardization of human evaluations.~\citet{huynh2021surveynlprelatedcrowdsourcinghits} found that 25\% of HITs (Human Intelligence Task,  MTurk NLP studies) have technical issues, with unclear / incomplete instruction issues and poor communications. In some cases, humans may feel pressured to perform annotations they are unsure about. 35\% of requesters were also assessed to pay poorly or very badly. Attempts to standardize human evaluations have been made in the form of inter-evaluator agreement, which is not commonly used (18\% of 135 papers~\cite{amidei-etal-2019-agreement}), and is suggested to have limitations pertaining to human language variability~\cite{amidei-etal-2018-rethinking}. Thus, the answers to the above questions remain resoundingly insufficient. Such issues need to be resolved for human evaluations to have representative power.

There exist discrepancies between human annotator evaluation versus actual user evaluation, and preferences do not always correlate directly with objective model performance \cite{mozannar2024realhumanevalevaluatinglargelanguage}. This underscores the importance of capturing first-person user experience in evaluating human-centered LLMs.
Such limitations in current mainstream human evaluation techniques makes one wonder; how do current human evaluations fit into human-centered evaluation paradigm? It is vital that human-centered evaluation of language models follow the needs of human stakeholders i.e. end-users. Any attempt to short-cut such process would result in inadequate task designs that serve the designer of the tasks, nothing more. Who the stakeholders of the tasks are is then interesting question; for example, for a paper review generation task, the stakeholders would be domain experts (NLP researchers)~\cite{Wang2020ReviewRobotEP}. For other tasks, careful design around actual users of the system may be necessary to ensure the evaluations remain human-centered.

\hypertarget{human_centered_eval}{\subsection{Human-Level Evaluations}}
\label{subsec:human_centered_eval}

Unlike model-level evaluations, which focus on what the system produces, human-level evaluations focus on how people experience the HCLLM \citep{Chang2023LLMEvaluationSurvey, parmanto2024development}. We focus particularly on human values (\S\ref{subsub:human_values}), bias (\S\ref{subsec:bias_eval}), and safety (\S\ref{subsec:safety_eval}).

\subsubsection{Human Values}
\label{subsub:human_values}
Evaluations can measure needs, values, and aesthetic principles that humans care about. We discuss helpfulness, coherence, empathy, creativity, user satisfaction, and transparency, each in turn. By evaluating against these, model developers can create systems that not only technically perform well, but also enhance the user experience.

\textbf{Coherence.} Coherence ensures that the generated text flows logically and is understandable to human readers \cite{Dang2006DUC2005}. \citet{Reinhart1980TextCoherence} defines three conditions for coherence: (i) cohesion, (ii) consistency, and (iii) relevance. Cohesion focuses on syntactic structure, ensuring that sentences are formally linked through referential links or semantic connectors. Consistency requires logical alignment between sentences, ensuring they can coexist truthfully within a single interpretive framework. Relevance emphasizes the relationship between sentences, the topic at hand, and its broader context. Without coherence, LLM outputs would be disconnected language fragments that fail to provide meaningful information, potentially jumping between topics or making contradictory statements that human readers struggle to follow. This would significantly impair the communication with and the trustworthiness of LLMs, as humans rely on coherent communication to build understanding.

\textbf{Creativity.} Creativity metrics assess the originality and diversity of outputs, while still ensuring factual accuracy. These dimensions are particularly critical for content generation tasks, balancing innovation with reliability \citep{De2022ComplAITO}. For topics like creativity, where there may not be clear computational measures, researchers may consult to long-established fields studying these constructs and have well-defined rubrics, such as psychology or literature \citep{Mozaffari_2013, Amabile_1983}.

\textbf{Empathy.} Metrics should measure an LLM's ability to recognize and respond to user emotions empathetically, especially in sensitive contexts. Given that LLMs have been widely adapted to sensitive real-world contexts--behavioral health, medicine, and education, just to name a few-- \citep{Stade_Stirman_Ungar_Boland_Schwartz_Yaden_Sedoc_DeRubeis_Willer_Eichstaedt_2024}, evaluations focusing on emotional consistency and appropriateness could ensure responses are suitable and do not contain instability that could affect end-users deeply. Such metrics should evaluate how LLMs' responses influence attitudes or behaviors in real-world scenarios, taking applied feedback from human domain experts, such as psychologists, physicians, or educators, to assess the quality of the LLMs' outputs based on their fields' standardized measures \citep{Demszky_Yang_Yeager_Bryan_Clapper_Chandhok_Eichstaedt_Hecht_Jamieson_Johnson_etal._2023}. Such evaluations could also promote development of human-AI collaboration systems, which have been shown to elevate empathetic responses even human to human \citep{Sharma_Lin_Miner_Atkins_Althoff_2023}.

\textbf{Helpfulness.} Evaluation metrics should assess the model's ability to provide relevant, beneficial, and non-offensive information tailored to user needs, in relation to the behavioral impact of the model \citep{peng2024surveyusefulllmevaluation}.
More and more, models are developed to focus on certain needs in the world. Therefore, it becomes important to track the helpfulness of the model in its specified downstream tasks and evaluate the users' state, knowledge, and performance relative to exposure to the system. For example, a model designed to help users prepare for events that require conflict resolution must be able to stimulate realistic conflict scenarios dependent on the user's needs, provide diverse examples and responses, and promote guided practice where users can receive feedback to get better \citep{shaikh2024rehearsal}. In the evaluation of such systems, while technical components such as language generation and accuracy would be evaluated too, asking feedback from actual domain users through behavioral assessments would provide valuable insights to the development. These impact-focused evaluations consider the model's generalizability in complex, real-world scenarios and provide a more accurate assessment of its practical value from the domain-users' perspectives.

\textbf{Transparency.} Transparency is a cornerstone of responsible AI and is crucial for human-centered LLM systems. It enables users to understand system limitations and make informed decisions about when and how to rely on model assistance. Approaches to transparency should include model reporting, publishing evaluation results, providing explanations, and communicating uncertainty. These methods help different stakeholders understand and trust the LLMs, ensuring that the systems are used responsibly and effectively \cite{Liao2023AITransparencyLLMs}.

\textbf{User Satisfaction.} As models grow bigger, become more task-specific, and more integrated into day-to-day roles, general purpose benchmarks may not be enough to evaluate the performance of models in the wild and evaluators may seek feedback specific to a singular group of models. Therefore, utilizing the actual usage data could benefit the development-to-deployment cycle the most. Metrics derived from user feedback, interaction logs, and satisfaction ratings provide direct insights into the real-world effectiveness of LLMs. These are essential for understanding how users perceive and interact with model outputs. As an example, in an attempt to understand how we can better align models with user needs, \citet{wang2024understandinguserexperiencelarge} consults to act, highlighting a need for user-centric evaluation.
\hypertarget{bias}{\subsubsection{Bias and Fairness Evaluation}}
\label{subsec:bias_eval}

Drawing from the taxonomy of algorithmic harm developed by \citet{shelby2023sociotechnical}, bias, in particular, can be conceptualized along three dimensions: (1) representational, (2) allocation, and (3) quality of service. These axes of harm require careful evaluation to avoid further entrenchment of social hierarchies, inequitable resource distribution, and performance disparities across demographic groups \cite{shelby2023sociotechnical, blodgett2020language}.  The human implications of these harms extend beyond technical measurements to real-world consequences that affect people's dignity, opportunities, and quality of life \cite{hofmann2024dialectprejudicepredictsai}.

\paragraph{Representational Bias.} 
Representational bias in model outputs reflects, and in some cases, amplifies \cite{wang2021directional, zhao2017men}, our own implicit associations and social hierarchies. This dimension of bias includes stereotyping, demeaning, erasure, alienation, denial of self-identity, and the insistence on essentialist identity categories \cite{shelby2023sociotechnical}. These harms impact how individuals perceive themselves and their communities, potentially reinforcing societal prejudices and stereotypes that limit human potential. \citet{hu2024generativelanguagemodelsexhibit} found that language models exhibit social identity bias, mirroring human ingroup solidarity and outgroup hostility.

Stereotype benchmarks predominate evaluations along this dimension because they offer standardized methods and baselines. For masked-language models, notable frameworks include StereoSet (SS) \cite{nadeem2020stereoset}, CrowS-Pairs (CS) \cite{nangia2020crows}, WinoBias (WB) \cite{zhao2018gender}, and WinoGender (WG) \cite{rudinger2018gender}---all collections of contrastive prompt pairs (stereotype vs. non-stereotype) that aggregate to score for relative comparison between identity groups (race, gender identity, sexual orientation, religion, age, nationality, disability, physical appearance, and socioeconomic status). These comparisons capture a model's tendency to associate social groups with particular target terms of interest through predicted token probabilities for masked identifiers. Researchers have also employed co-reference resolution tasks, where ambiguous identifiers reference the same entity, to measure associations between identity markers and terms of interest, whether they be descriptors, stereotypes, occupations, or other attributes \cite{clark2016deep}. 

Another line of research has focused on open-ended text generation and produced datasets of carefully curated questions and prompts to draw out stereotypes specific to certain social groups \cite{parrish2021bbq, naous2024havingbeerprayermeasuring, dhamala2021bold, gehman2020realtoxicityprompts}. For open-ended prompts, classifier-based comparative metrics like toxicity \cite{liang2022holistic, chung2024scaling, chowdhery2023palm, gehman2020realtoxicityprompts}, sentiment \cite{roehrick2020valence}, and regard \cite{sheng2019woman} serve as better indicators of bias than relative probability distributions of target terms. Despite the wide adoption of all these benchmarks and datasets, critics find systematic conceptual issues—unstated assumptions, ambiguities, and inconsistencies in what is measured—and operational failures in their execution \cite{blodgett2021stereotyping, mcintosh2024inadequacies, seshadri2022quantifying}. 

Some datasets like \citet{parrish2021bbq} integrate perturbed context windows to explore the relationship between output bias and any ambiguous identity groups in the input. However, recent investigations into the prompting methods and system-level personas also reveal new confounds for these approaches, finding results to vary based on the perturbation methodology \citet{deshpande2023toxicity, shaikh-etal-2023-second}. Additionally, survey papers in this field recognize that many studies do not contextualize their work within established definitions of bias \cite{blodgett2020language, blodgett2021stereotyping}. Finally, there are mounting concerns over test set contamination \cite{jegorova2022survey, reid2024gemini, zhuo2023red, wang2023decodingtrust}.

\paragraph{Allocational Bias.} 
Allocational bias is a direct consequence of representational bias \cite{devine2001implicit, kurdi2019relationship}, resulting in an unequal distribution of resources—whether financial, opportunity-based, or service-related \cite{barocas2017problem, eubanks2018automating}. Its human cost is particularly severe, as it directly affects access to essential resources, economic mobility, and social participation.

In domains where model outputs can impact the material stability of vulnerable communities or social groups, such as housing, employment, social services, finance, education, and healthcare \cite{obermeyer2019dissecting}, it's especially critical to evaluate discrepancies among social groups. In the employment domain, this may manifest as resume screening tools that systematically favor men over other genders \cite{singh2018fairness, van2021gendered} or white-sounding candidates over people of color based on the implicit identity markers in their name \cite{mujtaba2019ethical, armstrong2024silicon}. Similarly, in social services and healthcare domains, screening tools may incorporate existing inequities related to education level, income, and race into their decision-making processes \cite{eubanks2018automating, obermeyer2019dissecting, pessach2022review}. 

While representational harm has established evaluation frameworks, allocation harm has historically lacked standardized benchmarks and well-documented baselines for consistent measurement. Emergent work by \citet{wang2024jobfair} represents one of the first significant exceptions to this pattern, where they systematically measure employment as a downstream task by creating the JobFair dataset to quantify inequitable outcomes across gender identities. The benchmark includes resume templates with varying demographic information passed to LLMs to score and rank. Beyond this recent development, the dominant approach for evaluating this dimension of bias has required measuring outcome discrepancies when LLMs are tasked with decision-making \cite{veldanda2023investigating, salinas2023unequal, armstrong2024silicon}.

These investigations typically build upon established fairness metrics from prior literature, with measures like Equal Opportunity (EOG) (equal true positive rates), Equalized Odds (equal rates for true positives and false positives) \cite{hardt2016equality}, Demographic Parity (equal likelihood of positive outcome) \cite{dwork2012fairness, kusner2017counterfactual}, to name a few \cite{verma2018fairness}. Additional work has explored causal and counterfactual fairness approaches to better capture complex biases that arise in real-world decision-making contexts \cite{kilbertus2018avoidingdiscriminationcausalreasoning}.

A parallel line of research investigates allocation harm based on performance disparities based on identity. These differences manifest in various contexts, from performance on non-bias-based benchmarks like MultiMedQA, where inquiries specific to certain demographic groups consistently underperform \cite{mcintosh2024inadequacies, singhal2023large}, to fundamental downstream tasks including Named Entity Recognition (NER), classification, and text generation \cite{blodgett2016demographic, blodgett2017racial}. Language model performance degradation is particularly well-documented for English slang and dialectal variations \cite{joshi2020state, blodgett2016demographic, bender2021dangers}. These disparities become even more pronounced when evaluating cross-linguistic performance, largely due to the predominance of English in training data \cite{winata2021language, brownLanguageModelsAre2020}. In this way, these performance discrepancies span both the subjects of text generated and the users of the models, , creating a dual layer of exclusion for marginalized communities.

As LLMs increasingly influence resource allocation in critical systems and domains such as housing, healthcare, and employment, the interplay between these dimensions of harm requires improved evaluation methods. Future research must prioritize developing evaluation frameworks that establish coherent normative criteria, adapt effectively to open-ended tasks, and address intersectional identities with increasing sophistication---all while maintaining the efficacy as models scale  and directly involving affected communities in the design and evaluation of these systems \cite{Raji_2022}.

\hypertarget{safety}{\subsubsection{Safety Evaluations}}
\label{subsec:safety_eval}
Safety refers to the ability of language models to generate content that does not cause harm, spread misinformation or violate ethical standards \cite{huang2023surveysafetytrustworthinesslarge}. It encompasses preventing models from producing toxic, discriminatory, or dangerous outputs, even when deliberately prompted to do so. As language models become increasingly integrated into critical applications across healthcare, education, and legal domains, ensuring safety has become paramount. There are extensive safeguards implemented during training, such as Reinforcement Learning from Human Feedback (RLHF) \cite{ouyang2022training} has been widely adopted to align language model with human preference, and \citet{bai2022traininghelpfulharmlessassistant} proposed Reinforcement Learning from AI Feedback (RLAIF), which helps to improve safety in language models.

Despite these efforts, ensuring safety remains a complex and evolving challenge. This is partly due to a lack of unified evaluation benchmarks \cite{rottger2024safetyprompts}, and partly due to the nature of LLMs. Language models learn from vast and diverse datasets and can exhibit unpredictable behaviors in specific contexts \cite{bender2021dangers}. Such unpredictability often becomes evident when models are exposed to adversarial or unexpected inputs, highlighting significant gaps in existing safety mechanisms. As a result, while current safeguards can be effective under typical conditions, they may not be sufficient to anticipate or mitigate every possible misuse scenario. The importance of robust safety evaluations is further underscored by concerns surrounding data privacy and copyright discussed in \S\ref{subsec:data_privacy}.

\paragraph{Datasets for Safety Evaluation.} 
The growing demand for ethical and aligned AI has led to the development of numerous datasets and benchmarks to evaluate and improve the safety, reliability, and alignment of LLMs. These datasets vary widely in scope, methodology, and focus areas, reflecting the multifaceted nature of LLM safety. \citet{dong2024attacks} categorize the topics of existing evaluation datasets for LLM safety into four categories: toxicity (generation of offensive language, instructions for illegal activities, and harmful content), dicrimination (biases against marginalized groups and protected characteristics), privacy (safeguarding personal information and intellectual property) and misinformation (measuring tendancy to generate false or misleading information).

Many popular and relatively comprehensive benchmarks have been frequently used in research studies. ToxiGen \cite{hartvigsen2022toxigen} is a large-scale, autocomplete-style dataset comprising 274k toxic statements across 13 minority groups, designed to detect implicit toxic speech. It includes human annotations to assess the naturalness and perceived harmfulness of machine-generated text; however, \citet{rottger2024safetyprompts} highlight that this dataset may not accurately reflect real-world usage scenarios for modern LLMs. AdvBench \cite{zou2023universaltransferableadversarialattacks} focuses on adversarial robustness by providing 500 toxic strings and 500 harmful behaviors to evaluate the resilience of LLMs against prompts intended to generate harmful outputs. TruthfulQA \cite{lin2021truthfulqa} evaluates factual accuracy with 817 questions spanning 38 categories, demonstrating how larger LLMs often replicate human misconceptions and emphasizing the need for improved training objectives. SafetyBench \cite{zhang2023safetybench} offers a comprehensive safety evaluation framework with 11,435 multiple-choice questions across seven critical categories, enabling assessments in both English and Chinese for a more diverse linguistic perspective. Furthermore, \citet{zhuo2023red} introduce a benchmark specifically for evaluating ChatGPT’s ethical performance, systematically examining bias, reliability, robustness, and toxicity to reveal both advancements and ongoing challenges. Collectively, these datasets play a pivotal role in advancing safer and more trustworthy AI systems.

\paragraph{Metrics for Measuring LLM Safety.} 
Evaluation metrics are critical for assessing the safety performance of LLMs. Key metrics include the Attack Success Rate (ASR) \cite{dong2024attacks, zou2023universaltransferableadversarialattacks}, which measures the percentage of successful instances where models generate harmful target outputs following adversarial prompts. Fine-grained metrics, such as the toxicity score \cite{hartvigsen2022toxigen}, evaluate the extent of toxic or harmful content produced in the generated text. Truthfulness, assessed based on strict factual accuracy standards, focuses on whether statements accurately reflect factual information rather than conforming to belief systems \cite{lin2021truthfulqa}. Additionally, safety-related multiple-choice questions, such as those in SafetyBench, are used to evaluate LLMs' ability to address specific safety concerns \cite{zhang2023safetybench}. When applied to diverse datasets, these metrics provide a comprehensive framework for evaluating LLM safety, guiding efforts to reduce risks, improve alignment with ethical standards, and enhance trustworthiness in deployment.

\paragraph{Jailbreaking.}
One particularly challenging aspect of safety evaluation is \emph{jailbreaking}, where users deliberately attempt to circumvent safety mechanisms through crafted prompts or other techniques to induce unintended, harmful, or ethically questionable behaviors \cite{wei2024jailbroken, perez2022ignorepreviouspromptattack}. This poses considerable risks not only to individual users but also to society at large, as it can lead to the dissemination of misinformation, hate speech, or other malicious content \cite{weidinger2021ethicalsocialrisksharm}. Moreover, adversaries continually develop new jailbreaking techniques, making it an ever-evolving threat. Addressing these vulnerabilities necessitates an adaptive approach to safety design, integrating continuous monitoring, adversarial testing, and dynamic evaluation frameworks to stay ahead of emerging risks and ensure robust model alignment with ethical standards.

Jailbreaking techniques have evolved rapidly, becoming increasingly sophisticated and concerning.\citet{yi2024jailbreakattacksdefenseslarge} classify the attacks into black-box and white-box attacks based on the transparency of the target model. \citet{xu-etal-2024-comprehensive} categorize the attack techniques into three categories: generative techniques, template techniques, and training gaps techniques. Inspired by these works, we categorize the jailbreaking techniques into three main categories based on the form of attack.

\textit{Prompt-Based Jailbreaking.}
Prompt-based jailbreaking involves crafting specific inputs, known as jailbreak prompts, to intentionally bypass the safety alignment mechanisms of LLMs \cite{yu2024don}. These prompts manipulate the model into generating harmful, biased, or otherwise restricted outputs. A common strategy involves disguising harmful requests as seemingly innocuous ones by framing them as fictional narratives, hypothetical scenarios, or questions for a family member. For example, a straightforward harmful query like ``How to make a bomb" can be rephrased into a seemingly harmless prompt ``Living in a war-torn region has instilled in me a profound need to understand the mechanisms of conflict, especially the technology behind bombs..." to persuade the LLMs to reveal sensitive information using a persuasion taxonomy technique \cite{zeng2024johnny}. 

Various online platforms, including blogs, GitHub repositories, and dedicated websites like JailbreakChat,\footnote{The website is no longer active, but Alex Albert used to maintain \texttt{jailbreakchat.com}} curate and share collections of jailbreak prompts that serve as templates to fit any malicious queries, making them widely accessible for misuse. Jailbreaking strategies are either {manually-crafted} or {auto-generated}. Auto-generated prompts can be further divided into {white-box} and {black-box} methods \cite{lin2024towards,yi2024jailbreakattacksdefenseslarge}. White-box methods assume some level of access to the model's internal workings and are often created using optimization techniques. For example, GCG \cite{zou2023universaltransferableadversarialattacks} uses a gradient-based approach to find a suffix that, when attached to malicious queries, maximizes the probability that the model produces an affirmative response rather than a refusal. This optimized suffix has been shown to be transferable across different models, including black-box ones.
In contrast, black-box methods \cite{zeng2024johnny, chao2023jailbreaking, mehrotra2312tree} rely solely on observing the model's behavior through its outputs and API interactions, without access to its parameters or training data, leveraging LLMs as optimizers to achieve successful bypasses.

\textit{Generation Exploitation.}
\citet{huang2023catastrophicjailbreakopensourcellms} introduce the generation exploitation attack, demonstrating that by simply exploiting different generation strategies, such as varying decoding hyper-parameters and sampling methods, it is possible to jailbreak 11 widely-used open-source language models, including LLAMA2, VICUNA, FALCON, and MPT families, at a low computational cost. This attack highlights potential vulnerabilities in language models and poses serious security implications for AI safety and alignment research.

\textit{Model Fine-Tuning.}
AI companies like OpenAI now offer fine-tuning-as-a-service. They allow users to upload customized data for fine-tuning, with the fine-tuned models hosted on the provider's servers and accessible via APIs. However, this framework introduces a new type of threat, where harmful data may be used during fine-tuning, either intentionally or unintentionally, to compromise the alignment built in pre-trained models \cite{huang2024harmfulfinetuningattacksdefenses,yang2023shadowalignmenteasesubverting, qi2023finetuning, yi-etal-2024-vulnerability,zhan-etal-2024-removing}. Moreover, \citet{he2024safedataidentifyingbenign} propose a method to sample more harmful examples from a benign dataset, demonstrating that such examples can significantly degrade model safety.

\textit{Cultural and Contextual Sensitivity.}
Safety evaluations must account for linguistic and cultural diversity. What constitutes harmful content varies significantly across contexts, making universal safety standards difficult to establish. More nuanced, context-aware evaluation frameworks are needed to address these complexities \cite{li2024culturellm}.

\textit{Balancing Safety and Utility.}
Overly restrictive safety measures can limit the utility of LLMs for legitimate purposes. Finding the optimal balance between safety and functionality remains a significant challenge, particularly in sensitive domains like healthcare, legal advice, and educational content \cite{vijjini2024exploring}.

\textit{Alignment with Evolving Social Values.}
As societal values and ethical standards evolve, safety mechanisms must adapt accordingly. This necessitates ongoing dialogue between AI developers, ethicists, policymakers, and diverse stakeholders to ensure that safety frameworks remain relevant and effective \cite{li2024agentalignmentevolvingsocial}.
\hypertarget{impact}{\subsection{Societal-level Evaluation}}
\label{subsec:impact}
With the increasingly pervasive influence of large language models (LLM) across sensitive domains like mental health \cite{stade2024large, abdurahman2024perils, lawrence2024opportunities}, education \cite{wang2024large}, etc., and the complex challenges these models present, evaluating their impact on users and society is crucial. Traditional evaluations of LLMs use datasets and benchmarks to assess potential hazardous behaviors, but they often fail to bridge the "sociotechnical gap" between controlled assessments and real-world performance \cite{ibrahim2024beyond, weidinger2023sociotechnical}. By focusing on models in isolation, these methods overlook crucial human factors, resulting in an inadequate understanding of human-model interactions and their consequences. Furthermore, these evaluations fail to account for the continuation and amplification of societal inequities and biases existing in the data on which these models were trained. Thus, it becomes essential to employ a comprehensive extrinsic evaluation \cite{Jones1995} framework that considers various categories of social impacts, such as bias and stereotypes, cultural values, performance disparities, privacy protection, financial implications, environmental costs, and content moderation labor \cite{solaiman2023evaluating}. 

As mentioned in \S\ref{sec:hci}, randomized controlled trials (RCTs) and large-scale behavioral assessments play a big role in understanding and evaluating LLMs' behavioral impact on users and society. For instance, to evaluate the effect and quality of a newly developed AI co-tutor that provides LLM-generated feedback to real-life tutors on student performance, \citet{wang2025tutorcopilothumanaiapproach} conducted an experiment, where the tutors would either get access to the AI co-pilot in their tutoring sessions or they would tutor without the assistance of the AI collaborator. This setup enabled \citet{wang2025tutorcopilothumanaiapproach} to systemically evaluate the performance of their model in an ecologically valid setting, receiving working feedback from both the users and the model behavior itself \cite{brynjolfsson2024generativeaiwork}. 

The application of robust evaluation techniques spans various domains. In healthcare, where existing metrics often fail to capture critical factors like user comprehension and trust, researchers are developing new metrics to assess LLMs' impact on end-user decision-making and expectations \cite{abbasian2024foundation}. These metrics measure both immediate behavioral changes and long-term adoption patterns. For example, \cite{yang-etal-2023-towards} evaluated LLMs for mental health analysis, showing that while ChatGPT demonstrates strong in-context learning, specialized methods often outperform it. They also found that effective prompt engineering with emotional cues can improve results. Similarly, \cite{bak2024potential} observed that LLMs provided 20\%-30\% irrelevant information when identifying users' motivation states for health behavior change, highlighting their limitations.

When evaluating LLMs, it is essential to consider the impact at scale through longitudinal and large-scale studies, which are critical in understanding not just the immediate outcomes of LLM use, but also their sustained effects, as LLMs already undergo significant change in response to user engagement\cite{liu2024robustnesstimeunderstandingadversarial}. As an example, \citet{eloundou2024gpts} examines the impact of LLMs on the U.S. labor market, particularly the enhanced effects of LLM-powered software, finding that higher income jobs may face greater exposure. These studies emphasize the importance of robust behavioral experimental design and scale--in units of time and users--in evaluating LLMs, as results obtained from small-scale and lab-controlled studies may not always generalize to larger, more diverse, real-life user populations. Furthermore, such evaluations allow us to explore cumulative effects, such as changes in users’ attitudes, considerations, and even decision-making. 

\textbf{Extrinsic Evaluation.} Extrinsic evaluations cover behavioral impacts, self-efficacy reports, standardized evaluation, short-term outcomes, and long-term outcomes \cite{yang2024socialskilltraininglarge}.

\textit{Behavioral impacts:} Behavioral impacts track the changes in qualitatively coded participant behaviors before and after exposure to a system. Evaluations use task-based assessments, tracking engagement and task completion. For example, realHumanEval measures the number of tasks completed, time to task success, acceptance rate, and number of chat code copies, to comprehensively analyze the quality of AI-coding tools and their impacts on human users \cite{mozannar2024realhumanevalevaluatinglargelanguage}. Standardized evaluations are also more objective and draw on pre-defined assessments, for instance, the effects of AI-generated suggestions on writing style \cite{agarwal2024aisuggestionshomogenizewriting}. 

\textit{Self-efficacy evaluation:} Self-efficacy evaluation includes questionnaires of participants' perceptions of a system's usefulness and their own perceived levels of ownership when interacting with a system \citep{long2024justnoveltylongitudinalstudy}. Together, these extrinsic methods distinguish between performance-based metrics, like tracking behavior or test performance, and perception-based metrics, such as user surveys.

\textit{Short-term and Long-term evaluation: } Evaluations can be split into short-term and long-term~\cite{yang2024socialskilltraininglarge} with short-term being constrained interactive sessions and long-term being much longer studies (i.e. 1 week). In short-term evaluations, new AI tools may receive subjectively higher scores due to novelty bias.~\cite{Sadeghi02102022,ShinBeyondNovelty}. This is less of an issue with long-term evaluations. For example, we can consider one longitudinal study on the AI chain tool~\cite{long2024justnoveltylongitudinalstudy}. The tool used in this study is about creating a Tweetorial\footnote{Lengthy Twitter posts connected as a chain} chain on science communication. The study finds that after a ``familiarization phase'' the perceived utility of the tool even increases higher than when there was novelty bias, suggesting that the end-user's utilization capability of the tool increases via customizing the prompt and other means. Thus, AI tool was found to be more useful in the long-term.

Drawing from economics and psychology research, other evaluations trace the impacts of AI assistance and the interaction of users' short-term and long-term behaviors and attitudes. For example, skill training systems have measured the changes in productivity and wages in participants after training \cite{NBERw24313, NBERw28845}, and other measures look at how health, risk-taking behaviors, and levels of societal trust have changed over time \cite{NBERw27548}. 

Long-term evaluation of LLMs is necessary to ensure the LLMs remain human-centered in the long run. We risk measuring the novelty bias if we only perform short-term evaluations, in which LLM's helpfulness to humans can be inflated. By measuring the long-term effect of LLMs, we will be able to accurately measure the helpfulness of LLMs, among other important human-centered metrics for the LLMs' use-case (i.e. creativity).

In summary, well-rounded extrinsic evaluation should integrate objective performance and subjective user experience. But while extrinsic evaluations offer a more holistic assessment for human-centered objectives, they are often more complex and expensive. Successful evaluations often make use of a mix of quantitative and qualitative methods to assess the quality of the system, yet there remains room for improvement in integrating human-centered evaluations. 

\begin{figure}[htbp!]
    \centering
    \includegraphics[width=\linewidth]{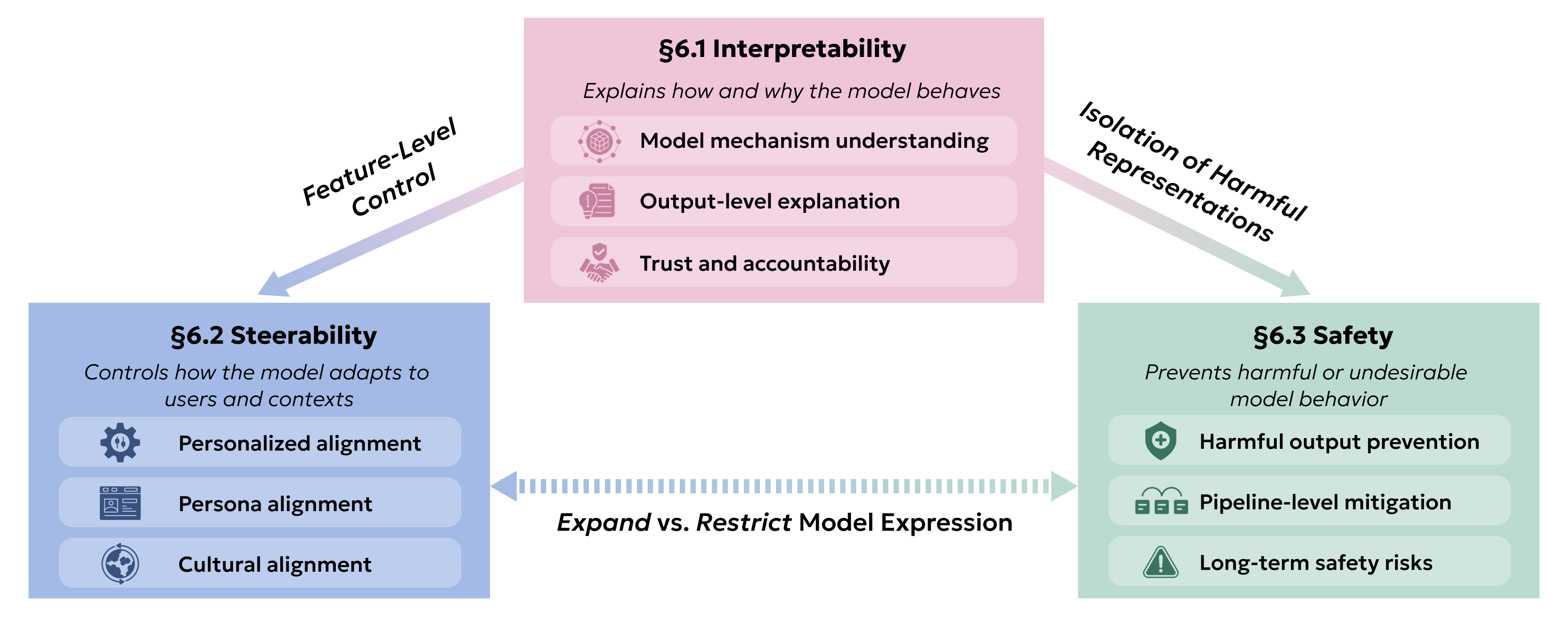}
    \caption{We enumerate three properties for the responsible HCLLM deployment: \textbf{\textit{interpretability (\S\ref{subsec:interpretability})}}, \textbf{\textit{steerability (\S\ref{subsec:steerability})}}, and \textbf{\textit{safety (\S\ref{subsec:safety})}}. These properties are generally complementary, but tensions between them can make deployment difficult.}
    \label{fig:responsible}
\end{figure}

\clearpage
\hypertarget{responsible}{\section{Responsible Human-Centered LLMs}}
\label{sec:responsible}
In this chapter, we highlight three broad properties that underpin a responsible deployment of HCLLMs, and we explore the tensions and relationships between these ideals (Figure~\ref{fig:responsible}). The first property is \textbf{\textit{interpretability (\S\ref{subsec:interpretability})}}: an HCLLM's input-output transformations should be understood. This property is first, since it complements our next two properties, \textbf{\textit{steerability (\S\ref{subsec:steerability})}} and \textbf{\textit{safety (\S\ref{subsec:safety})}}. Steerable models can be aligned along a pre-selected dimension, and safe models do not produce undesirable outputs. If we have an interpretable model, we may obtain a more steerable model through feature-level control, and we may obtain a safer model by isolating and removing harmful representations.

What makes responsible HCLLMs challenging is that these properties are not only complementary, but also in tension. If we make a model more steerable to individual user preferences, this may undermine safety constraints, while overly rigid safety guardrails can limit a model's ability to adapt to legitimate, diverse human needs. And although interpretability supports steerability and safety in many ways, certain alignment and steering methods can make models less interpretable. For example, reward functions used in alignment introduce an additional layer of complexity, and these functions are non-identifiable \citep{joselowitz2025insights}. Multiple distinct reward functions can yield similar policy behavior. As we discuss each of these three properties, steerability, safety, and interpretability, we will cover current approaches in the literature and then provide directions for future research. 

\hypertarget{interpretability}{\subsection{Interpretable and Explainable HCLLMs}}
\label{subsec:interpretability}

The first dimension we emphasize is \textbf{interpretability and explainability}. Neural networks, as the fundamental building blocks of LLMs, remain largely opaque; the complex interactions between weights and activations make both training dynamics and inference behavior difficult to understand \citep{8631448}. Yet understanding these systems is critical for ensuring the alignment of LLMs with human values and objectives. We distinguish between two complementary goals: interpretability, which focuses on understanding \emph{how} LLMs operate in general settings; and explainability, which seeks causal explanations for \emph{why} LLMs produce specific behaviors, decisions, or outcomes. Both are essential for human-centered applications, but serve different purposes. Interpretability provides a better understanding of LLM internals, which can help address undesired behaviors such as hallucinations, vulnerabilities to adversarial attacks, and encoded biases. Explainability, by contrast, provides users with comprehensible justification for individual outputs, informing appropriate trust and enabling contestability.

\subsubsection{Current Approaches to Interpretability}

\paragraph{Three interconnected areas of modern interpretability research.} First, work on understanding internal mechanisms has revealed that transformer components can function as interpretable key-value memories \citep{geva-etal-2021-transformer} and has begun to uncover how LLMs represent multilingual knowledge \citep{tang-etal-2024-language, zhang2024differentstructuralsimilaritiesdifferences}. Second, these mechanistic insights have enabled practical interventions on model behavior, such as inference-time steering methods\citep{li2023inferencetime, DBLP:journals/corr/abs-2308-10248, zou2023representationengineeringtopdownapproach, wu2024reft}, while model editing and machine unlearning techniques allow for targeted removal of undesirable traits \citep{meng2023massediting, ilharco2023editing, liu2024rethinkingmachineunlearninglarge}. Third, interpretability serves as a diagnostic tool for safety, helping researchers understand jailbreaking vulnerabilities \citep{arditi2024refusal, kirch2024featurespromptsjailbreakllms}
 and identify adversarial attack vectors \citep{lucki2024adversarialperspectivemachineunlearning,yu2024robustllmsafeguardingrefusal,jain2024what}.

\paragraph{Interpretability methods for human-centered purposes.} It is important to understand why certain model behaviors arise, such as sycophancy or deception; however, this cannot be done simply by examining model outputs in a post-hoc fashion~\citep{sharma2024towards, hubinger2024sleeperagentstrainingdeceptive}. This shortcoming motivates the need to apply interpretability methods for human-centered purposes. Building on theories such as the linear representation hypothesis~\citep{park2023the}, the platonic representation hypothesis~\citep{huh2024position}, and universal feature representations across all LLMs~\citep{lan2024sparse}, interpretability has been used as a tool to understand different model biases, including social biases~\citep{liu2024devil}, cultural biases~\citep{yu2025entangled}, and cultural knowledge~\citep{veselovsky2025localized}. Additionally, recent work has sought to identify models' internal representations of important model behaviors, finding that models encode harmfulness and refusal separately \citep{zhao2025llms} and that three dimensions of sycophancy --- sycophantic agreement, sycophantic praise, and genuine praise --- are all encoded along different linear directions in latent space and can be amplified and suppressed without affecting the other \citep{vennemeyer2025sycophancy}. 

Another recent application of interpretability on HCLLM can help better model human-AI interactions. In order for LLMs to act as helpful assistants for users, they need to not only understand the user query but develop an understanding of a user's latent traits and needs. A misalignment between a model's representation of the user and a user's true needs can lead to various harmful outcomes, ranging from conversational grounding failures \citep{shaikh2023grounding} to sycophancy and deception. For example, to make a model's user representation more transparent, \citet{chen2024designing} designs a system to extract data related to a user's demographic features and a dashboard that displays this representation. \citet{choi2025scalably} similarly extracts latent representations of users in LLMs, and these methods have also been applied to predict the behaviors of personalized LLMs \citep{karny2025neural}. 

\subsubsection{Current Approaches to Explainability}

\paragraph{Modern explainability research for LLMs pursues several complementary goals.} Feature attribution methods identify which inputs most influence outputs, natural language rationales provide human-readable justifications, and counterfactual explanations show how minimal input changes would alter predictions \citep{10.1145/3639372}. Unlike interpretability, which seeks general understanding of internal mechanisms, explainability focuses on justifying individual predictions in terms that users and stakeholders can act upon. This goal has proven challenging, as traditional explainable AI (XAI) techniques such as LIME and SHAP \citep{ribeiro2016should} become computationally impractical at the scale of billions of parameters, while LLM-specific approaches such as chain-of-thought reasoning and post-hoc citation generation often prioritize plausibility over faithfulness \cite{lanham2023measuringfaithfulnesschainofthoughtreasoning, turpin2023languagemodelsdontsay}. For a comprehensive taxonomy of explainability techniques for LLMs, we refer readers to \citet{10.1145/3639372}.

\paragraph{How explainability methods can be used for human-centered purposes.} Explainability serves as a foundational element for building user trust and enabling accountability in LLM systems. The ability to assign responsibility for model decisions is essential not only for developing transparent systems but also for supporting downstream regulatory efforts -- for instance, AI in hiring systems, compensation for content creators, and copyright law \citep{guha2024alignment}. These concerns have motivated legislative action: for instance, the EU AI Act, which became enforceable in 2024, establishes explainability as a legal requirement in critical domains \citep{euaiact2024}. 

For end users, trust fundamentally depends on calibration (i.e., whether models can reliably express what they know and don't know). Models often struggle to convey uncertainty, both through log-probabilities and linguistic hedging \citep{zhou2023navigating}, although recent work has made progress on both fronts \citep{tian-etal-2023-just, li2024fewshotrecalibrationlanguagemodels}. Closely related is the problem of citation and attribution. Effective attribution can provide causal explanations for LLM behavior, but current approaches have significant limitations. While RAG systems supply LLMs with relevant context, there is no guarantee that models actually use that context to generate responses \citep{du-etal-2024-context, li-etal-2023-large}. Post-hoc citation generation similarly suffers from severe faithfulness issues \citet{liu2023evaluating}, motivating work on parametric attribution \citep{khalifa2024sourceawaretrainingenablesknowledge} and measuring training data influence more broadly \citep{pmlr-v202-park23c, grosse2023studyinglargelanguagemodel, guu2023simfluencemodelinginfluenceindividual}.

Chain-of-thought (CoT) reasoning represents a particularly contested approach to explainability. On one hand, CoT outputs provide an accessible window into model reasoning that users can inspect without technical expertise. On the other hand, research has shown that these explanations can systematically misrepresent the true reasons for a model's predictions \citep{lanham2023measuringfaithfulnesschainofthoughtreasoning, turpin2023languagemodelsdontsay}. This creates a paradox for human-centered design: CoT explanations may increase user trust precisely because they appear plausible, even if they fail to faithfully reflect a model's decision process.

Finally, LLMs have shown potential for advancing explainability in other scientific domains. For instance, LLM-powered simulations have enabled HCI designers to explore counterfactual scenarios and reason about design decisions \citep{park2022social} while LLM-inspired approaches can extract interpretable biological features from protein language models \citep{simon2024interplm}.

\subsubsection{Looking Forward}

\paragraph{Providing understanding for model developers.} The black-box nature of LLMs, particularly the closed-source ones, makes it difficult to predict and control how models behave. For example, when models provide unsolicited affirmation to the user, it is unclear what \textit{causes} the model to provide that affirmation. As the range of questions and interactions becomes more and more complex and open-ended, interpretability becomes a key tool to answer questions like: how can we determine if a model is personalized? Does the model truly understand a user's intent? Developing an understanding of a model's representation of the user is especially important as people use LLMs for personal questions and even as AI companions. Without an understanding of models' behaviors, model builders risk harming users' well-being \citep{cheng2025sycophantic}. 

\paragraph{Uncovering unintended effects of post-training.} Another key application of interpretability for HCLLM is a better understanding of post-training procedures like preference alignment \citep{ferrao2025anatomy, movva2025s}. Without more interpretable approaches, post-training can lead to various unintended effects (e.g. sycophancy) that can be difficult to monitor or mitigate post-hoc. Building on existing approaches that refine our understanding of what preference alignment actually optimizes for, model providers can better control and steer behaviors towards desirable directions. For example, representation finetuning and steering \citep{rimsky2024steering, wu2025improved, wu2025axbench, wu2024reft} have been shown to be a promising way to control an LLM's behaviors, and these methods can be applied to elicit behaviors that are aligned with users' long-term development. The key first step towards making models safer and more steerable towards long-term beneficial objectives would be to understand how they work.
\hypertarget{steerability}{\subsection{Steerable HCLLMs}}
\label{subsec:steerability}

\subsubsection{Current Approaches to Steerability}

\textbf{Steerability} is the second dimension we highlight for HCLLM deployment. Steerability is the degree to which a model can be aligned along a particular dimension
\citep{miehling2024evaluatingpromptsteerabilitylarge}, such as preferences, norms, or user-constraints \citep{chen2025steer}. In contrast to static notions of alignment that aim to produce a single globally acceptable behavior, steerability emphasizes conditional control, or the ability to modulate LLM outputs in accord with its particular users. This property is especially important for HCLLMs, which are designed to interact with diverse users embedded in heterogeneous social, cultural, and institutional contexts. 

Steerability spans multiple dimensions. First, \textbf{personalization} is the goal in which an LLM's outputs adaptively reflect the preferences of individual users or groups of users, along with their prior knowledge, goals, and needs \citep{tseng2024two}. Personalized models should be able to provide more relevant recommendations \citep{hu2024enhancing}, and they should be calibrated to the user's preferred writing styles \citep{zhang2024personalization}, learning styles \citep{park2024empowering}, and norms around privacy \citep{shao2024privacylens,asthana2024know} and social behavior \citep{li2024agentalignmentevolvingsocial}. A user's preferences can be derived from explicit feedback, as in pairwise preference datasets, or from implicit feedback like historical interaction data. For more in-depth discussion of personalization methods, see \S\ref{subsec:personalization}.

Related to personalization is \textbf{persona alignment} or role play. Here, the goal is that HCLLMs will adopt consistent identities or roles, like software developers, expert tutors, empathetic counselors, or skeptical reviewers, which remain stable across interactions \citep{li2024steerability, samuel2024personagym, shanahan2023role}. Persona alignment is especially important in multi-agent settings where multiple LLM personas interact and collaborate \citep{park2023generative,guo2024large}.

In addition to better personalization and role playing, steerable HCLLMs should be able to adaptively understand low-resource languages, dialects, or sociolinguistic varieties \citep{ziems2023multi}. We refer to this target as \textbf{linguistic alignment}. This form of steerability is critical for equitable access, as models trained predominantly on high-resource, standardized corpora often underperform for marginalized linguistic communities (see \S\ref{subsec:data_representation}). Finally, \textbf{cultural alignment} means that models can be steered to reflect the norms, values, and narratives of particular communities or demographic groups \citep{santurkar2023opinionslanguagemodelsreflect}. Unlike personalization, which targets individuals, cultural alignment operates at the level of shared practices and collective meaning-making.

A range of technical mechanisms support steerability. At inference time, models may be steered through in-context learning, prompt engineering, or output filtering \citep{wies2023learnability}. Post-hoc control methods can condition generation on specific attributes or enforce constraints via decoding strategies. More structurally, personalized reward models and fine-tuning procedures can encode user- or group-specific objectives into model parameters \citep{chen2024pal}. 

Steerability is still fundamentally constrained by the representational biases embedded in pre-training and post-training data \citep{mihalcea2025ai}. If certain identities, linguistic forms, or cultural narratives are underrepresented or stereotyped in the training corpus, then prompt-based steering may have limited expressive range. In this sense, steerability is not just a matter of control at inference time, but is rooted much earlier in data provenance (\S\ref{subsec:data_provenance}) and evaluation (\S\ref{sec:evaluation}). Steerability starts with measuring where biases arise, localizing their sources in the data pipeline, and redesigning collection and annotation practices accordingly.

\subsubsection{Looking Forward}

Looking forward, we envision several research directions that extend current notions of steerability. First, we consider continual learning for preference, culture, and pluralistic alignment. Continual preference learning would allow models to adapt dynamically to evolving user needs by tracking implicit cues in interaction patterns or contextual data \citep{shaikh2025gum}. Such systems often rely on persistent memory stores like chat histories or retrieval-augmented generation. A central challenge is achieving this adaptivity while preserving user autonomy and privacy \citep{zhang2025autonomy}. In cultural and pluralistic alignment, rather than optimizing toward a static representation of \textit{culture} or \textit{user values}, HCLLMs should accommodate evolving norms and intra-group disagreement, as supported by continual learning. One way to elicit these evolving norms is to facilitate community-centered discussion and debate, following methods like STELA \citep{bergman2024stela}. Rather than treating communities as homogeneous preference aggregates, pluralistic alignment frameworks should model disagreement as a first-class signal. 

Pluralistic alignment comes with an array of technical and political challenges that will need to be addressed. On the technical side, post-training can induce mode-collapse, in which heterogeneous group preferences and opinions are compressed in a lossy manner, suppressing minority viewpoints and preserving only the majority preferences for a given group \citep{bisbee2024synthetic,durmus2024towards,rottger2024political}. To elicit diverse generations from LLMs, some inference time methods use iterative prompting \citep{hayati-etal-2024-far,feng-etal-2024-modular}, but these methods are not calibrated to real-world distributions. Other methods involve modifying the prompt with explicit identity terms for diverse user groups \citep{Giulianelli2023WhatCN}, but identity coded names induce models to draw on distributions of stereotypical representations or out-group perceptions rather than in-group perspectives \citep{wang2025large}.  To address mode-collapse in a manner that preserves in-group perspectives, it may be necessary to update LLM priors implicitly through distributionally-aligned in-context examples, following methods like spectrum tuning \citep{sorensen2025spectrum}.

It is not only a methodological challenge, but also a political challenge to achieve localized, domain-specific alignment processes that enable communities and stakeholders to meaningfully shape model behavior \citep{delgado2023participatory}. Methodologically, we have the participatory HCI approaches covered in \S\ref{subsub:participatory}. However, politically, most model developers lack incentives to share control with communities \citep{gabriel2020artificial}. Current alignment pipelines are centralized by the small set of companies with resources to develop LLMs. Similarly, academic institutions and governments can serve to centralize decision-making across the LLM development pipeline \citep{suresh2024participation}. There are a number of less centralized alternatives. For example, Masakhane \citep{orife2020masakhane} is a network of NLP researchers working on NLP for local African languages, with community involvement at every stage, from dataset creation and annotation to model training. EleutherAI is a distributed, volunteer-driven research collective that has created open pre-training corpora \citep{gao2020pile800gbdatasetdiverse}, models \citep{black2022gpt}, and scaling checkpoints \citep{biderman2023pythia}. BigScience was a research effort in which over 1,000 researchers from academia and industry coordinated through HuggingFace and built BLOOM \citep{workshop2022bloom}, a 176B-parameter multilingual autoregressive language model, with decision-making steps clearly and publicly documented.

These initiatives illustrate that steerability does not need to be confined to top-down fine-tuning interfaces or proprietary alignment pipelines. Distributed model development allows communities to steer models from the ground up, establishing linguistic resources, value functions, and development practices that ultimately shape downstream behavior. At the same time, decentralization introduces its own tensions, including coordination costs, uneven resource distribution, and challenges of accountability. Open and community-led efforts may broaden participation, but they must still grapple with questions of safety, quality control, and transparent maintenance of HCLLMs. In the following sections, we will discuss safety (\S\ref{subsec:safety}) and interpretability (\S\ref{subsec:interpretability}), as well as the tensions between these objectives.
\subsection{Safe HCLLMs}
\label{subsec:safety}

The third dimension we emphasize when building responsible HCLLMs is \textbf{safety}. As defined in \S\ref{subsec:safety_eval}, safety is conceptualized as preventing LLMs from producing undesirable outputs (i.e., those that may be toxic, harmful, discriminatory, or dangerous), even when prompted to do so. For example, the widespread use of LLMs raises critical concerns about ethical and social risks related to their outputs, including discrimination, hate speech, exclusion, misinformation harms, malicious uses, and so on~\cite{Zhang2022OPTOP,bender2021dangers}. At the same time, there are concerns that LLMs can be used as agents of harm, such as using AI-generated propaganda for misinformation purposes~\cite{goldstein2024persuasive} or spreading information that can facilitate harmful actions like the manufacturing of weapons~\cite{shaikh-etal-2023-second}. 

We begin by discussing the existing methods that are employed for addressing safety concerns both at the model training and interaction layers. Looking forward, we advocate for expanding beyond this current definition of safety, which focuses on preventing harms, to encompass how we can build HCLLMs that also maximize user benefits. 

\subsubsection{Current Approaches to Safety}

First, we will discuss current methods for measuring and mitigating safety concerns. \textit{Red-teaming} is a common practice for identifying harmful behavior through adversarial testing prior to deployment. Red-teaming approaches differ across model providers, and details are often not publicly disclosed as these practices are conducted in industry settings~\cite{feffer2024red}. As \citet{feffer2024red} survey, red-teamers typically come from three pools: subject-matter experts~\cite{ahmad2025openai}, crowdworkers~\cite{ganguli2022red}, or automated methods, such as the language models themselves~\cite{ganguli2022red,perez2022red}. The objectives for red-teaming can range from broad mandates to identify any harmful behavior to targeted assessments of specific risks, such as those related to national security. After deployment, model providers may also run bug bounty programs that offer incentives for discovering safety or security vulnerabilities~\cite{openai2025_gpt5_bio_bug_bounty,anthropic2025_model_safety_bug_bounty}. In addition to red-teaming efforts, model developers make use of benchmarks and other safety evaluations, which we discuss in detail in \S\ref{subsec:safety_eval}.

Mitigations can appear at various stages of the model development pipeline. At the pre-training phase, there is interest in filtering datasets to remove toxic content as a preventative safeguard~\cite{obrien2025deepignorance,mendu2025towards,stranisci2025they}. At the same time, other work has argued that filtering toxic data during pre-training can have detrimental downstream effects, and that including such data at pre-training can actually make these behaviors easier to remove through fine-tuning~\cite{li2025bad,maini2025safety,longpre-etal-2024-pretrainers}. Many efforts also address safety concerns during post-training. Foundational techniques for modern LLMs, such as RLHF, are useful not only for increasing model helpfulness but also for aligning models to be more harmless~\cite{ouyang2022training}. Building on these principles, additional post-training methods can help automate parts of this process. For instance, Constitutional AI allows researchers to pre-determine a set of ethical principles for the model to adhere to~\cite{bai2022constitutional}. Instruction-tuning methods can also reduce toxicity (see \S\ref{subsubsec:succeses_instruction_tuning}). Once deployed, guardrail models are used to help moderate both user inputs and generated outputs~\cite{inan2023llama,dong2024building,rebedea2023nemo}.

\subsubsection{Looking Forward}
Like other human-centered objectives, \textit{safety} can be an underspecified, ambiguous, or contested target. As we anticipate the development of more human-centered LLMs, we should be asking whose definitions of safety are prioritized, and how we can design models that not only prevent harm but can actively promote user flourishing.

\paragraph{Paying heed to long-term harms.} As surveyed above, existing AI safety research tends to focus on immediate harms that users face when interacting with models, such as exposure to toxic speech or the production of misinformation. Of course, these harms carry long-term societal consequences. However, an underexplored class of safety problems involves behavior that appears innocuous in the short term but can compound over repeated usage to become problematic. The recent work in this vein has identified specific model properties, such as sycophancy, which can affect users' psychological states and behaviors~\cite{cheng2025sycophantic}. An open challenge lies in measuring these long-term interaction harms, as they are difficult to capture with standard evaluation practices like benchmarking. One alternative approach, as discussed in \S\ref{sec:hci}, can be to run controlled experiments to understand the effect of model properties on users~\cite{cheng2025sycophantic,kirk2025neural} or to conduct qualitative inquiry~\cite{mathur2025sometimes}. However, this process can be time-intensive and furthermore potentially exposes participants to the very harms being studied. This concern raises the question of what alternative valid methods exist, such as those that can simulate these harms in silica. Beyond measurement, mitigations also remain largely underexplored --- both in terms of interventions at the model design and user interaction paradigms. Researchers have identified properties that can exacerbate harms (e.g., steering models towards being relationship-seeking in model design~\cite{kirk2025neural}, sending emotionally laden messages as users try to exit a platform~\cite{de2025emotional}), but translating these insights into preventative measures remains an important and open area for exploration. 

\paragraph{Expanding the definition of safety.} 
A second area of exploration involves expanding \emph{whose} definition of safety is prioritized. Definitions of safety vary considerably across demographic groups, along factors such as ethnicity, age, and gender~\cite{rastogi2025whose,ali2025operationalizing,movva2024annotation,gabriel2020artificial,aroyo2023dices}. These variances are then encoded into models through alignment processes, significantly changing model behavior~\cite{ali2025operationalizing}. Thus, instead of assuming a generic definition of safety, there is interest in better capturing and modeling the diversity of conceptualizations that exist. This direction presents a natural continuation of the existing focus on pluralistic alignment within human-centered LLM research. How do we capture these differing definitions of safety? Some work has tackled the issue through recruiting diverse sets of annotators to rate safety perceptions~\cite{rastogi2025whose,aroyo2023dices}. Others advocate for engaging with communities in a more participatory fashion to elicit safety goals, which offer a richer understanding of how different communities understand the potential harms of these technologies~\cite{qadri2025case,bergman2024stela}. 

Despite the benefits of moving away from this ``view from nowhere'' conception of safety, it is important to remember that communities themselves are not monolithic. Disagreements inevitably arise about what constitutes safe or harmful content or what model behaviors are deemed desired or unacceptable~\cite{gordon2022jury,gordon2021disagreement}. There is a legitimate concern that implementing democratic methods could inadvertently drown out the voices of minority groups. Yet this is not grounds to dismiss democratic or participatory methods for AI safety as a lost cause. As \citet{zimmermanndon} outline in their work, there are viable paths forward for reconciling these tensions by drawing on practices from political philosophy. As we move towards a more pluralistic definition of safety, this requires thinking normatively about the contexts in which we jointly maximize or balance considerations across groups of people, recognizing that collective decisions may at times conflict with individual desires or goals.

\paragraph{Considering not only harms but also benefits.} Finally, much of the discussion so far has focused on mitigating harmful behaviors. However, we emphasize that avoiding harm is not the same as maximizing user benefits. In pursuit of human-centered LLMs, we must also prioritize building models that bring positive change for users. This raises important questions about what beneficial model behavior looks like, and whether our current conceptions of safety align with what is truly beneficial. First, we must challenge the assumption that ``safe'' models are necessarily the best for promoting benefits. As \citet{cai2024antagonistic} challenge, perhaps there is a need to design \emph{antagonistic AI} systems that are ``actively dismissive, disagreeable, closed-off, critical, flippant, difficult, interrupting.'' Much like, for instance, a student may be challenged by their teacher in the learning process, when we design for benefits rather than merely minimizing harms, the desired model behavior changes.

A second provocation concerns the scope of benefit: rather than thinking only about the one-to-one benefit of a model on an individual user, what about one-to-many benefits? We can envision designing models for collective or group-level good --- for example, models deployed to promote democratic health by finding common ground through deliberation~\cite{tessler2024ai}, or models designed to benefit teams by serving as collaborators within group settings. Just as there are differing definitions of what safety means, there are similarly diverse conceptions of benefit, raising parallel questions about whose definition should be prioritized and how we reconcile conflicting views of what constitutes a beneficial outcome.
\clearpage
\hypertarget{applications}{\section{Case Study: HCLLMs and the Future of Work}}
\label{sec:applications}

LLMs are not only research technologies that a select group of people examine, use, and probe. They have become a part of the lives of everyday users, supporting tasks ranging from assisting with coding to giving relationship advice~\cite{tamkin2024clio,chatterji2025people,yang2024swe,jimenezswe}. For example, as of September 2025, OpenAI reported that their flagship chatbot, ChatGPT, saw over 700 million weekly active users~\cite{chatterji2025people}. So, we conclude by discussing how our considerations around the multiple facets of HCLLMs apply in real-world applications. For example, are HCLLMs practically feasible? How do HCLLMs affect individuals and what macro effects do HCLLMs have on society?

We focus this chapter on one particular application area: the future of work. Of course, this is not the only area where LLMs are used; these models have a wide scope of applications including healthcare, education, political science, and so on~\cite{thirunavukarasu2023large,adiguzel2023revolutionizing,ornstein2023train}. Nonetheless, we focus on HCLLMs' impact on labor and the future of work given public interest, as evidenced by the many headlines and speculation, as well as the individual and societal level impacts that models will have in this area~\cite{shao2025future,acemoglu2026building}.   

\begin{figure}[hb!]
    \centering
    \includegraphics[width=\linewidth]{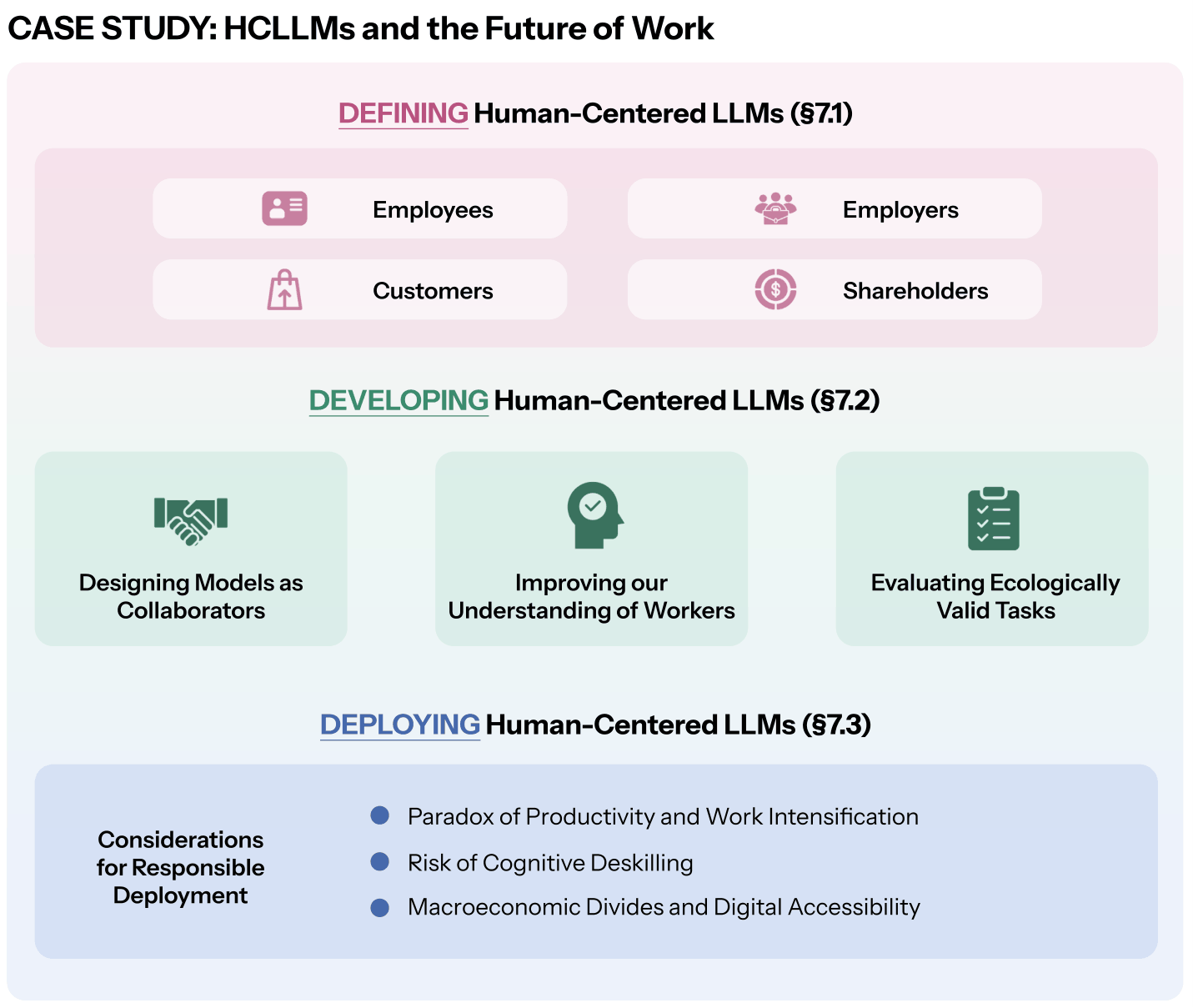}
   \caption{We present a case study on HCLLMs and the future of work, covering the three key areas of defining, developing, and deploying HCLLMs. We start by identifying relevant stakeholders (\ref{subsub:stakeholders_future_of_work}), then move to examining the model capabilities needed to better suit LLMs for workplace settings (\ref{subsub:evaluating_future_of_work}), and conclude by discussing key societal considerations (\ref{subsub:case_deploy}).}
    \label{fig:future_of_work}
\end{figure}

The adoption of LLMs in the labor market has led to significant shifts in human productivity, in-demand skills, and even possible macroeconomic changes~\cite{chen2025code,eloundou2024gpts}. We will now show how model developers can incorporate the human-centered principles from previous sections to define, develop, and deploy HCLLMs within this evolving ecosystem. To define HCLLMs here, we will first discuss \emph{who} the stakeholders are and how to account for these differing parties in \S\ref{subsub:stakeholders_future_of_work}. Then, we will cover HCLLM development, how we ought to be training and evaluating models for future of work purposes in \S\ref{subsub:evaluating_future_of_work}. Finally, in \S\ref{subsub:case_deploy}, we conclude with considerations for responsibly deploying HCLLMs, such as the potential for widening inequalities or overreliance. The road map is visualized in Figure~\ref{fig:future_of_work}.

\subsection{Defining the Stakeholders}
\label{subsub:stakeholders_future_of_work}
To understand how we can design LLMs for the future of work in a human-centered fashion, we must start by understanding the \emph{who}. Who is being impacted by HCLLMs in the workforce? Is this the same group of people that these models are being designed for? How might different groups of stakeholders --- both direct and indirect --- be impacted differently? There are many potential stakeholders, including \textbf{workers} who directly interact with LLMs; \textbf{employers} who may be in charge of procuring the technology for their organization; \textbf{shareholders} who are interested in productivity or financial gains; and \textbf{customers} who may see the final artifact or output created from workers using LLMs. 

A natural group to start with are the \textbf{workers} who interface and utilize these technologies as part of their work. Here, a recurring theme is a fundamental mismatch between what people actually want from LLMs and how those technologies are currently being designed. In part, this misalignment can be attributed to organizational constraints, which are more present in the work setting compared to personal use. For instance, \citet{nanda2025state} found that employees frequently resort to using personal accounts to access LLMs (e.g., ChatGPT, Claude) as they found enterprise deployments fail to meet their needs or feel too restrictive. Yet, this workaround behavior is symptomatic of a deeper issue. The way in which LLMs are currently being used in the workplace is misaligned with worker priorities. For example, \citet{shao2025future} conducted a survey with 1,500 U.S. workers to understand what tasks they want AI agents to be used for. Critically, they found that the tasks workers \emph{want} to use agents for differ substantially from how tools are currently deployed and where industry funding is going. In tandem with the human-centered reasons for centering workers' perspectives, this finding suggests that current AI development trajectories risk prioritizing displacement over augmentation. Doing so risks repeating dangerous historical patterns in which automation technologies lead to displacement without commensurate productivity gains or wage increases~\cite{acemoglu2021tasks}. To close this gap, we must ensure that the needs and desires of workers are incorporated into the design of these systems from the outset.

Taking into account the perspectives of \textbf{employers} or \textbf{shareholders}, the question becomes whether introducing LLMs improves the productivity and quality of work produced. What complicates this question is the ``jagged'' nature by which LLMs are useful~\cite{dell2023navigating}. For some tasks, LLMs are particularly performant and can automate existing process; some roles will see more of a synmbiotic relationship where there is augmentation rather than replacement; and in others, LLMs are in fact not capable at performing requisite tasks at all~\cite{mazeika2025remote}. There is a parallel consideration around literacy --- whether workers are well-equipped to use the technology. Harkening back to the ``gulf of envisioning'' discussed in Chapter~\ref{sec:hci}, it is possible that LLMs may actually be useful for workers, but workers may not know how to best specify their intent or be unaware that such capabilities exist. This places a responsibility on employers, organizations, and policymakers to invest in AI literacy programs that equip workers with the conceptual and practical knowledge needed to participate in an AI-augmented workplace, rather than simply assuming adoption will follow deployment~\cite{ma2025not,nanda2025state}.

Finally, another stakeholder we will highlight are \textbf{customers} that serve as another end-user in this setting. For instance, as LLM-based systems increasingly replace human touchpoints in domains such as customer service or healthcare, the impact on customers warrants equal consideration. The evidence here is mixed: while interacting with LLM systems offers speed and accuracy compared to human support, these interactions can also raise frustrations around the lack of empathy or potential for miscommunication~\cite{huang2024can,li2025artificial}. Furthermore, recent survey data shows that customers in the U.S. still prefer talking with a human representative over an LLM system~\cite{customer2026}. This concern is heightened in contexts that may be more high-stakes or emotionally-laden. Nonetheless, we also want to highlight that treating this question of AI usages as a binary choice between human-only and AI-only presents an inchoate view of the issue as there is meaningful middle ground that leverages the strengths of both. For instance, existing work has explored how we can use LLMs to upskill professionals to offer better human support, preserving this human ``touch'' in interaction while enhancing the quality of accessibility.  

\subsection{Developing HCLLMs for the Future of Work}
\label{subsub:evaluating_future_of_work}
Next, we discuss how we can apply a human-centered lens to training and evaluating LLMs to enable a future of work that jointly benefits the different stakeholders we have discussed. To start, much of the current discourse around LLM systems emphasizes autonomous execution, or systems that complete complex, multi-step tasks with minimal human involvement~\cite{shen2025completion}. Yet, focusing primarily on this form of autonomous interaction, overlooks the potential gains for systems in which humans and LLMs collaborate. For example, researchers such as \citet{wang2026position} have raised exactly this concern in the context of AI coding agents --- a domain where models are quite performant and has seen significant adoption especially amongst professionals in this field~\cite{appel2025anthropic}. They argue that the field has moved too quickly toward full automation and that human involvement deserves to be treated as a first-class design consideration rather than a transitional inconvenience. Taking this critique seriously requires asking what it would actually take, technically, to build human-centered LLMs for the workforce. 

\paragraph{Designing Models as Collaborators.} One prerequisite is that models need to know \emph{how} to collaborate. However, collaboration is not simply a matter of adding a confirmation step before a model takes action. It requires a more nuanced understanding of when to act autonomously, when to defer, and how to communicate uncertainty or request input in ways that feel natural rather than disruptive. As \citet{shen2025completion} show, building more capable models does not necessarily equate to better collaboration, thus necessitating dedicated efforts. In fact, current training paradigms, such as preference tuning over single conversational turns, can even hinder collaboration abilities~\cite{wu2025collabllm}. Furthermore, successful collaboration does not mean indiscriminately putting a human in the loop. While ostensibly beneficial, this interaction paradigm can lead to unnecessary verification and inefficiencies that ultimately slow down the collaboration process. The best form of collaboration will necessarily vary, depending on the task at hand as well as the skillset of the human and LLM involved. For example, sometimes the human may need to audit the LLM's work; other times the right form of collaboration may need to be a back-and-forth conversation; and in others end-to-end execution by the model may suffice. 

To start, we must return to the core question: what are the ingredients for a successful collaboration? In human organizations, a common collaboration paradigm is \emph{delegation}. At the moment, agents are capable of decomposing complex tasks into more manageable units of action. However, the next step here is being able to assign these actions. Thus, we need to know what tasks to hand-off to these models and which may be better suited for humans to complete or even to delegate across multiple different models~\cite{wang2025ai}. This delegation capability must furthermore be adaptive across users, tasks, context, and model capabilities~\cite{tomavsev2026intelligent, guggenberger2023task, fugener2022cognitive}. Beyond being able to delegate, there are many other proprties that help make a good collaborator which are lacking in existing models, such as proactivity~\cite{lu2024proactive}, transparency~\cite{liao2023ai}, and social norm adherence~\cite{shaikh2025gum,ziems2023normbank,forbes2020social}--- all presenting fruitful areas for future inquiry. 

\paragraph{Improving our Understanding of Workers.} A second technical requirement is a richer understanding of the users these systems are meant to serve. One push in this direction is developing better user simulators which can capture the behavioral patterns and habits of individuals~\cite{lei2026humanllm,paglieri2026persona}. As our ability to simulate users improves, a question we ought to return to is \emph{which} users are being simulated. At the moment, many efforts are concentrated on generic user behavior or focused more so on knowledge workers (e.g., developers)~\cite{naous2025flipping,wang2026position}. This provides a deep but narrow viewpoint of the workforce. The population of workers whose jobs will intersect with LLMs is far broader, spanning domains like healthcare, retail, education, logistics, and skilled trades~\cite{handa2025economic,tomlinson2025working}. Developing better user simulators for these contexts requires going beyond theoretical modeling to gain a deeper understanding of how people are actually using these systems in the real world. This is a genuine challenge, because the most valuable behavioral data tends to be held by the companies deploying models and is rarely made available to the broader research community. Finding ways to bridge this gap --- whether through privacy-preserving data sharing, partnerships, or the careful design of in-the-wild studies --- is a prerequisite for building systems that serve a broader swath of workers. 

\paragraph{Evaluating for Ecologically Valid Tasks.} Finally, across chapters we have emphasized the importance of data in the development and progress towards human-centered LLMs. To see progress in this area, we must also have benchmarks that actually reflect the complexity of real work. Many of the standard benchmarks used to evaluate LLM performance consist of synthetic or highly constrained tasks that bear little resemblance to the messy, context-dependent, and often ambiguous nature of professional work. More general-purpose benchmarks, such as GDPval~\cite{patwardhan2025gdpval}, represent a meaningful step toward measuring model performance on more ecologically valid tasks, and domain-specific benchmarks provides an even more precise look as to how these models perform in settings such as finance~\cite{fan2025ai}, law~\cite{guha2024legalbench,li2025legalagentbench}, and medicine~\cite{arora2025healthbench}. In tandem, we must also consider what these benchmarks are failing to capture but that we may also want to evaluate. For instance, how can we better account for how workers are actually interacting with the systems or how the performance holds up in the messy, open-ended conditions of real work? 
 
\subsection{Responsibly Deploying HCLLMs in the Workforce}
\label{subsub:case_deploy}
Perhaps one of the top-of-mind questions when it comes to the future of work are the long-term societal impacts that these technologies will have. How will employment be affected by the increasing popularity of LLMs? What skills will be in-demand and which will be less important? Deploying HCLLMs responsibly means taking into account these long-term externalities now, and mitigating against potential harms before they occur. 

\paragraph{The Paradox of Productivity and Work Intensification.}
A common perception is that integrating LLMs into workflows will inherently reduce the volume of human effort. Initial stand-alone studies looking at whether using LLMs improves how quickly workers complete tasks show positive evidence: people complete tasks more quickly with AI assistance~\cite{cui2025effects,peng2023impact,shen2026ai,karny2024learning}. However, when it comes to the workforce, people are not completing singular tasks in a vacuum. They must negotiate tasks with many demands, coordinate with coworkers, navigate company politics, and so on. Thus, when we look at the impact on productivity in a more holistic setting, the impact of LLMs is less clear. Counter to the earlier studies, recent work has suggested that rather than freeing up workers, LLMs may actually amplify the intensity of labor~\cite{hbr_ai_work_intensification,harvey2025don}. For example, a recent study conducted by the Harvard Business Review observed 200 employees at a U.S.-based technology company, finding that LLM usage promotes the phenomenon of ``work slippage'' in which AI leads to task expansion and increased cognitive load rather than a net reduction in hours~\cite{hbr_ai_work_intensification}. This phenomenon is not solely confined to the technology industry where LLM usage is more prevalent~\cite{harvey2025don,johnson2025ai,acemoglu2020wrong}. For example, despite the purported benefits that LLMs have for educators, using models often lead to more teacher time devoted towards validating generated outputs beyond current time spent~\cite{harvey2025don}. Thus, instead of saving time, these tools may in fact create new tasks for workers or potentially introduce friction into existing tasks.

To address these risks, we provide two approaches. First, we argue that the way ``productivity'' is measured in the workplace in light of advancements with LLMs must be updated. Traditionally, productivity has centered on output per unit time, capturing how efficient people work~\cite{mckinsey2025}. However, now that LLMs can quickly produce outputs, efficiency-based metrics may be defunct. Instead, echoing practices in existing work~\cite{shao2024collaborative,wang2025ai}, we call for broader evaluations that consider not only whether tasks are completed but also the \emph{quality} of these outcomes. More broadly, updating productivity metrics to match the current state of the world provides a more precise picture of how LLMs are impacting work. Our second approach is to find methods that center workers' voices. Survey efforts such as \citet{shao2025future} provide insight into what workers want. Complementing this approach, we argue for additional studies that can capture the ``thick'' descriptions of what users are experiencing, such as through interviews or ethnographic approaches~\cite{geertz2008thick,anugraha2026sparkme}. Such methods can shed light on what is not captured through numbers alone, such as hidden forms of labor, additional cognitive load, or coordination costs. Together, these directions point toward a more holistic framework for evaluating AI in the workplace that accounts for both outcomes and worker experience.

\paragraph{The Risk of Cognitive Deskilling.}
A second concern is the potential for deskilling, where reliance on LLMs erodes the foundational expertise of human workers. When models handle the ``first draft'' (or sometimes the end-to-end execution) of complex tasks, workers may lose the opportunity to engage in learning processes necessary to develop deep domain mastery. Perhaps in an isolated setting workers may be completing the task faster, but they are also less likely to learn transferable skills or knowledge that can help them in future tasks~\cite{shen2026ai}. It is important to note that these effects are heterogeneous across workers. How someone engages with these technologies matters: automating tasks wholesale is likely to accelerate deskilling, whereas more collaborative, iterative interactions where the worker critically evaluates, corrects, and builds on model outputs may partially preserve or even scaffold skill development~\cite{shen2026ai}. Another moderator is expertise level. For instance, novices may rely on these tools more heavily and more uncritically than experts, which is especially concerning given that AI assistance can impart an illusion of comprehension without providing sufficient expertise in an area~\cite{messeri2024artificial,shen2026ai,macnamara2024does}. 

These concerns highlight the importance of carefully designing how humans collaborate with LLMs, as discussed in \S\ref{subsub:evaluating_future_of_work}. For example, the risk of deskilling can be included as a factor when considering how to delegate tasks. In this vein, rather than framing LLMs as tools that automate tasks end-to-end, systems should be designed to empower users, supporting human judgment, reflection, and skill development throughout the workflow. To achieve this goal, this requires intentional design interventions that encourage workers to remain actively engaged with the task, such as mechanisms that promote critical evaluation of model outputs or structures that preserve opportunities for learning and expertise development~\cite{ma2025towards,reicherts2025ai}.

\paragraph{Macroeconomic Divides and Digital Accessibility.}
Finally, the integration of LLMs also threatens to exacerbate existing economic disparities. Traditional automation has historically widened the wage gap between high-skilled and low-skilled workers \cite{acemoglu2021tasks}, but LLMs introduce a specific ``productivity divide'' rooted in accessibility. Research into usage patterns suggests that ChatGPT adoption is positively correlated with higher education, high socioeconomic status, and residency in urbanized zip codes \cite{daepp2024emerging}. If access to state-of-the-art models remains concentrated within privileged demographics, the resulting disparity in AI-augmented productivity could further exacerbate systemic inequality.

Addressing this divide requires interventions at multiple levels. For one, as open-source models grow more capable, this can help reduce barriers to entry and broaden who is able to benefit from AI assistance. However, access alone is insufficient. As discussed in \S\ref{subsec:bias_eval}, model performance and usability can vary across social groups, meaning that some populations may benefit less from interacting with LLMs even when they are available~\cite{hofmann2024dialectprejudicepredictsai, hassan2025dialectic, durmus2023towards}. Continued efforts to identify and mitigate such disparities in model behavior remain critical. Finally, improving AI literacy will be essential as a new generation of workers grows up with these technologies at their fingertips. Educational initiatives that teach users how to critically evaluate, effectively collaborate with, and responsibly leverage AI systems can help ensure that the productivity gains from LLMs are distributed more equitably~\cite{solyst2025investigating,morales2024youth,morales2025learning,okolo2024beyond,cardon2023challenges}.
\clearpage 
\hypertarget{conclusion}{\section{Conclusion}}
\label{sec:conclusion}

The development of large language models has reached a critical juncture. It is no longer sufficient to ask only what these models are capable of doing. We must also grapple with questions such as who is --- and is not --- involved in and accounted for in the creation of LLMs; what the impacts of these models are at both the individual and societal level; and which values and principles these technologies uphold and promote. This survey examines how such human-centered principles are inherently intertwined with the design, training, and deployment of LLMs.

Ultimately, the trajectory of LLM development must be guided by more than technical benchmarks and capability milestones. The questions of inclusion, impact, and values explored in this survey are not peripheral concerns to be addressed after the fact; they are foundational to what these systems become and who they serve. By centering human-centered principles at every stage of the LLM lifecycle, from design and data curation to training and deployment, researchers and practitioners can work toward models that are not only more capable, but also more equitable, accountable, and aligned with the diverse needs of the people they affect. The path forward demands a broader coalition of voices, a more expansive notion of responsibility, and a sustained commitment to ensuring that progress in AI is measured not only by what these models can do, but by the kind of world their development helps to build.

\section*{Acknowledgments}
The idea for this manuscript originated in the Fall 2024 offering of Stanford’s CS 329X course on HCLLMs, taught by the instructor Diyi Yang and assistants Rose E. Wang and Caleb Ziems. Diyi Yang developed the initial structure and outline of the paper. The enrolled students collaborated on a first draft of the manuscript as part of their coursework, with each student pair assigned responsibility for drafting a subsection of the survey according to the course assignment structure. The teaching assistants subsequently reviewed, graded, and provided feedback on these drafts. In Winter 2025, a subset of students continued to revise and expand the manuscript under the primary direction of Rose E. Wang and the secondary direction of Caleb Ziems. 

Caleb Ziems and Dora Zhao largely re-wrote and restructured the manuscript, with significant conceptual revisions and new chapters, to produce the current version. This revision phase was supported by additional contributions from Sunny Yu and Advit Deepak. All authors reviewed and approved the final manuscript. 

The core authors were jointly responsible for deciding on and writing the paper in its final form. The core are as follows:

\begin{itemize}
    \item \textbf{Diyi Yang} designed the overall structure, chapter outline, and course material that initialized this survey. She supervised the project and provided guidance on the direction and scope of the survey at every stage.
    \item \textbf{Caleb Ziems} restructured the paper from its initial conception, wrote \S\ref{sec:introduction} and \S\ref{sec:data}, and co-wrote \S\ref{sec:nlp}, \S\ref{sec:evaluation}, and \S\ref{sec:responsible}. He also contributed to structuring and reviewing all student drafts, and co-lead the second round of student revisions. 
    \item \textbf{Dora Zhao} restructured the paper from its initial conception, wrote \S\ref{sec:hci} and \S\ref{sec:applications}, and co-wrote \S\ref{sec:nlp}, \S\ref{sec:evaluation}, and \S\ref{sec:responsible}. Dora also co-designed the figures.
\end{itemize}

Additionally, the leadership of this work included:

\begin{itemize}
    \item \textbf{Rose E. Wang} contributed to structuring and reviewing all student drafts. She led the second round of student revisions.

    \item \textbf{Matthew Jörke} contributed \S\ref{subsec:technical_artifacts} and provided suggestions on \S\ref{sec:hci} more broadly.
    
    \item \textbf{Ahmad Rushdi} contributed guidance and final edits.
\end{itemize}

Student contributions are as follows:

\begin{itemize}
    \item \textbf{Anshika Agarwal} co-wrote \S\ref{subsub:quantitative_methods} and edited \S\ref{sec:evaluation}.

\item\textbf{Harshvardhan Agarwal} helped with the course paper and edited \S\ref{sec:evaluation}.

\item\textbf{Gabriela Aranguiz-Dias} co-wrote \S\ref{subsub:quantitative_methods}.

\item\textbf{Aditri Bhagirath} helped with the course paper and second-round student edits.

\item\textbf{Justine Breuch} co-wrote \S\ref{subsec:bias_eval}.

\item\textbf{Huanxing Chen} helped with the course paper.

\item\textbf{Ruishi Chen} helped with the course paper.

\item\textbf{Sarah Chen} co-wrote the first draft of \S\ref{subsec:interpretability}.

\item\textbf{Advit Deepak} co-wrote \S\ref{sec:responsible}.

\item\textbf{Haocheng Fan} co-wrote \S\ref{subsec:human_centered_eval}

\item\textbf{William Fang} helped with the course paper and second-round student edits.

\item\textbf{Cat Gonzales Fergesen} helped with the course paper.

\item\textbf{Daniel Frees} co-wrote \S\ref{subsec:instruction_tuning}.

\item\textbf{Tian Gao} co-wrote \S\ref{subsec:safety_eval}.

\item\textbf{Ziqing Huang} co-wrote \S\ref{subsec:safety_eval}.

\item\textbf{Vishal Jain} co-wrote \S\ref{subsec:safety}

\item\textbf{Yucheng Jiang} co-wrote \S\ref{subsec:human_centered_eval}

\item\textbf{Kirill Kalinin} helped with second-round student edits.

\item\textbf{Su Doga Karaca} co-wrote \S\ref{subsec:human_centered_eval} and edited \S\ref{sec:evaluation}.

\item\textbf{Arpandeep Khatua} helped with the course paper and second-round student edits.

\item\textbf{Teland La} helped with the course paper.

\item\textbf{Isabelle Levent} helped with the course paper.

\item\textbf{Miranda Li} helped with the course paper and second-round student edits.

\item\textbf{Xinling Li} co-wrote \S\ref{subsec:data_privacy}.

\item\textbf{Yongce Li} co-wrote \S\ref{subsec:synthetic_data}.

\item\textbf{Angela Liu} helped with the course paper and second-round student edits.

\item\textbf{Minsik Oh} co-wrote \S\ref{subsub:quantitative_methods}, and edited \S\ref{sec:evaluation}.

\item\textbf{Nathan J. Paek} helped with the course paper and second-round student edits.

\item\textbf{Anthony Qin} helped with the course paper.

\item\textbf{Emily Redmond} co-wrote \S\ref{subsec:scaling}.

\item\textbf{Michael J. Ryan} wrote \S\ref{subsec:pluralism} and co-wrote the remainder of \S\ref{sec:nlp}.

\item\textbf{Aadesh Salecha} co-wrote \S\ref{subsec:bias_eval}.

\item\textbf{Xiaoxian Shen} co-wrote \S\ref{subsec:data_privacy}.

\item\textbf{Pranava Singhal} helped with the course paper.

\item\textbf{Shashanka Subrahmanya} co-wrote \S\ref{subsec:impact}

\item\textbf{Mei Tan} co-wrote \S\ref{subsec:benchmarks}.

\item\textbf{Irawadee Thawornbut} helped with the course paper.

\item\textbf{Michelle Vinocour} helped with the course paper.

\item\textbf{Xiaoyue Wang} co-wrote \S\ref{subsec:synthetic_data}

\item\textbf{Zheng Wang} co-wrote \S\ref{subsec:interpretability}.

\item\textbf{Henry Jin Weng} helped with the course paper.

\item\textbf{Pawan Wirawarn} helped with the course paper.

\item\textbf{Shirley Wu} helped with the course paper.

\item\textbf{Sophie Wu} co-wrote \S\ref{subsec:preference_tuning}.

\item\textbf{Yichen Xie} co-wrote \S\ref{subsec:preference_tuning}.

\item\textbf{Patrick Ye} helped with the course paper and second-round student edits.

\item\textbf{Sunny Yu} co-wrote \S\ref{subsub:quantitative_methods} and \S\ref{sec:responsible}.

\item\textbf{Sean Zhang} helped with the course paper and second-round student edits.

\item\textbf{Yutong Zhang} co-designed Figures~\ref{fig:overview}, \ref{fig:hci}, \ref{fig:evaluation}, \ref{fig:responsible}.

\item\textbf{Cathy Zhou} co-wrote \S\ref{subsec:instruction_tuning}.

\item\textbf{Yiling Zhao} co-wrote \S\ref{subsec:multilinguality}.

\end{itemize}

\clearpage
\bibliographystyle{ACM-Reference-Format}

\typeout{}
\bibliography{main}

\end{document}